\documentclass{article}

 \usepackage[main, final]{neurips_2026}

\usepackage[utf8]{inputenc} 
\usepackage[T1]{fontenc}    
\usepackage{hyperref}       
\usepackage{url}            
\usepackage{booktabs}       
\usepackage{amsfonts}       
\usepackage{nicefrac}       
\usepackage{microtype}      
\usepackage{xcolor}         
\usepackage{amsmath}
\usepackage{amsfonts}
\usepackage[ruled,vlined]{algorithm2e}
\usepackage{makecell}
\usepackage{multirow}
\usepackage{multicol}
\usepackage{graphicx}
\usepackage{subfigure}
\usepackage{float}
\usepackage{subcaption}
\usepackage{caption}

\usepackage{xcolor}

\definecolor{tscolor}{RGB}{52, 120, 180}     
\definecolor{imgcolor}{RGB}{80, 160, 100}    
\definecolor{textcolor}{RGB}{230, 140, 60}   

\title{SSDA: Bridging Spectral and Structural Gaps via Dual Adaptation for Vision-Based Time Series Forecasting}

\author{%
Mingrui Zhang$^*$\quad Hanchen Yang$^*$ \quad Wengen Li$^\dagger$  \quad Xudong Jiang \\  \quad  \textbf{Yichao Zhang} \quad \textbf{Jihong Guan} \quad \textbf{Shuigeng Zhou} \\
\texttt{lwengen@tongji.edu.cn} \\
}

\begin{document}

\maketitle

\let\thefootnote\relax\footnote{$*$ Equal contribution, $\dagger$ Corresponding authors}

\begin{abstract}

Large vision models (LVMs) have recently proven to be  surprisingly effective time series forecasters, simply by rendering temporal data as images. This success, however, rests on a largely unexamined premise: the rendered time series images are sufficiently close to natural images for knowledge in pre-trained models to transfer effectively. We argue that two gaps still remain, i.e., \textbf{spectral and structural gaps}, fundamentally limiting the potential of LVMs for time series forecasting. \emph{Spectrally}, we systematically reveal that rendered time series images exhibit a markedly shallower power spectrum than the natural images LVMs are pre-trained to recognize. \emph{Structurally}, reshaping 1D temporal sequences into 2D grids fabricates spurious spatial adjacencies while severing genuine temporal continuities, misleading the spatial inductive biases of pre-trained LVMs. To bridge these gaps, we propose \textbf{SSDA}, a dual-branch network that spectrally and structurally adapts to unlock the full potential of LVMs for time series forecasting. At the data level, a \textbf{Spectral Magnitude Aligner (SMA)} applies 2D FFT to selectively enhance the magnitude spectrum toward natural-image statistics while preserving phase. At the model level, a \textbf{Structural-Guided Low-Rank Adaptation (SG-LoRA)} injects position-aware temporal encodings into patch embeddings and adapts attention via low-rank updates. The two branches are further adaptively fused to produce the final forecast. Extensive experiments on seven real-world benchmarks demonstrate that SSDA consistently outperforms strong LVM- and LLM-based baselines under both full-shot and few-shot settings. Code is publicly available at \url{https://anonymous.4open.science/r/SSDA-8C5B}.

\end{abstract}

\section{Introduction}
\label{introduction}

Time series forecasting (TSF) is central to decision-making  across a wide range of domains, including energy~\cite{Koprinska2018}, transportation~\cite{yang2025towards}, healthcare~\cite{Morid2023}, climate science~\cite{zhang2026piformer}, and economics~\cite{yu2023harnessing}.
Inspired by the success of foundation models in natural language processing (NLP) and computer vision (CV), a growing body of work adapts large pre-trained backbones, such as GPT-2, BERT, CLIP, and LLaMA, into universal forecasters~\cite{liang2024foundation}. 
Such backbones offer rich, transferable representations learned from large-scale data, motivating their reuse as general feature encoders.
Early efforts, e.g., GPT4TS, UniTS, and Time-LLM, treat time series as text and feed them into large language models (LLMs), achieving superior gains over traditional methods~\cite{goswami2024,niu2024,zhou2023,timellm2024,loftllm2026}.
More recently, large vision models (LVMs) have emerged as an especially promising direction. By rendering 1D time series as 2D images and processing them with a pre-trained vision encoder, methods such as VisionTS~\cite{visionts2025} and DMMV~\cite{dmmv2025} outperform LLM-based counterparts on many TSF benchmarks~\cite{Harnessing2025,vilt2021}.
The rationale is intuitive: while both text and images differ from time series in modality, images align with the continuous structure of time series far more naturally. Tokenizing continuous values into discrete symbols,  as required by LLM-based approaches, is inherently lossy and disrupts numerical continuity. Rendering time series as 2D images, by contrast, turns trends, periodicity, and fluctuations into the form of gradients, textures, and edges that pre-trained LVMs are already trained to recognize~\cite{visionts2025}.

However, the success of LVMs relies on a largely unexamined assumption: the rendered time series images are sufficiently similar to natural images for the pre-trained knowledge of LVMs to transfer effectively.
Existing LVM-based forecasters typically resort to ad hoc strategies: directly feeding rendered time series images into a frozen Masked Autoencoder (MAE)~\cite{feichtenhofer2022masked}, fusing them with textual or numerical views~\cite{timevlm2025}, or filtering training data to mitigate severe mismatches~\cite{visionts++}. None of these approaches, however, explicitly answer the fundamental question: \emph{how far do rendered time series images deviate from natural images, and along which dimensions do such gaps hinder the reuse of LVM pre-training?} Without this understanding, efforts to adopt LVMs for time series forecasting remain largely heuristic and the full potential of LVMs remains out of reach.

We argue that this modality gap has two distinct facets: a \textbf{spectral gap} in the input data distribution and a \textbf{structural gap} in how time series are rendered to images.


\begin{figure}
    \centering
    \includegraphics[width=1\linewidth]{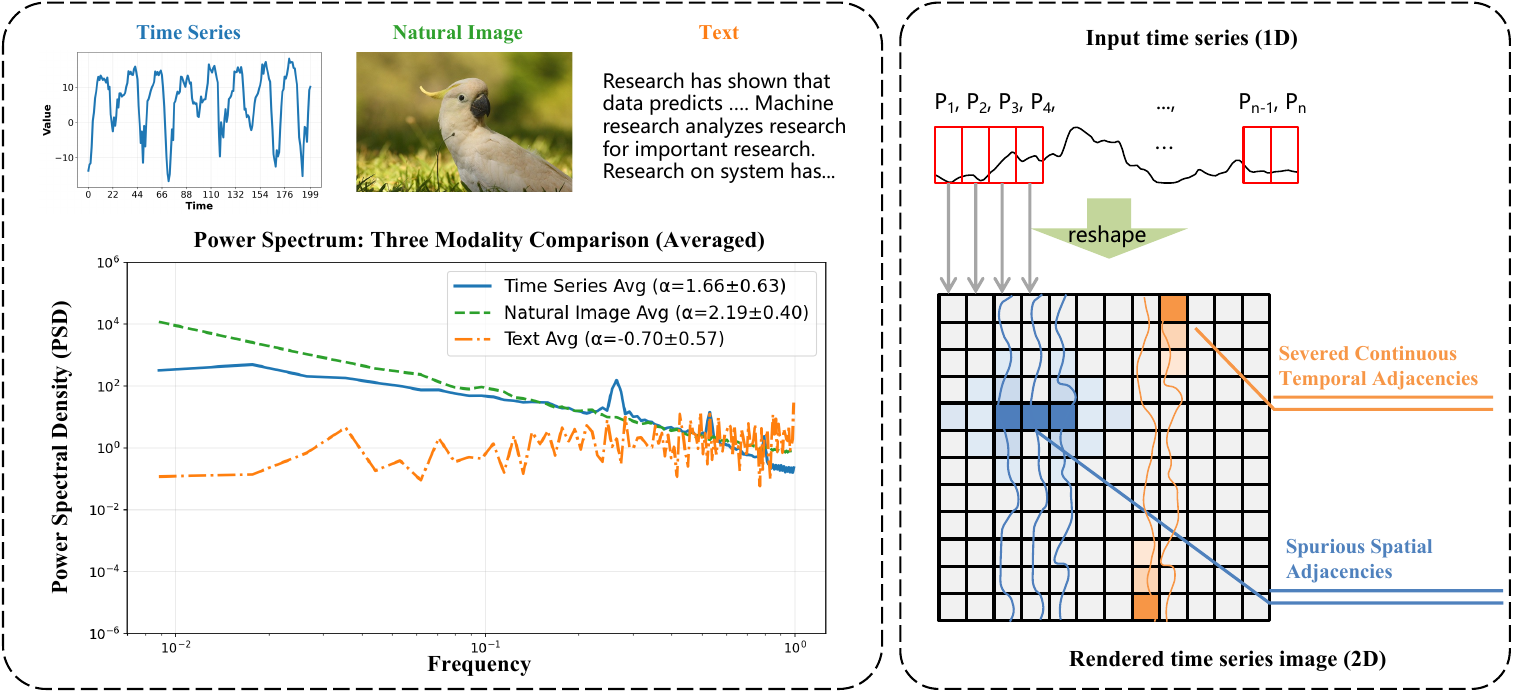}
    \caption{Illustration of the spectral and structural modality gaps. 
    \textbf{Left:} Spectral gap. Top: one representative sample from three modalities, i.e., a time series from ETT, a natural image from ImageNet, and a text snippet from Wikipedia. 
    Bottom: power spectra {averaged over a large set of samples per modality}, fitted with the power-law model $P(f) \propto f^{-\alpha}$. The fitted exponents \textcolor{tscolor}{Time Series ($\alpha=1.66\pm 0.63$)}, \textcolor{imgcolor}{Natural Images ($\alpha=2.19\pm 0.40$)}, and \textcolor{orange}{Text ($\alpha=-0.70\pm 0.57$)} reveal substantial distributional differences across modalities.
    \textbf{Right:} Structural gap. Reshaping a 1D time series into a 2D image via periodic folding introduces {spurious spatial adjacencies} between temporally distant points  and {severs continuous temporal adjacencies} at column boundaries.}
    \label{fig:modality_comparison}
\end{figure}

    

The \textbf{spectral gap} concerns the low-level statistics of the image itself. Natural images are well known to follow a $1/f^{\alpha}$ power spectrum with $\alpha \approx 2.0$~\cite{Field1987,ruderman1994}, a regularity that LVM pre-training implicitly exploits. To probe whether rendered time series images share this property, we adopt the Power Spectrum Slope (PSS)~\cite{pss2022,fredn2026}, a compact frequency-domain descriptor that has been used to characterize signals across modalities~\cite{timevlm2025,zeng2024,text2freq2024,nguyen2026}. 
As shown in the left panel of Figure~\ref{fig:modality_comparison}, samples from the ETT dataset exhibit an average power spectrum slope of $\alpha \approx 1.66$, significantly lower than that of natural images ($\alpha \approx 2.19$), while text rendered as a visual signal yields $\alpha \approx -0.70$. The three modalities thus occupy distinctly different regions of frequency space: the gap between rendered time series and natural images ($\Delta\alpha \approx 0.53$) is substantially smaller than that between time series and text ($\Delta\alpha \approx 2.36$), confirming that the visual modality aligns with time series far more closely than text.
To our knowledge, this is the first cross-modal spectral comparison of time series, images, and text in the context of LVM-based forecasting. Full details are given in Appendix~\ref{apd:modality_gap_analysis}.
   
The \textbf{structural gap} arises in how the model constructs and captures relations between patches. As illustrated in the right panel of Figure~\ref{fig:modality_comparison}, reshaping a 1D sequence into a 2D grid via periodic folding creates two simultaneous distortions. First, it fabricates spurious spatial adjacencies. Existing methods usually arrange consecutive time steps as columns~\cite{visionts2025}, but pixels in the same row yet different columns are treated as neighbors by the LVM's spatial attention, even though they may be hours or days apart in the original series. Second, it severs genuine temporal adjacencies. The last point of one column and the first point of the next are temporally consecutive yet spatially distant in the rendered image. A pre-trained LVM, whose inductive biases are calibrated to true spatial structure, fails to distinguish these distortions, leading to degraded forecasting performance. 

Motivated by the above observations, we propose \textbf{SSDA}, a dual-branch network in which each branch is the natural remedy for one gap at the level where that gap actually arises. The \textbf{spectral branch} addresses the spectral gap at the \emph{data level} through a \textbf{Spectral Magnitude Aligner (SMA)}: rendered images are decomposed by 2D Fast Fourier Transform (FFT), their magnitude spectrum is selectively enhanced by a lightweight learnable module while the phase is preserved to retain temporal structure, and the result is passed to a frozen MAE, maximally preserving the visual knowledge accumulated during pre-training. The \textbf{structural branch} addresses the structural gap at the \emph{model level} through \textbf{Structural-Guided LoRA (SG-LoRA)}: a Temporal Grounding Adapter (TGA) maps each spatial patch back to its original 1D temporal index, and injects a sinusoidal temporal encoding into the patch embedding via a learnable fusion gate, and LoRA adapters fine-tune the attention projections so that the model reconciles spatial layout with true temporal order. The two branches are fused by a learnable scalar weight and the result is denormalized to produce the final forecast.

Our contributions are threefold:
\begin{itemize}
    \item We revisit a largely unexamined premise behind LVM-based 
    time series forecasting and present the first systematic characterization 
    of the modality gap between rendered time series images and 
    natural images, identifying two distinct issues: a 
    \textbf{spectral} gap in data statistics and a \textbf{structural} gap in rendering time series images.
    
    \item 
    We propose \textbf{SSDA}, a dual-branch network with two complementary adapters: at the {data level}, the \textbf{SMA} reshapes the Fourier magnitude spectrum of rendered images while preserving its phase to retain temporal periodicity; at the {model level}, \textbf{SG-LoRA} injects temporal awareness into attention via low-rank adaptation to restore true continuity.

    \item Extensive experiments on seven real-world benchmarks demonstrate that SSDA achieves state-of-the-art performance in both full-shot and few-shot settings, with consistent improvements over 11 strong baselines spanning LVM-based, LLM-based, and classical methods.
\end{itemize}

\section{Related Work}
\subsection{LVM-based Time Series Forecasting}

The success of vision models has spurred their application in TSF. Early approaches applied CNNs such as ResNet~\cite{resnet2015} and VGG-19~\cite{vgg192015} to time series for feature extraction~\cite{Li2020}. More recently, large-scale vision models including ViT~\cite{vit2021}, MAE~\cite{mae2021}, and CLIP~\cite{clip2021} have been adapted for TSF~\cite{timevlm2025,visionts2025,vitime2025,zeng2024}. For example, VisionTS~\cite{visionts2025} applies a frozen MAE to rendered time series images, achieving competitive performance. Time-VLM~\cite{timevlm2025} models time series as images and text, encoding these visual and textual views via ViLT to enhance forecasting. DMMV~\cite{dmmv2025} leverages MAE's periodicity bias by decomposing series into seasonal and trend components within a multimodal fusion framework for temporal modeling. VisionTS++~\cite{visionts++} further introduces continual pre-training on large-scale time series, and proposes targeted strategies to mitigate cross-modal discrepancies, including data filtering, multivariate visual encoding, and quantile-based forecasting.
Despite promising results, these methods establish LVMs as effective forecasters, yet they focus on input-side designs such as rendering and fusion, rather than cross-modal adaptation of the LVMs to the time-series modality.

 \subsection{Cross-modal Adaptation for Time Series Forecasting}
Cross-modal adaptation reuses pre-trained models across modalities by aligning data distributions, enabling effective knowledge transfer~\cite{shen2023}. Following its success in vision~\cite{Verma2024,Lu2021PretrainedTA} and audio~\cite{ghosal2023,Hassid2023}, this paradigm has been extended to TSF along two routes. The language route reprograms continuous signals into discrete tokens for LLM processing, exemplified by Time-LLM~\cite{timellm2024}, TEST~\cite{test2024}, and TimeCMA~\cite{timecma2025}; although effective, such tokenization is inherently lossy and disrupts numerical continuity. The vision route is a more natural alternative, since continuous signals align with image patches far better than with discrete tokens, prompting a growing line of LVM adaptations~\cite{visionts2025,timevlm2025,dmmv2025}. However, current LVM adaptations rely on constrained strategies such as freezing the backbone, attaching lightweight adapters, or performing costly continual pre-training~\cite{visionts++}. These strategies adjust what the model sees but not how it reads patches; neither the spectral statistics of the rendered input nor the structural ordering of patches is aligned with the LVM's pre-training distribution, leaving both modality gaps unaddressed.

\begin{figure}[t]
    \centering
    \includegraphics[width=1\textwidth]{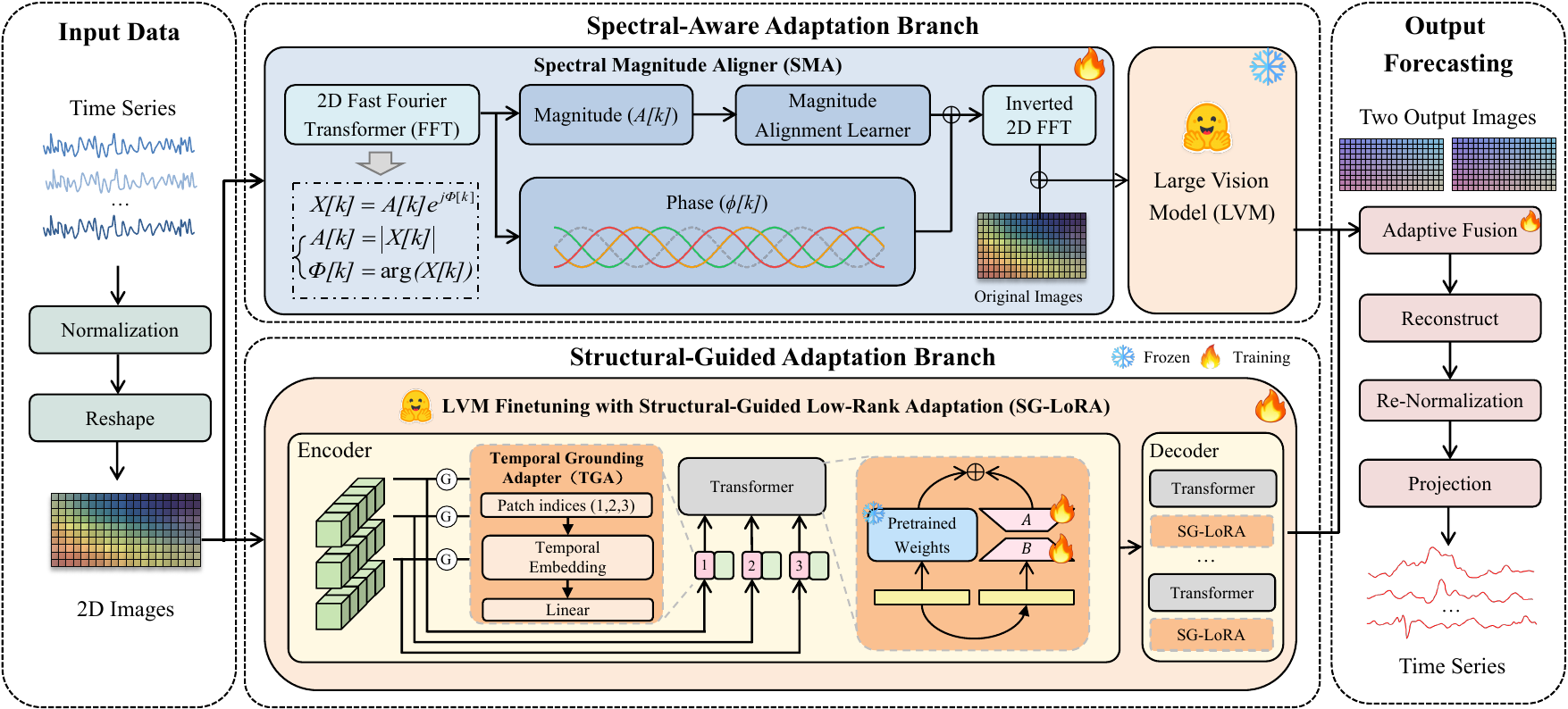}
    \caption{The Overall architecture of SSDA.}
    \label{fig:structure}
\end{figure}

\section{Methodology}

\subsection{Model Structure}

As shown in Figure~\ref{fig:structure}, SSDA processes a normalized, image-rendered time series through two parallel branches that each target one identified gap. The \textbf{spectral branch} closes the spectral gap at the \emph{data level}: a \textbf{Spectral Magnitude Aligner (SMA)} reshapes the frequency-domain statistics of the rendered image toward those of natural images before forwarding it to a frozen LVM. The \textbf{structural branch} closes the structural gap at the \emph{model level}: \textbf{Structural-Guided LoRA (SG-LoRA)}, comprising a \textbf{Temporal Grounding Adapter (TGA)} and low-rank attention adapters, restores true temporal ordering within a trainable LVM. The two branch outputs are combined by an \textbf{Adaptive Fusion} module, then reconstructed and denormalized to produce the final forecast.


\subsection{Spectral Magnitude Aligner}


\textbf{Spectral Magnitude Aligner (SMA)} decomposes the input image 
into magnitude and phase components. The magnitude is selectively enhanced by a learnable frequency enhancer to shift the spectral statistics of the rendered time series toward those of natural images, while the phase is kept unchanged to preserve temporal structure, following recent evidence that amplitude components yield more stable optimization dynamics~\cite{adamamba2026}. The enhanced spectrum is recombined and passed through inverse FFT before being forwarded to the frozen MAE encoder.

Concretely, given an input 2D time series image $\mathbf{I} \in \mathbb{R}^{B \times 1 \times H \times W}$, where $B$ is the batch size and $H, W$ are spatial dimensions, SMA first computes the 2D Real Fast Fourier Transform (RFFT):
\begin{equation}
\mathbf{F} = \mathcal{F}_\text{RFFT}(\mathbf{I}),
\end{equation}
where $\mathbf{F} \in \mathbb{C}^{B \times 1 \times H \times (W/2+1)}$ contains the complex frequency coefficients. The transform decomposes $\mathbf{F}$ into magnitude and phase components:
\begin{align}
\mathbf{A} = |\mathbf{F}|,\
\mathbf{\varphi} = \angle \mathbf{F},
\end{align}
with $\mathbf{A}, \mathbf{\varphi} \in \mathbb{R}^{B \times 1 \times H \times (W/2+1)}$ representing the magnitude and phase spectra, respectively.
The key innovation lies in selectively enhancing only the magnitude spectrum while preserving the phase information. A learnable convolutional network $\mathcal{E}_\theta$ processes the magnitude spectrum:
\begin{equation}
\mathbf{A}' = \mathcal{E}_\theta(\mathbf{A}),
\end{equation}
where $\mathcal{E}_\theta$ consists of two convolutional layers:
\begin{equation}
\mathcal{E}_\theta = \text{Conv}_{3\times3}(1,16) \rightarrow \text{BN} \rightarrow \text{ReLU} \rightarrow \text{Dropout}(0.1) \rightarrow \text{Conv}_{3\times3}(16,1).
\end{equation}
The enhanced frequency representation is reconstructed by combining the processed magnitude with the original phase:
\begin{equation}
\mathbf{F}' = \mathbf{A}' \odot e^{j\mathbf{\varphi}},
\end{equation}
where $\odot$ denotes element-wise multiplication. The enhanced spatial image is obtained via inverse RFFT:
\begin{equation}
\mathbf{I}_\text{enhanced} = \mathcal{F}_\text{IRFFT}(\mathbf{F}').
\end{equation}
To prevent over-enhancement and maintain training stability, a residual blending is applied:
\begin{equation}
\mathbf{I}_\text{sp} = \mathbf{I} + \lambda (\mathbf{I}_\text{enhanced} - \mathbf{I}),
\label{eq:residual}
\end{equation}
where $\lambda$ is a blending coefficient. We set $\lambda=0.05$, validated through ablation as the optimal trade-off: larger values risk over-enhancement and information loss, while smaller values yield negligible spectral correction.

The enhanced single-channel image is replicated across three channels to satisfy MAE's RGB input format:
\begin{equation}
\mathbf{I}^\text{3c}_\text{sp} = \text{repeat}(\mathbf{I}_\text{sp}, \text{channels}=3).
\end{equation}
The resulting $\mathbf{I}^\text{3c}_\text{sp} \in \mathbb{R}^{B \times 3 \times H \times W}$ is forwarded to the frozen MAE encoder to produce $y_{\text{sp}}$.

\subsection{Structural-Guided  Low-Rank Adaptation}

To close the structural gap between rendered time series images and pre-trained vision models, we propose \textbf{Structural-Guided Low-Rank Adaptation (SG-LoRA)}, a parameter-efficient fine-tuning strategy that explicitly injects temporal ordering into patch-level representations while preserving the spatial reasoning capacity of the pre-trained backbone.

\subsubsection{Temporal Grounding Adapter}

The core innovation of the structural branch is the \textbf{Temporal Grounding Adapter (TGA)}, which re-establishes the correspondence between spatial patch positions in the rendered image and their original temporal positions in the 1D series, thereby resolving the structural misalignment introduced by image reshaping.

Given the original rendered image $\mathbf{I} \in \mathbb{R}^{B \times 1 \times H \times W}$, we first divide it into non-overlapping patches following the standard ViT preprocessing. Each patch is then linearly projected through a learnable embedding layer to obtain the patch embeddings $\mathbf{X} \in \mathbb{R}^{B \times L \times D}$, where $L = \frac{H}{P} \times \frac{W}{P}$ denotes the number of image patches with patch size $P$. 
Based on these patch embeddings, the TGA module then generates the original temporal indices:
\begin{equation}
    \mathbf{T} = [0, 1, 2, \ldots, L-1] \in \mathbb{R}^{1 \times L},
\end{equation}
where each index corresponds to the original temporal position in the time series. Subsequently, temporal positional embeddings are generated via 1D sinusoidal position encoding:
\begin{equation}
    \mathbf{P}_{\mathrm{temp}}[i, 2k] = \sin\left(\frac{i}{\omega_k}\right), \quad \mathbf{P}_{\mathrm{temp}}[i, 2k+1] = \cos\left(\frac{i}{\omega_k}\right),
\end{equation}
where $\omega_k = 10000^{2k/D}$ is the frequency base and $k \in \{0, \ldots, D/2-1\}$. This sinusoidal encoding captures multi-frequency temporal patterns, consistent with standard position encoding in Transformers.
To enhance representational capacity and task adaptability, TGA introduces a learnable projection layer:
\begin{equation}
    \mathbf{P}'_{\mathrm{temp}} = \mathbf{W}_{\mathrm{proj}} \mathbf{P}_{\mathrm{temp}},
\end{equation}
where $\mathbf{W}_{\mathrm{proj}} \in \mathbb{R}^{D \times D}$ is a learnable linear projection matrix. Finally, temporal positional information is injected into the original patch embeddings via a learnable fusion gating mechanism:
\begin{equation}
    g = \sigma(w_{\mathrm{fusion}}), \quad \mathbf{X}_{\mathrm{enhanced}} = \mathbf{X} + g \cdot \mathbf{P}'_{\mathrm{temp}},
\end{equation}
where $w_{\mathrm{fusion}}$ is a learnable parameter, $\sigma(\cdot)$ is the sigmoid function, and $g \in [0, 1]$ adaptively controls the fusion strength. This design enables the model to dynamically balance original visual features with temporal structural information based on task requirements.

\subsubsection{Low-Rank Adaptation}

TGA is combined with Low-Rank Adaptation (LoRA) to form the complete SG-LoRA module. LoRA achieves parameter-efficient fine-tuning by imposing a low-rank decomposition on the query, key, and value projection matrices of the attention layers:
\begin{equation}
    \mathbf{W}' = \mathbf{W} + \frac{\alpha_{\text{lora}}}{r} \mathbf{B}\mathbf{A},
\end{equation}
where $\mathbf{W} \in \mathbb{R}^{D \times D}$ is the original weight matrix, $\mathbf{B} \in \mathbb{R}^{D \times r}$ and $\mathbf{A} \in \mathbb{R}^{r \times D}$ are the low-rank decomposition matrices, $r$ is the rank, and $\alpha_{\text{lora}}$ is the scaling factor.

The output of the MAE after SG-LoRA fine-tuning is the prediction $y_{\text{st}}$.

\subsection{Adaptive Fusion}
\label{sec:fusion}
The two branches address complementary aspects of the modality gap, but their relative importance varies across datasets and forecasting horizons. To adaptively balance their contributions, we introduce a learnable scalar $\beta \in \mathbb{R}$, initialized to $0.5$ and constrained to a valid mixing range:
\begin{equation}
    \beta = \text{clamp}\left(\beta,\; 0,\; 1\right).
\end{equation}
The final prediction is a convex combination of both branch outputs:
\begin{equation}
    \label{eq:fusion}
    \hat{y} = \beta \cdot y_{\text{st}} + (1 - \beta) \cdot y_{\text{sp}}.
\end{equation}
where $\hat{y}$ is the final output of the SSDA. A larger $\beta$ emphasizes temporal dependencies recovered by SG-LoRA, while a smaller $\beta$ relies more on the visual priors of the frozen LVM. We empirically observe that $\beta$ remains within $[0,1]$ throughout training and find this simple parameterization to be as effective as a sigmoid-based gate while avoiding saturation in extreme cases.















\section{Experiments}


\textbf{Datasets}.
We use seven public benchmarks commonly used in time‑series forecasting: ETT (Electricity Transformer Temperature)~\cite{informer2021}, including ETTh1, ETTh2, ETTm1 and ETTm2; Weather~\cite{autoformer2022}; Traffic~\cite{autoformer2022}; and Electricity~\cite{Trindade2015}. For the full-shot experiments, 
we use chronological splits of 60\%/20\%/20\% (train/val/test) for the ETT datasets, and chronological splits of 70\%/10\%/20\% for the remaining datasets. For the few-shot experiments, we use only 5\% or 10\% of the original training data as the training set, while the validation and test sets remain unchanged.


\textbf{Baselines}.
We benchmark SSDA against 11 baselines spanning three categories. LVM-based methods comprise VisionTS~\cite{visionts2025}, DMMV~\cite{dmmv2025}, and Time-VLM~\cite{timevlm2025}. LLM-based methods include Time-LLM~\cite{timellm2024}, TimeCMA~\cite{timecma2025}, and GPT4TS~\cite{zhou2023}. Classical forecasters cover PatchTST~\cite{patchtst2023}, FEDformer~\cite{fedformer2022}, Informer~\cite{informer2021}, TimesNet~\cite{timesnet2023}, and DLinear~\cite{dlinear2022}. Several baseline results are reproduced from~\cite{timevlm2025}.

\textbf{Metrics}. 
Forecast accuracy is evaluated using Mean Squared Error (MSE) and Mean Absolute Error (MAE), both computed on denormalized predictions following standard practice in the TSF field.

\subsection{Full-shot Forecasting}

We evaluate SSDA against a diverse set of baselines on seven real-world benchmarks under four forecasting horizons, with results summarized in Table~\ref{tab:main_results}. The following observations highlight the effectiveness of SSDA:
\textbf{(1) Consistent state-of-the-art performance.} SSDA achieves \textbf{48 first-place results} out of all metric-horizon combinations. This demonstrates that SSDA delivers not only strong average performance but also consistent superiority across diverse datasets and prediction horizons.
\textbf{(2) Significant gains over the strongest LVM-based baseline.} Compared with DMMV, the second-best model, SSDA reduces MSE and MAE significantly. These improvements are consistent across all datasets, indicating that explicitly addressing the spectral and structural modality gaps leads to substantial and reliable performance gains beyond what ad hoc adaptation strategies can achieve.
\textbf{(3) Clear advantage over LLM-based methods.} SSDA outperforms LLM-based forecasters (Time-LLM, GPT4TS and TimeCMA) by a wider margin, with the performance gap widening further on long-horizon settings. This suggests that visual representations are more transferable to time series than language representations, consistent with our spectral analysis showing that the modality gap between rendered time series images and natural images is substantially smaller than that between time series and text.
\textbf{(4) Superiority of LVM-based approaches over classical methods.} Across all datasets, the top-performing models are uniformly vision-based (SSDA, DMMV, VisionTS), while traditional deep forecasters (PatchTST, TimesNet, DLinear) lag considerably behind, especially on datasets with strong periodic structure. This observation underscores the value of leveraging the rich visual priors accumulated during large-scale pre-training on natural images.

\begin{table}[h]
\centering
\caption{\small Full TSF results with forecasting horizons $H \in \{96, 192, 336, 720\}$. 
\textcolor{red}{Red} indicates the best performance, and \textcolor{blue}{Blue} indicates the second best. ``OOM'' denotes the out-of-memory failure.}
\label{tab:main_results}
\resizebox{1\textwidth}{!}{
\begin{tabular}{ll|cc|cc|cc|cc|cc|cc|cc|cc|cc|cc|cc}
\toprule
\multicolumn{2}{c}{Models} &
\multicolumn{2}{c}{SSDA} &
\multicolumn{2}{c}{DMMV} &
\multicolumn{2}{c}{Time-VLM} &
\multicolumn{2}{c}{VisionTS} &
\multicolumn{2}{c}{Time-LLM} &
\multicolumn{2}{c}{TimeCMA} &
\multicolumn{2}{c}{GPT4TS} &
\multicolumn{2}{c}{PatchTST} &
\multicolumn{2}{c}{TimesNet} &
\multicolumn{2}{c}{FEDformer} &
\multicolumn{2}{c}{DLinear} \\
\cmidrule(lr){3-4}\cmidrule(lr){5-6}\cmidrule(lr){7-8}\cmidrule(lr){9-10}
\cmidrule(lr){11-12}\cmidrule(lr){13-14}\cmidrule(lr){15-16}
\cmidrule(lr){17-18}\cmidrule(lr){19-20}\cmidrule(lr){21-22}\cmidrule(lr){23-24}
\multicolumn{2}{c}{Metrics} & MSE & MAE & MSE & MAE & MSE & MAE & MSE & MAE & MSE & MAE & MSE & MAE & MSE & MAE & MSE & MAE & MSE & MAE & MSE & MAE & MSE & MAE \\
\midrule

\multirow{5}{*}{\rotatebox{90}{ETTh1}}
& 96  & \textcolor{red}{0.343} & \textcolor{red}{0.372} & \textcolor{blue}{0.354} & 0.389 & 0.361 & \textcolor{blue}{0.386} & 0.355 & \textcolor{blue}{0.386} & 0.362 & 0.392 & 0.373 & 0.391 & 0.376 & 0.397 & 0.370 & 0.399 & 0.384 & 0.402 & 0.376 & 0.419 & 0.375 & 0.399 \\
& 192 & \textcolor{red}{0.387} & \textcolor{red}{0.402} & \textcolor{blue}{0.393} & \textcolor{blue}{0.405} & 0.397 & 0.415 & 0.395 & 0.407 & 0.398 & 0.418 & 0.427 & 0.421 & 0.416 & 0.418 & 0.413 & 0.421 & 0.436 & 0.429 & 0.420 & 0.448 & 0.405 & 0.416 \\
& 336 & \textcolor{blue}{0.409} & \textcolor{blue}{0.430} & \textcolor{red}{0.387} & \textcolor{red}{0.413} & 0.420 & 0.421 & 0.419 & 0.421 & 0.430 & 0.427 & 0.458 & 0.448 & 0.442 & 0.433 & 0.422 & 0.436 & 0.491 & 0.469 & 0.459 & 0.465 & 0.439 & 0.443 \\
& 720 & \textcolor{red}{0.408} & \textcolor{red}{0.430} & 0.445 & \textcolor{blue}{0.450} & \textcolor{blue}{0.441} & 0.458 & 0.458 & 0.460 & 0.442 & 0.457 & 0.449 & 0.460 & 0.477 & 0.456 & 0.447 & 0.466 & 0.521 & 0.500 & 0.506 & 0.507 & 0.472 & 0.490 \\
& Avg & \textcolor{red}{0.387} & \textcolor{red}{0.406} & \textcolor{blue}{0.395} & \textcolor{blue}{0.414} & 0.405 & 0.420 & 0.407 & 0.419 & 0.408 & 0.423 & 0.423 & 0.431 & 0.465 & 0.455 & 0.413 & 0.430 & 0.458 & 0.450 & 0.440 & 0.460 & 0.422 & 0.437 \\
\midrule

\multirow{5}{*}{\rotatebox{90}{ETTh2}}
& 96  & \textcolor{red}{0.267} & \textcolor{red}{0.328} & 0.294 & 0.349 & \textcolor{red}{0.267} & 0.335 & 0.288 & \textcolor{blue}{0.334} & 0.286 & 0.346 & 0.286 & 0.336 & 0.285 & 0.342 & \textcolor{blue}{0.274} & 0.336 & 0.340 & 0.374 & 0.358 & 0.397 & 0.289 & 0.353 \\
& 192 & \textcolor{red}{0.324} & \textcolor{red}{0.368} & 0.339 & 0.395 & \textcolor{blue}{0.326} & \textcolor{blue}{0.373} & 0.349 & 0.380 & 0.329 & 0.375 & 0.363 & 0.387 & 0.354 & 0.389 & 0.339 & 0.379 & 0.402 & 0.414 & 0.429 & 0.439 & 0.383 & 0.418 \\
& 336 & 0.347 & 0.392 & \textcolor{red}{0.322} & \textcolor{blue}{0.384} & 0.357 & 0.406 & 0.364 & 0.398 & 0.368 & 0.409 & 0.406 & 0.421 & 0.373 & 0.407 & \textcolor{blue}{0.329} & \textcolor{red}{0.380} & 0.452 & 0.452 & 0.496 & 0.487 & 0.448 & 0.465 \\
& 720 & \textcolor{red}{0.370} & \textcolor{red}{0.415} & 0.392 & 0.425 & 0.412 & 0.449 & 0.403 & 0.431 & 0.405 & 0.434 & 0.417 & 0.438 & 0.406 & 0.441 & \textcolor{blue}{0.379} & \textcolor{blue}{0.422} & 0.462 & 0.468 & 0.463 & 0.474 & 0.605 & 0.551 \\
& Avg & \textcolor{red}{0.327} & \textcolor{red}{0.376} & 0.337 & 0.388 & 0.341 & 0.391 & 0.351 & 0.386 & 0.334 & 0.383 & 0.372 & 0.397 & 0.381 & 0.412 & \textcolor{blue}{0.330} & \textcolor{blue}{0.379} & 0.414 & 0.427 & 0.437 & 0.449 & 0.431 & 0.446 \\
\midrule

\multirow{5}{*}{\rotatebox{90}{ETTm1}}
& 96  & \textcolor{blue}{0.283} & \textcolor{red}{0.328} & \textcolor{red}{0.279} & \textcolor{blue}{0.329} & 0.304 & 0.346 & 0.284 & 0.332 & 0.291 & 0.341 & 0.312 & 0.351 & 0.292 & 0.346 & 0.290 & 0.342 & 0.338 & 0.375 & 0.379 & 0.419 & 0.299 & 0.343 \\
& 192 & \textcolor{red}{0.316} & \textcolor{red}{0.353} & \textcolor{blue}{0.317} & \textcolor{blue}{0.357} & 0.332 & 0.366 & 0.327 & 0.362 & 0.341 & 0.369 & 0.361 & 0.378 & 0.332 & 0.372 & 0.332 & 0.369 & 0.374 & 0.387 & 0.426 & 0.441 & 0.335 & 0.365 \\
& 336 & \textcolor{red}{0.340} & \textcolor{red}{0.370} & \textcolor{blue}{0.351} & \textcolor{blue}{0.381} & 0.364 & 0.383 & 0.354 & 0.382 & 0.359 & 0.379 & 0.392 & 0.401 & 0.366 & 0.394 & 0.366 & 0.392 & 0.410 & 0.411 & 0.445 & 0.459 & 0.369 & 0.386 \\
& 720 & \textcolor{blue}{0.388} & \textcolor{red}{0.400} & 0.411 & 0.415 & \textcolor{red}{0.364} & \textcolor{blue}{0.410} & 0.411 & 0.415 & 0.433 & 0.419 & 0.453 & 0.438 & 0.417 & 0.421 & 0.416 & 0.420 & 0.478 & 0.450 & 0.543 & 0.490 & 0.425 & 0.421 \\
& Avg & \textcolor{red}{0.332} & \textcolor{red}{0.363} & \textcolor{blue}{0.340} & \textcolor{blue}{0.371} & 0.351 & 0.376 & 0.344 & 0.373 & 0.356 & 0.377 & 0.380 & 0.392 & 0.388 & 0.403 & 0.351 & 0.380 & 0.400 & 0.406 & 0.448 & 0.452 & 0.357 & 0.378 \\
\midrule

\multirow{5}{*}{\rotatebox{90}{ETTm2}}
& 96  & 0.162 & \textcolor{red}{0.249} & 0.172 & 0.260 & \textcolor{red}{0.160} & \textcolor{blue}{0.252} & 0.174 & 0.262 & \textcolor{blue}{0.161} & 0.253 & 0.173 & 0.258 & 0.173 & 0.262 & 0.165 & 0.255 & 0.187 & 0.267 & 0.203 & 0.287 & 0.167 & 0.269 \\
& 192 & \textcolor{red}{0.217} & \textcolor{red}{0.290} & 0.227 & 0.298 & 0.220 & \textcolor{blue}{0.291} & 0.228 & 0.297 & \textcolor{blue}{0.219} & 0.293 & 0.238 & 0.301 & 0.229 & 0.301 & 0.220 & 0.292 & 0.249 & 0.309 & 0.269 & 0.328 & 0.224 & 0.303 \\
& 336 & \textcolor{red}{0.268} & \textcolor{red}{0.323} & 0.272 & \textcolor{blue}{0.327} & \textcolor{blue}{0.270} & 0.325 & 0.281 & 0.337 & 0.271 & 0.329 & 0.297 & 0.338 & 0.286 & 0.341 & 0.274 & 0.329 & 0.321 & 0.351 & 0.325 & 0.366 & 0.281 & 0.342 \\
& 720 & \textcolor{red}{0.341} & \textcolor{red}{0.375} & 0.351 & 0.381 & \textcolor{blue}{0.348} & \textcolor{blue}{0.378} & 0.384 & 0.410 & 0.352 & 0.379 & 0.393 & 0.394 & 0.378 & 0.401 & 0.362 & 0.385 & 0.408 & 0.403 & 0.421 & 0.415 & 0.397 & 0.421 \\
& Avg & \textcolor{red}{0.249} & \textcolor{red}{0.311} & 0.256 & 0.317 & \textcolor{blue}{0.250} & \textcolor{blue}{0.312} & 0.267 & 0.327 & 0.251 & 0.313 & 0.275 & 0.323 & 0.284 & 0.339 & 0.255 & 0.315 & 0.291 & 0.333 & 0.305 & 0.349 & 0.267 & 0.333 \\
\midrule

\multirow{5}{*}{\rotatebox{90}{Weather}}
& 96  & \textcolor{red}{0.139} & \textcolor{red}{0.193} & \textcolor{blue}{0.143} & 0.195 & 0.148 & 0.200 & 0.146 & \textcolor{blue}{0.194} & 0.147 & 0.201 & 0.167 & 0.211 & 0.162 & 0.212 & 0.149 & 0.198 & 0.172 & 0.220 & 0.217 & 0.296 & 0.176 & 0.237 \\
& 192 & \textcolor{red}{0.185} & \textcolor{blue}{0.236} & \textcolor{blue}{0.187} & 0.242 & 0.193 & 0.240 & 0.194 & 0.238 & 0.189 & \textcolor{red}{0.234} & 0.212 & 0.253 & 0.204 & 0.248 & 0.194 & 0.241 & 0.219 & 0.261 & 0.276 & 0.336 & 0.220 & 0.282 \\
& 336 & \textcolor{red}{0.230} & \textcolor{red}{0.272} & \textcolor{blue}{0.237} & \textcolor{blue}{0.273} & 0.243 & 0.281 & 0.243 & 0.275 & 0.262 & 0.279 & 0.270 & 0.292 & 0.254 & 0.286 & 0.245 & 0.282 & 0.280 & 0.306 & 0.339 & 0.380 & 0.265 & 0.319 \\
& 720 & 0.314 & 0.329 & \textcolor{red}{0.302} & \textcolor{red}{0.315} & 0.312 & 0.332 & 0.318 & 0.328 & \textcolor{blue}{0.304} & \textcolor{blue}{0.316} & 0.350 & 0.348 & 0.326 & 0.337 & 0.314 & 0.334 & 0.365 & 0.359 & 0.403 & 0.428 & 0.333 & 0.362 \\
& Avg & \textcolor{red}{0.217} & 0.258 & \textcolor{red}{0.217} & \textcolor{red}{0.256}& \textcolor{blue}{0.224} & 0.263 & 0.225 & 0.258 & 0.225 & \textcolor{blue}{0.257} & 0.250 & 0.276 & 0.237 & 0.270 & 0.225 & 0.264 & 0.259 & 0.287 & 0.309 & 0.360 & 0.248 & 0.300 \\
\midrule

\multirow{5}{*}{\rotatebox{90}{Electricity}}
& 96  & \textcolor{red}{0.126} & 0.219 & \textcolor{red}{0.126} & \textcolor{red}{0.213} & 0.142 & 0.245 & \textcolor{blue}{0.127} & \textcolor{blue}{0.217} & 0.131 & 0.224 & 0.143 & 0.238 & 0.139 & 0.238 & 0.129 & 0.222 & 0.168 & 0.272 & 0.193 & 0.308 & 0.140 & 0.237 \\
& 192 & \textcolor{red}{0.145} & \textcolor{blue}{0.238} & \textcolor{red}{0.145} & \textcolor{red}{0.237} & 0.157 & 0.260 & \textcolor{blue}{0.148} & \textcolor{red}{0.237} & 0.152 & 0.241 & 0.161 & 0.259 & 0.153 & 0.251 & 0.157 & 0.240 & 0.184 & 0.289 & 0.201 & 0.315 & 0.153 & 0.249 \\
& 336 & \textcolor{red}{0.160} & 0.255 & \textcolor{blue}{0.162} & \textcolor{blue}{0.254} & 0.174 & 0.276 & 0.163 & \textcolor{red}{0.253} & 0.160 & 0.248 & 0.169 & 0.261 & 0.169 & 0.266 & 0.163 & 0.259 & 0.198 & 0.300 & 0.214 & 0.329 & 0.169 & 0.267 \\
& 720 & \textcolor{red}{0.194} & \textcolor{blue}{0.288} & \textcolor{blue}{0.197} & \textcolor{red}{0.286} & 0.214 & 0.308 & 0.199 & 0.293 & 0.192 & 0.298 & 0.219 & 0.315 & 0.206 & 0.297 & \textcolor{blue}{0.197} & 0.290 & 0.220 & 0.320 & 0.246 & 0.355 & 0.203 & 0.301 \\
& Avg & \textcolor{red}{0.156} & \textcolor{blue}{0.250} & \textcolor{blue}{0.158} & \textcolor{red}{0.248} & 0.172 & 0.272 & 0.159 & \textcolor{blue}{0.250} & 0.158 & 0.252 & 0.174 & 0.269 & 0.167 & 0.263 & 0.161 & 0.252 & 0.192 & 0.295 & 0.214 & 0.327 & 0.166 & 0.263 \\
\midrule

\multirow{5}{*}{\rotatebox{90}{Traffic}}
& 96  & \textcolor{blue}{0.345} & 0.243 & \textcolor{red}{0.344} & \textcolor{blue}{0.237} & 0.393 & 0.290 & 0.346 & \textcolor{red}{0.234} & 0.362 & 0.248 & OOM & OOM & 0.388 & 0.282 & 0.360 & 0.249 & 0.593 & 0.321 & 0.587 & 0.366 & 0.410 & 0.282 \\
& 192 & \textcolor{red}{0.359} & \textcolor{blue}{0.247} & \textcolor{blue}{0.363} & 0.249 & 0.405 & 0.296 & 0.376 & \textcolor{red}{0.245} & 0.374 & 0.247 & OOM & OOM & 0.407 & 0.290 & 0.379 & 0.256 & 0.617 & 0.336 & 0.604 & 0.373 & 0.423 & 0.287 \\
& 336 & 0.395 & 0.267 & \textcolor{red}{0.387} & \textcolor{blue}{0.256} & 0.420 & 0.305 & \textcolor{blue}{0.389} & \textcolor{red}{0.252} & 0.385 & 0.271 & OOM & OOM & 0.412 & 0.294 & 0.392 & 0.264 & 0.629 & 0.336 & 0.621 & 0.383 & 0.436 & 0.296 \\
& 720 & \textcolor{red}{0.427} & \textcolor{red}{0.282} & 0.433 & \textcolor{blue}{0.284} & 0.459 & 0.323 & 0.432 & 0.293 & \textcolor{blue}{0.430} & 0.288 & OOM & OOM & 0.450 & 0.312 & 0.432 & 0.286 & 0.640 & 0.350 & 0.626 & 0.382 & 0.466 & 0.315 \\
& Avg & \textcolor{red}{0.381} & 0.261 & \textcolor{blue}{0.382} & \textcolor{blue}{0.257} & 0.419 & 0.304 & 0.386 & \textcolor{red}{0.256} & 0.388 & 0.264 & OOM & OOM & 0.414 & 0.294 & 0.390 & 0.263 & 0.620 & 0.336 & 0.610 & 0.376 & 0.433 & 0.295 \\
\midrule

\multicolumn{2}{c|}{\textbf{1st count}} 
& \multicolumn{2}{c|}{\textbf{48}} 
& \multicolumn{2}{c|}{16} 
& \multicolumn{2}{c|}{3} 
& \multicolumn{2}{c|}{6} 
& \multicolumn{2}{c|}{1} 
& \multicolumn{2}{c|}{0} 
& \multicolumn{2}{c|}{0} 
& \multicolumn{2}{c|}{1} 
& \multicolumn{2}{c|}{0} 
& \multicolumn{2}{c|}{0} 
& \multicolumn{2}{c}{0} \\

\bottomrule
\end{tabular}}
\end{table}

\subsection{Few-shot Forecasting}

Table~\ref{tab:few_shot_full} reports few-shot results using 10\% and 5\% training data. SSDA achieves \textbf{9} and \textbf{8} first-place results under the two ratios, respectively, consistently outperforming competing methods across the majority of datasets. In contrast, traditional deep models degrade sharply under data scarcity, while pre-trained model-based methods remain relatively stable, underscoring the practical value of transferable LVM representations in low-data regimes. Note that DMMV is excluded from the few-shot comparison as its official implementation does not support the few-shot training protocol used in this evaluation. Full per-horizon results are provided in Appendix~\ref{apd:few_shot}.

\begin{table*}[t]
\centering
\caption{\small Few-shot forecasting results on 10\% and 5\% training data. 
}
\label{tab:few_shot_full}
\footnotesize
\setlength{\tabcolsep}{3pt}
\renewcommand{\arraystretch}{1.05}

\resizebox{\textwidth}{!}{
\begin{tabular}{c|c|cc|cc|cc|cc|cc|cc|cc|cc|cc|cc}
\toprule
Ratio & Dataset
& \multicolumn{2}{c|}{SSDA}
& \multicolumn{2}{c|}{Time-VLM}
& \multicolumn{2}{c|}{VisionTS}
& \multicolumn{2}{c|}{Time-LLM}
& \multicolumn{2}{c|}{GPT4TS}
& \multicolumn{2}{c|}{PatchTST}
& \multicolumn{2}{c|}{TimesNet}
& \multicolumn{2}{c|}{FEDformer}
& \multicolumn{2}{c|}{Informer}
& \multicolumn{2}{c}{DLinear} \\

\cmidrule(lr){3-22}

& Metric
& MSE& MAE& MSE& MAE& MSE& MAE& MSE& MAE& MSE& MAE
& MSE& MAE& MSE& MAE& MSE& MAE& MSE& MAE& MSE& MAE \\

\midrule
\multirow{7}{*}{10\%}
& ETTh1 & \textcolor{red}{0.396}&\textcolor{red}{0.410} &0.431&0.442 &\textcolor{blue}{0.409}&\textcolor{blue}{0.412} &0.556&0.522 &0.590&0.525 &0.633&0.542 &0.869&0.628 &0.639&0.561 &1.199&0.809 &0.691&0.600 \\
& ETTh2 & \textcolor{red}{0.324}&\textcolor{red}{0.373} &0.361&0.405 &\textcolor{blue}{0.353}&\textcolor{blue}{0.389} &0.370&0.394 &0.397&0.421 &0.415&0.431 &0.479&0.465 &0.466&0.475 &3.872&1.513 &0.605&0.538 \\
& ETTm1 & \textcolor{red}{0.360}&\textcolor{red}{0.369} &\textcolor{blue}{0.360}&\textcolor{blue}{0.382} &0.595&0.508 &0.404&0.427 &0.464&0.441 &0.501&0.466 &0.677&0.537 &0.722&0.605 &1.192&0.821 &0.411&0.429 \\
& ETTm2 & \textcolor{blue}{0.270}&\textcolor{red}{0.320} &\textcolor{red}{0.263}&\textcolor{blue}{0.323} &0.292&0.342 &0.277&\textcolor{blue}{0.323} &0.293&0.335 &0.296&0.343 &0.320&0.353 &0.463&0.488 &3.370&1.440 &0.316&0.368 \\
& Weather & \textcolor{red}{0.230}&\textcolor{red}{0.271} &0.245&0.282 &0.247&0.285 &\textcolor{blue}{0.234}&\textcolor{blue}{0.273} &0.238&0.275 &0.242&0.279 &0.279&0.301 &0.284&0.324 &0.597&0.495 &0.241&0.283 \\
& Electricity & 0.199&0.290 &0.198&0.291 &0.180&\textcolor{red}{0.265} &\textcolor{red}{0.175}&0.270 &\textcolor{blue}{0.176}&\textcolor{blue}{0.269} &0.180&0.273 &0.323&0.392 &0.346&0.427 &1.195&0.891 &0.180&0.280 \\
& Traffic & 0.468&0.310 &0.484&0.357 &0.483&0.344 &\textcolor{red}{0.429}&\textcolor{red}{0.306} &0.440&\textcolor{blue}{0.310} &\textcolor{blue}{0.430}&0.305 &0.951&0.535 &0.663&0.425 &1.534&0.811 &0.447&0.313 \\
\midrule
\multicolumn{2}{c|}{\textbf{1st count}} & \multicolumn{2}{c|}{\textbf{9}} & \multicolumn{2}{c|}{1} & \multicolumn{2}{c|}{1} & \multicolumn{2}{c|}{3} & \multicolumn{2}{c|}{0} & \multicolumn{2}{c|}{0} & \multicolumn{2}{c|}{0} & \multicolumn{2}{c|}{0} & \multicolumn{2}{c|}{0} & \multicolumn{2}{c}{0} \\

\midrule
\multirow{7}{*}{5\%}
& ETTh1 & \textcolor{red}{0.390}&\textcolor{red}{0.401} &0.442&0.453 &\textcolor{blue}{0.409}&\textcolor{blue}{0.403} &0.627&0.543 &0.681&0.560 &0.694&0.569 &0.925&0.647 &0.658&0.562 &1.225&0.817 &0.750&0.611 \\
& ETTh2 & \textcolor{red}{0.311}&\textcolor{red}{0.358} &0.354&0.402 &\textcolor{blue}{0.348}&\textcolor{blue}{0.376} &0.382&0.418 &0.400&0.433 &0.827&0.615 &0.439&0.448 &0.463&0.454 &3.922&1.653 &0.694&0.577 \\
& ETTm1 & \textcolor{red}{0.363}&\textcolor{red}{0.369} &\textcolor{blue}{0.364}&\textcolor{blue}{0.385} &0.574&0.498 &0.425&0.434 &0.472&0.450 &0.526&0.476 &0.717&0.561 &0.730&0.592 &1.163&0.791 &0.400&0.417 \\
& ETTm2 & 0.281&\textcolor{blue}{0.326} &\textcolor{red}{0.262}&\textcolor{red}{0.323} &0.308&0.350 &\textcolor{blue}{0.274}&0.323 &0.308&0.346 &0.314&0.352 &0.344&0.372 &0.381&0.404 &3.658&1.489 &0.399&0.426 \\
& Weather & \textcolor{red}{0.244}&\textcolor{red}{0.275} &\textcolor{blue}{0.246}&\textcolor{blue}{0.284} &0.255&0.296 &0.260&0.309 &0.263&0.301 &0.269&0.303 &0.298&0.318 &0.309&0.353 &0.584&0.527 &0.263&0.308 \\
& Electricity & 0.191&0.276 &0.218&0.315 &0.183&\textcolor{blue}{0.269} &0.179&\textcolor{red}{0.268} &\textcolor{blue}{0.178}&0.273 &0.181&0.277 &0.402&0.453 &0.266&0.353 &1.281&0.929 &\textcolor{red}{0.176}&0.275 \\
& Traffic & 0.493&0.330 &0.558&0.410 &0.715&0.469 &\textcolor{blue}{0.423}&\textcolor{blue}{0.298} &0.434&0.305 &\textcolor{red}{0.418}&\textcolor{red}{0.296} &0.867&0.493 &0.676&0.423 &1.591&0.832 &0.450&0.317 \\
\midrule
\multicolumn{2}{c|}{\textbf{1st count}} & \multicolumn{2}{c|}{\textbf{8}} & \multicolumn{2}{c|}{2} & \multicolumn{2}{c|}{0} & \multicolumn{2}{c|}{1} & \multicolumn{2}{c|}{0} & \multicolumn{2}{c|}{2} & \multicolumn{2}{c|}{0} & \multicolumn{2}{c|}{0} & \multicolumn{2}{c|}{0} & \multicolumn{2}{c}{1} \\

\bottomrule
\end{tabular}}
\end{table*}

\subsection{Ablation Study}
Table~\ref{tab:ablation_improve} reports ablation results on four ETT datasets, and full results are in Appendix~\ref{apd:ablation}. We study three ablation variants:
(1) \textbf{W/O Spectral branch}: removes SMA, retaining only SG-LoRA;
(2) \textbf{W/O Structural branch}: removes SG-LoRA, retaining only SMA;
(3) \textbf{W/O Spectral branch \& TGA}: further removes TGA from the structural branch, reducing it to vanilla LoRA fine-tuning.

\textbf{Both branches are necessary.} Removing either branch consistently degrades performance across all datasets, confirming that the spectral and structural gaps are complementary and must be addressed jointly. The structural branch contributes more since removing it (Variant 1) causes larger drops than removing the spectral branch (Variant 2), particularly on the minute-level datasets (ETTm1: $-9.04\%$, ETTm2: $-15.70\%$ in MSE).

\textbf{Vanilla LoRA alone is insufficient.} Variant 3 further isolates the role of TGA by reducing SG-LoRA to plain LoRA, and causes the largest degradation among all variants ($-14.01\%$ MSE / $-5.86\%$ MAE on average). This confirms that naive LoRA fine-tuning, without explicit temporal structure injection through TGA, fails to bridge the gap between visual pretraining and time series forecasting, and that TGA is the key driver enabling LoRA to transfer effectively.

\begin{table}[h]
\centering
\caption{Ablation study with relative change (\%). All degradations are highlighted in \textcolor{red}{red}.}
\label{tab:ablation_improve}
\footnotesize
\setlength{\tabcolsep}{3pt}
\renewcommand{\arraystretch}{1.0}
\resizebox{1\textwidth}{!}{
\begin{tabular}{ccccccccccc}
\toprule
Dataset 
& \multicolumn{2}{c}{ETTh1} 
& \multicolumn{2}{c}{ETTh2} 
& \multicolumn{2}{c}{ETTm1} 
& \multicolumn{2}{c}{ETTm2} 
& \multicolumn{2}{c}{Avg} \\
\cmidrule(lr){2-3} \cmidrule(lr){4-5} \cmidrule(lr){6-7} \cmidrule(lr){8-9} \cmidrule(lr){10-11}
Method($\downarrow$), Metric($\rightarrow$)
& MSE & MAE 
& MSE & MAE 
& MSE & MAE 
& MSE & MAE 
& MSE & MAE \\
\midrule
SSDA (full)
& \textbf{0.387} & \textbf{0.406} & \textbf{0.327} & \textbf{0.376} & \textbf{0.332} & \textbf{0.363} & \textbf{0.249} & \textbf{0.311} & \textbf{0.324} & \textbf{0.364} \\
\midrule
(1) W/O Spectral branch 
& 0.403 & 0.417 & 0.332 & 0.378 & 0.362 & 0.374 & 0.288 & 0.331 & 0.346 & 0.375 \\
\quad Drop
& \textcolor{red}{$-4.13\%$} & \textcolor{red}{$-2.71\%$} 
& \textcolor{red}{$-1.53\%$} & \textcolor{red}{$-0.53\%$} 
& \textcolor{red}{$-9.04\%$} & \textcolor{red}{$-3.03\%$} 
& \textcolor{red}{$-15.70\%$} & \textcolor{red}{$-6.43\%$}
& \textcolor{red}{$-7.60\%$} & \textcolor{red}{$-3.18\%$} \\
\midrule
(2) W/O Structural branch 
& 0.391 & 0.408 & 0.332 & 0.378 & 0.342 & 0.367 & 0.260 & 0.318 & 0.331 & 0.368 \\
\quad Drop
& \textcolor{red}{$-1.03\%$} & \textcolor{red}{$-0.49\%$} 
& \textcolor{red}{$-1.53\%$} & \textcolor{red}{$-0.53\%$} 
& \textcolor{red}{$-3.01\%$} & \textcolor{red}{$-1.10\%$} 
& \textcolor{red}{$-4.42\%$} & \textcolor{red}{$-2.25\%$}
& \textcolor{red}{$-2.50\%$} & \textcolor{red}{$-1.09\%$} \\
\midrule
(3) W/O Spectral branch \& TGA 
& 0.462 & 0.438 & 0.396 & 0.410 & 0.353 & 0.369 & 0.272 & 0.323 & 0.371 & 0.385 \\
\quad (Only LoRA) Drop
& \textcolor{red}{$-19.38\%$} & \textcolor{red}{$-7.88\%$} 
& \textcolor{red}{$-21.10\%$} & \textcolor{red}{$-9.04\%$} 
& \textcolor{red}{$-6.33\%$} & \textcolor{red}{$-1.65\%$} 
& \textcolor{red}{$-9.24\%$} & \textcolor{red}{$-4.86\%$}
& \textcolor{red}{$-14.01\%$} & \textcolor{red}{$-5.86\%$} \\
\bottomrule
\end{tabular}}
\end{table}

\begin{figure}[b]
\centering
\begin{minipage}{0.24\textwidth}
    \centering
    \includegraphics[width=\linewidth]{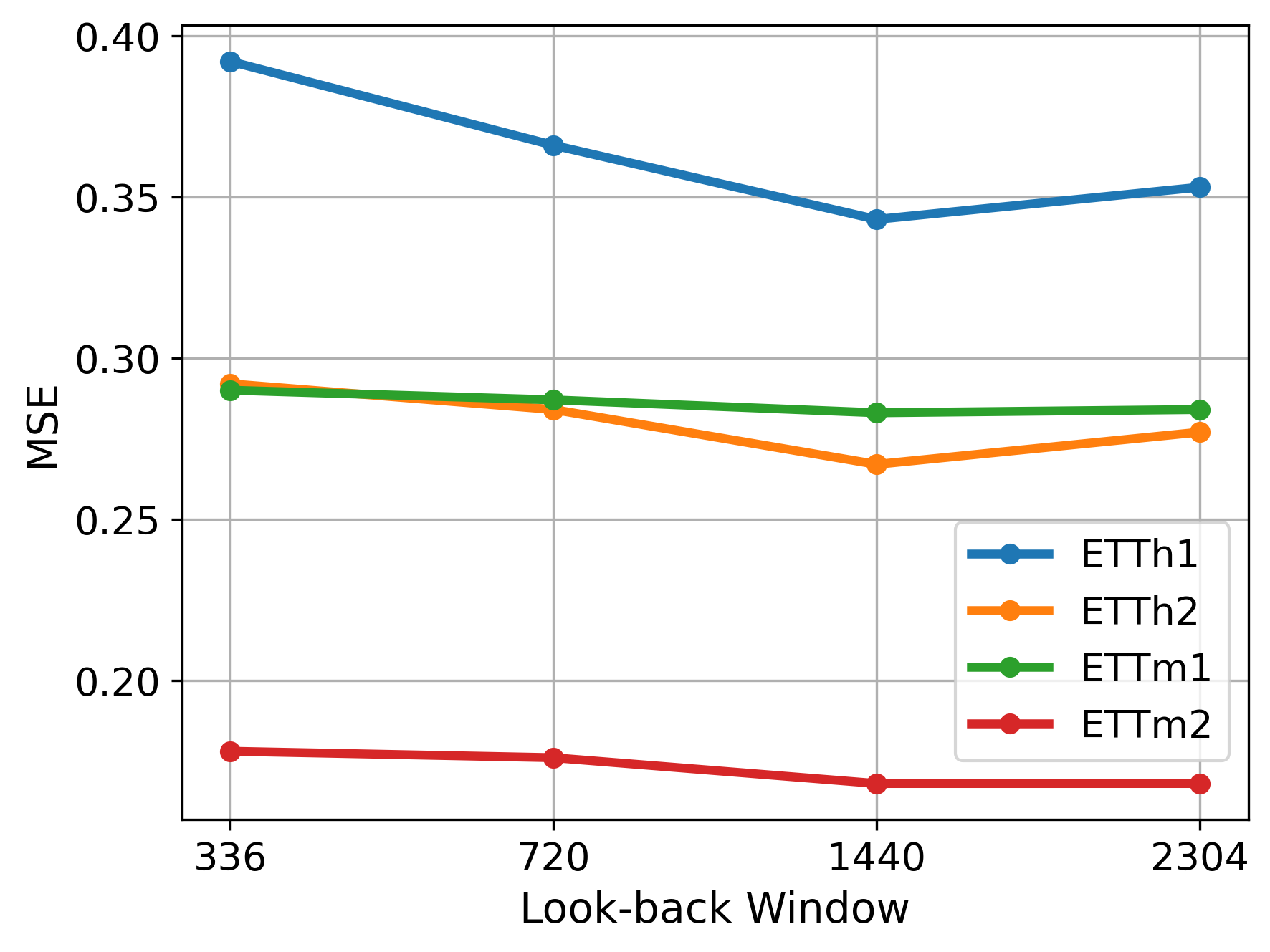}
    \small (a) Look-back window
\end{minipage}
\hfill
\begin{minipage}{0.24\textwidth}
    \centering
    \includegraphics[width=\linewidth]{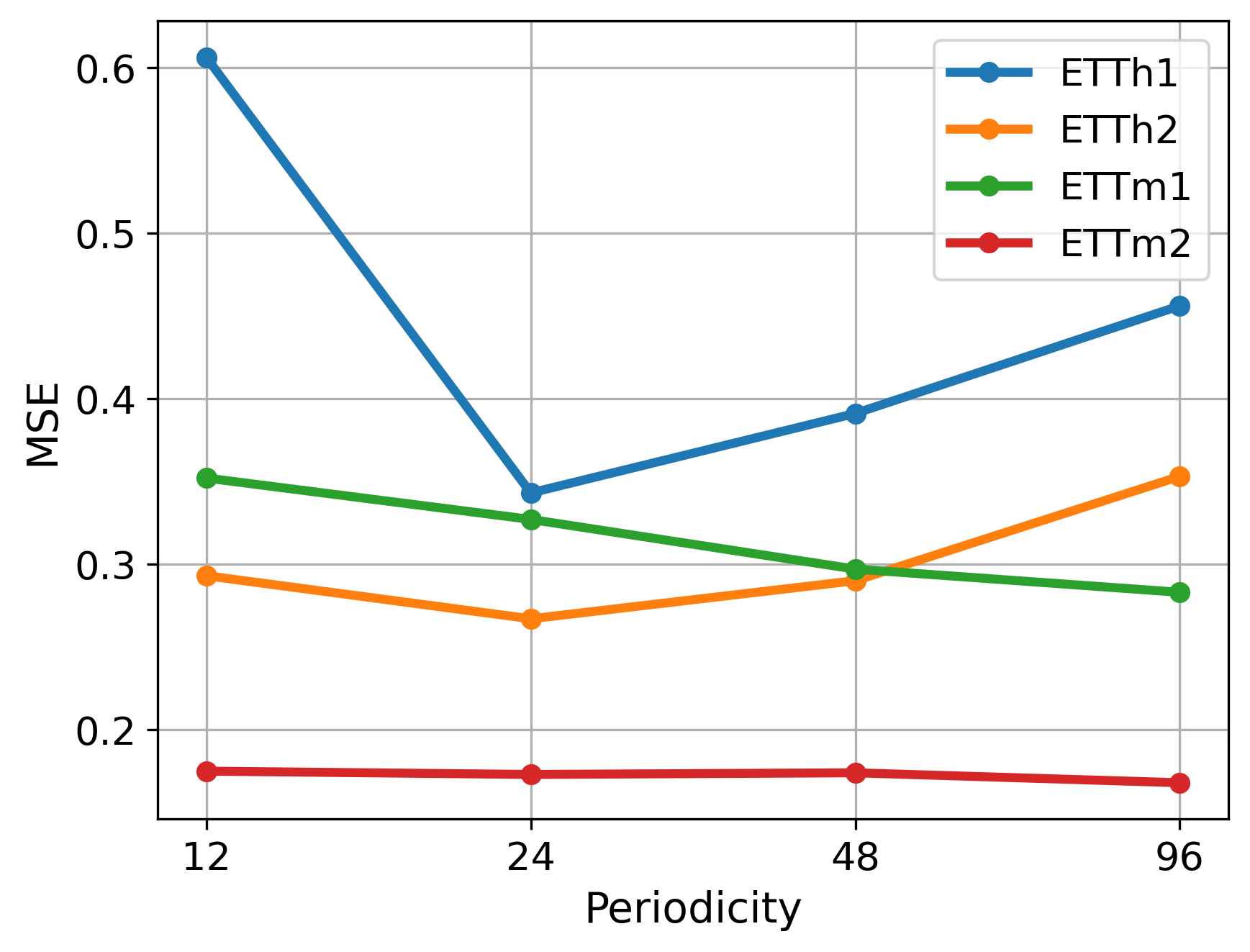}
    \small (b) Periodicity
\end{minipage}
\hfill
\begin{minipage}{0.24\textwidth}
    \centering
    \includegraphics[width=\linewidth]{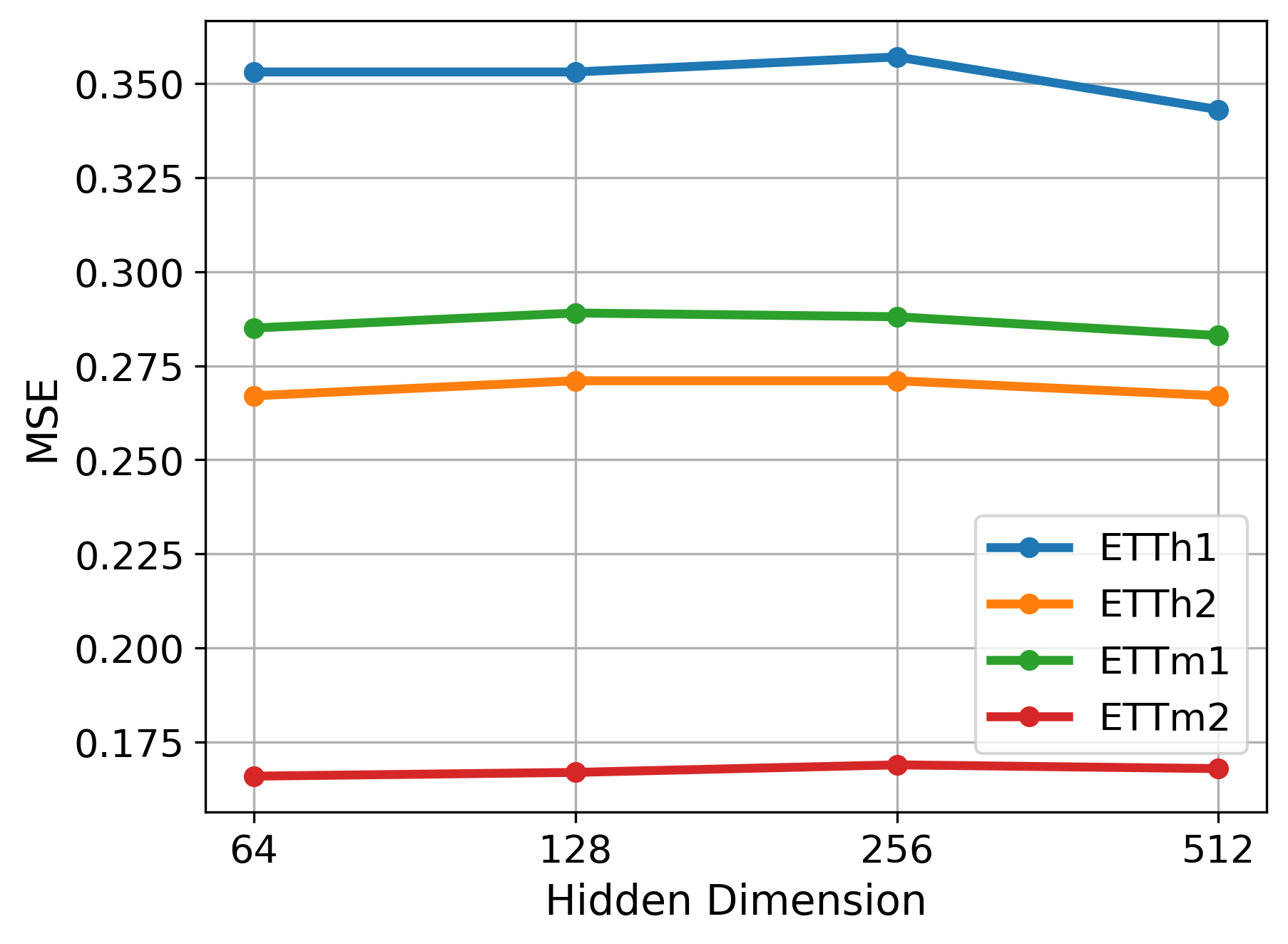}
    \small (c) Hidden dimension
\end{minipage}
\hfill
\begin{minipage}{0.24\textwidth}
    \centering
    \includegraphics[width=\linewidth]{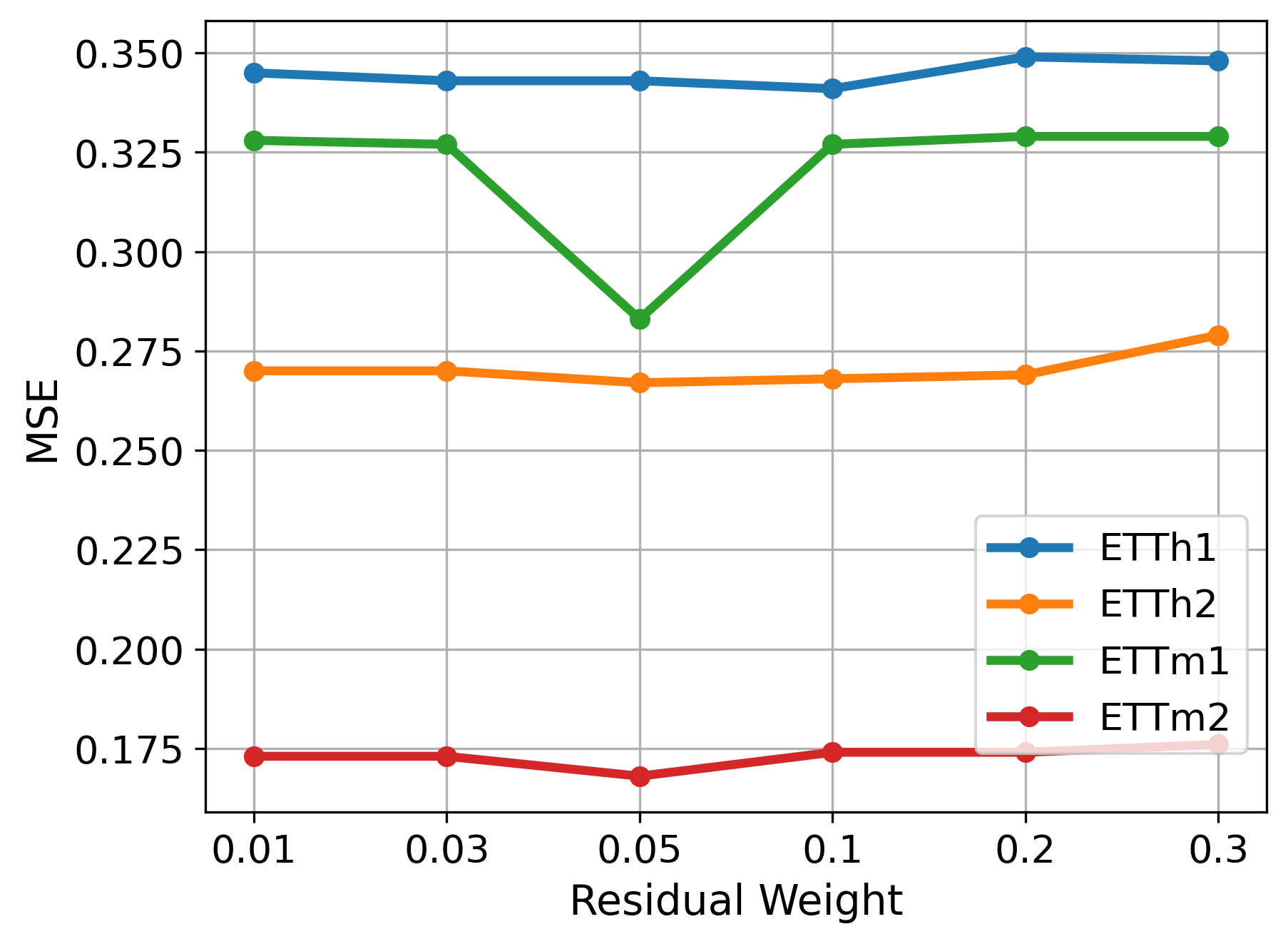}
    \small (d) Residual weight
\end{minipage}
\caption{Sensitivity analysis of key hyper-parameters.}
\label{fig:hyper_analysis}
\end{figure}

\subsection{Hyper-parameter Analysis}
We analyze four key hyper-parameters in Figure~\ref{fig:hyper_analysis}. \textbf{(a) Look-back window:} performance improves up to 1440 and saturates beyond, indicating that longer history helps but redundancy eventually sets in. \textbf{(b) Periodicity:} the optimum is 24 on ETTh and 96 on ETTm, matching the $4\times$ sampling-rate gap between the two and confirming that periodicity should align with the daily cycle of the data. \textbf{(c) Hidden dimension:} performance stays nearly flat across $d_{\text{model}} \in \{64,\dots,512\}$, showing the method is not capacity-bound. \textbf{(d) Residual weight $\lambda$:} the best result is achieved at $\lambda{=}0.05$; smaller values weaken SMA's enhancement, while larger values distort the original signal. Overall, SSDA is robust across a wide range of settings.

\subsection{Validation of Gap Mitigation}
\label{sec:effectiveness_analysis}

\subparagraph{Spectral gap mitigation.}
To verify that SMA effectively closes the spectral gap, we conduct a PSS analysis on both the original rendered images and the SMA-enhanced images. As shown in Table~\ref{tab:sae_pss}, the PSS values consistently increase after applying SMA across all datasets, indicating that SMA successfully shifts the frequency distribution of rendered time series images toward natural image statistics, thereby narrowing the spectral gap identified in Section~\ref{introduction}.

\begin{table}[h]
\centering
\caption{PSS comparison between original images and SMA-enhanced images.}
\label{tab:sae_pss}
\setlength{\tabcolsep}{6pt}
\renewcommand{\arraystretch}{1.1}
\begin{tabular}{lcccc}
\toprule
 & ETTh1 & ETTh2 & ETTm1 & ETTm2 \\
\midrule
Original TS images      & 1.37 & 1.98 & 1.68 & 1.62 \\
SMA enhanced TS images & \textbf{1.48} & \textbf{2.06} & \textbf{1.78} & \textbf{1.68} \\
\bottomrule
\end{tabular}
\end{table}

\subparagraph{Structural gap mitigation.}

\begin{figure}[H]
    \centering
    \begin{minipage}[t]{0.20\textwidth}
        \centering
        \includegraphics[width=1\textwidth]{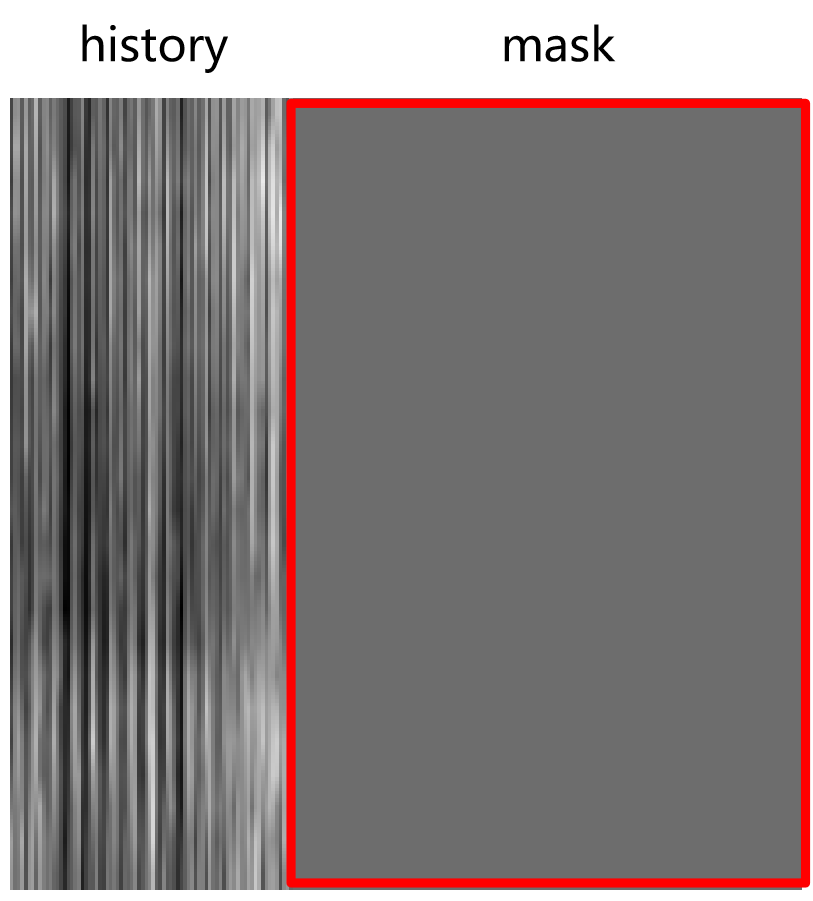}
        \label{fig:ettm1_var4_ti}
    \end{minipage}
    \hspace{2em}
    \begin{minipage}[t]{0.20\textwidth}
        \centering
        \includegraphics[width=1\textwidth]{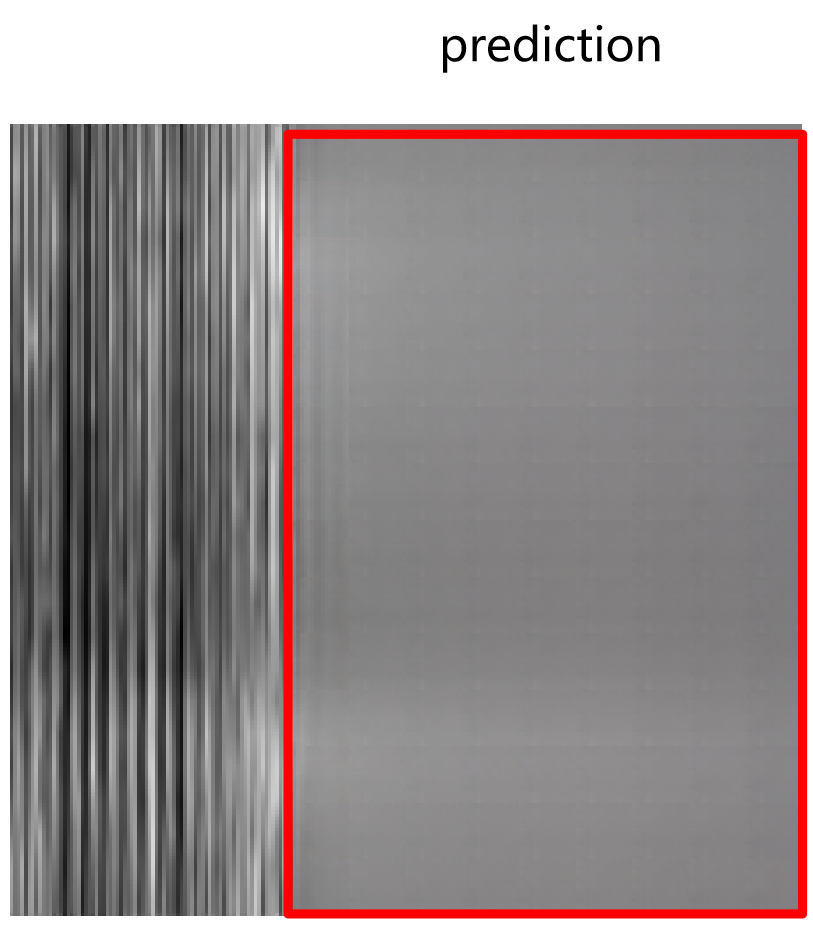}
        \label{fig:ettm1_var4_vr}
    \end{minipage}
    \hspace{2em}
    \begin{minipage}[t]{0.20\textwidth}
        \centering
        \includegraphics[width=1\textwidth]{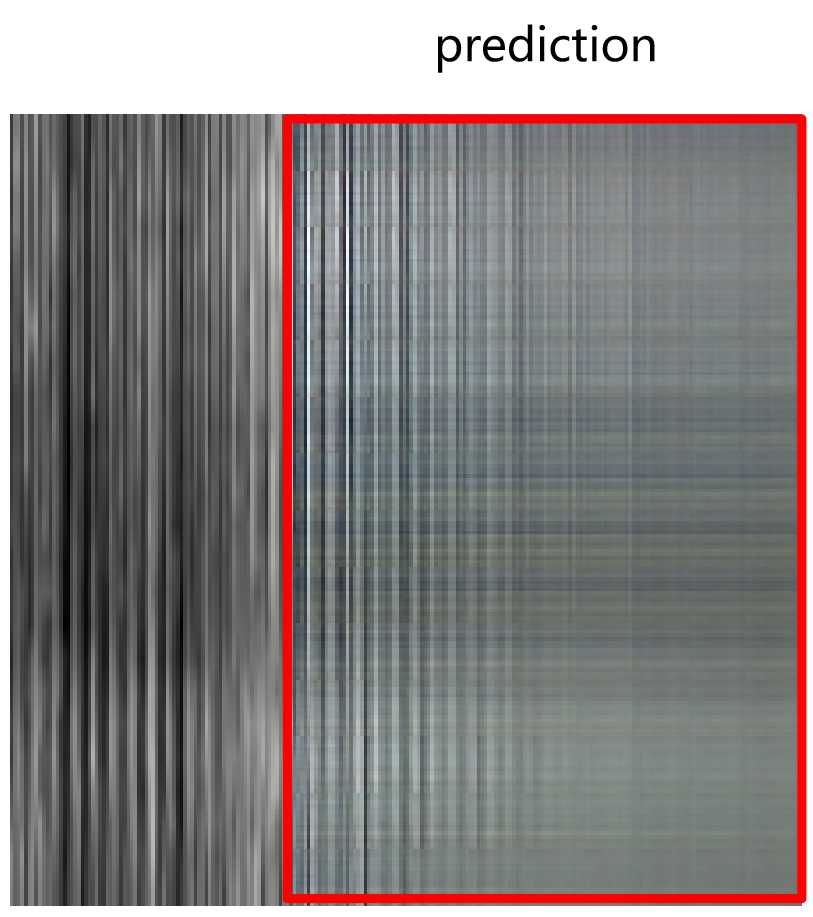}
        \label{fig:ettm1_var4_tr}
    \end{minipage}
\caption{Visualization of structural gap mitigation on an ETTm1 sample. \textbf{Left:} original rendered image. \textbf{Middle:} spectral branch output, which preserves the spurious 2D spatial layout. \textbf{Right:} structural branch output, which exhibits pronounced vertical patterns aligned with the 1D temporal ordering, indicates that SG-LoRA bridges the structural gap.}
    \label{fig:structural_branch_performance}
\end{figure}

Figure~\ref{fig:structural_branch_performance} presents a visualization comparison of the reconstructed images from both branches. The spectral branch produces relatively smooth structures that largely preserve the spurious spatial layout introduced by 2D reshaping, and therefore does not explicitly resolve the structural gap. In contrast, the structural branch yields reconstructed images with pronounced vertical patterns, which are consistent with the one-dimensional temporal ordering of the original series (where each column corresponds to a temporal slice).
This observation indicates that SG-LoRA effectively restores temporal coherence that is disrupted during the image transformation process, while suppressing misleading spatial correlations. More examples and corresponding prediction results are provided in Appendix~\ref{apd:visualizaiton}.



\section{Conclusion}

In this paper, we revisit a largely unexamined premise behind LVM-based time series forecasting and identify two distinct facets of the modality gap between rendered and natural images: a \emph{spectral} gap in data statistics and a \emph{structural} gap induced by 2D folding. Guided by this diagnosis, we propose \textbf{SSDA}, a dual-branch network that addresses each gap at the level where it originates. The Spectral Magnitude Aligner (SMA) reshapes frequency-domain statistics at the data level via selective magnitude enhancement under phase preservation, while Structural-Guided LoRA (SG-LoRA) restores temporal ordering at the model level through position-aware encodings and low-rank attention updates. Experiments on seven real-world benchmarks show that SSDA achieves state-of-the-art performance in both full-shot and few-shot settings over 12 strongest baselines. We hope our findings could open new avenues for further cross-modality research.

\newpage


\bibliographystyle{plain}
\bibliography{base}

@article{Morid2023,
  title={Sparse learned kernels for interpretable and efficient medical time series processing},
  author={Chen, Sully F and Guo, Zhicheng and Ding, Cheng and Hu, Xiao and Rudin, Cynthia},
  journal={Nature machine intelligence},
  volume={6},
  number={10},
  pages={1132--1144},
  year={2024},
  publisher={Nature Publishing Group UK London}
}

@inproceedings{liang2024foundation,
  title={Foundation models for time series analysis: A tutorial and survey},
  author={Liang, Yuxuan and Wen, Haomin and Nie, Yuqi and Jiang, Yushan and Jin, Ming and Song, Dongjin and Pan, Shirui and Wen, Qingsong},
  booktitle={Proceedings of the 30th ACM SIGKDD conference on knowledge discovery and data mining},
  pages={6555--6565},
  year={2024}
}

@inproceedings{feichtenhofer2022masked,
  title={Masked autoencoders as spatiotemporal learners},
  author={Feichtenhofer, Christoph and Fan, Haoqi and He, Yanghao Li Kaiming},
  booktitle={Proceedings of the 36th International Conference on Neural Information Processing Systems},
  pages={35946--35958},
  year={2022}
}

@inproceedings{yu2023harnessing,
  title={Harnessing LLMs for temporal data-a study on explainable financial time series forecasting},
  author={Yu, Xinli and Chen, Zheng and Lu, Yanbin},
  booktitle={Proceedings of the 2023 conference on empirical methods in natural language processing: industry track},
  pages={739--753},
  year={2023}
}

@inproceedings{Koprinska2018,
  title={Amplifier: Bringing attention to neglected low-energy components in time series forecasting},
  author={Fei, Jingru and Yi, Kun and Fan, Wei and Zhang, Qi and Niu, Zhendong},
  booktitle={Proceedings of the AAAI conference on artificial intelligence},
  volume={39},
  number={11},
  pages={11645--11653},
  year={2025}
}

@article{yang2025towards,
  title={Towards Robust and Interpretable Spatial-Temporal Graph Modeling for Traffic Prediction},
  author={Yang, Hanchen and Cao, Jiannong and Li, Wengen and Yang, Yu and Li, Xiaoyi and Kong, Lingbai and Zhang, Yichao and Guan, Jihong and Zhou, Shuigeng},
  journal={ACM Transactions on Knowledge Discovery from Data},
  volume={19},
  number={9},
  pages={1--20},
  year={2025},
  publisher={ACM New York, NY}
}

@inproceedings{zhou2023,
  title={One fits all: Power general time series analysis by pretrained lm},
  author={Zhou, Tian and Niu, Peisong and Sun, Liang and Jin, Rong and others},
  booktitle={Advances in neural information processing systems},
  volume={36},
  pages={43322--43355},
  year={2023}
}

@article{niu2024,
  title={Understanding the role of textual prompts in llm for time series forecasting: an adapter view},
  author={Niu, Peisong and Zhou, Tian and Wang, Xue and Sun, Liang and Jin, Rong},
  journal={arXiv preprint arXiv:2311.14782},
  year={2023}
}

@inproceedings{goswami2024,
  title={MOMENT: A Family of Open Time-series Foundation Models},
  author={Goswami, Mononito and Szafer, Konrad and Choudhry, Arjun and Cai, Yifu and Li, Shuo and Dubrawski, Artur},
  booktitle={International Conference on Machine Learning},
  pages={16115--16152},
  year={2024},
  organization={PMLR}
}

@inproceedings{visionts2025,
  title={VisionTS: Visual Masked Autoencoders Are Free-Lunch Zero-Shot Time Series Forecasters},
  author={Chen, Mouxiang and Shen, Lefei and Li, Zhuo and Wang, Xiaoyun Joy and Sun, Jianling and Liu, Chenghao},
  booktitle={International Conference on Machine Learning},
  pages={8979--9007},
  year={2025},
  organization={PMLR}
}

@inproceedings{dmmv2025,
      title={Multi-Modal View Enhanced Large Vision Models for Long-Term Time Series Forecasting}, 
      author={ChengAo Shen and Wenchao Yu and Ziming Zhao and Dongjin Song and Wei Cheng and Haifeng Chen and Jingchao Ni},
  booktitle={The Thirty-ninth Annual Conference on Neural Information Processing Systems},
      year={2025},
      pages={1--28},
}

@inproceedings{timevlm2025,
  title={Time-VLM: Exploring Multimodal Vision-Language Models for Augmented Time Series Forecasting},
  author={Zhong, Siru and Ruan, Weilin and Jin, Ming and Li, Huan and Wen, Qingsong and Liang, Yuxuan},
  booktitle={International Conference on Machine Learning},
  pages={78478--78497},
  year={2025},
  organization={PMLR}
}

@article{zhang2026piformer,
  title={PiFormer: Towards Subseasonal SST Prediction with Spatial-Patched Inverted Transformer},
  author={Zhang, Mingrui and Yang, Hanchen and Li, Wengen and Jiang, Xudong and Guan, Jihong and Zhang, Yichao and Zhou, Shuigeng},
  journal={Expert Systems with Applications},
  pages={131618},
  year={2026},
  publisher={Elsevier}
}

@inproceedings{timecma2025,
  title={Timecma: Towards llm-empowered multivariate time series forecasting via cross-modality alignment},
  author={Liu, Chenxi and Xu, Qianxiong and Miao, Hao and Yang, Sun and Zhang, Lingzheng and Long, Cheng and Li, Ziyue and Zhao, Rui},
  booktitle={Proceedings of the AAAI Conference on Artificial Intelligence},
  volume={39},
  number={18},
  pages={18780--18788},
  year={2025}
}

@article{Li2020,
title = {Forecasting with time series imaging},
journal = {Expert Systems with Applications},
volume = {160},
pages = {113680},
year = {2020},
issn = {0957-4174},
doi = {https://doi.org/10.1016/j.eswa.2020.113680},
url = {https://www.sciencedirect.com/science/article/pii/S0957417420305042},
author = {Xixi Li and Yanfei Kang and Feng Li},
}

@inproceedings{
vit2021,
title={An Image is Worth 16x16 Words: Transformers for Image Recognition at Scale},
author={Alexey Dosovitskiy and Lucas Beyer and Alexander Kolesnikov and Dirk Weissenborn and Xiaohua Zhai and Thomas Unterthiner and Mostafa Dehghani and Matthias Minderer and Georg Heigold and Sylvain Gelly and Jakob Uszkoreit and Neil Houlsby},
booktitle={International Conference on Learning Representations},
year={2021},
pages={1-21}
}

@inproceedings{mae2021,
  title={Masked autoencoders are scalable vision learners},
  author={He, Kaiming and Chen, Xinlei and Xie, Saining and Li, Yanghao and Doll{\'a}r, Piotr and Girshick, Ross},
  booktitle={Proceedings of the IEEE conference on computer vision and pattern recognition},
  pages={16000--16009},
  year={2022}
}

@inproceedings{clip2021,
  title={Learning transferable visual models from natural language supervision},
  author={Radford, Alec and Kim, Jong Wook and Hallacy, Chris and Ramesh, Aditya and Goh, Gabriel and Agarwal, Sandhini and Sastry, Girish and Askell, Amanda and Mishkin, Pamela and Clark, Jack and others},
  booktitle={International conference on machine learning},
  pages={8748--8763},
  year={2021},
  organization={PMLR}
}

@inproceedings{resnet2015,
  title={Deep residual learning for image recognition},
  author={He, Kaiming and Zhang, Xiangyu and Ren, Shaoqing and Sun, Jian},
  booktitle={Proceedings of the IEEE conference on computer vision and pattern recognition},
  pages={770--778},
  year={2016}
}

@inproceedings{vgg192015,
  edition = {},
  number = {},
  booktitle = {International Conference on Learning Representations},
  pages = {1-14},
  publisher = {Computational and Biological Learning Society},
  school = {},
  title = {Very deep convolutional networks for large-scale image recognition},
  volume = {},
  author = {Simonyan, K and Zisserman, A},
  editor = {},
  year = {2015},
  organizer = {},
  series = {}
}

@article{
vitime2025,
title={ViTime: Foundation Model for Time Series Forecasting Powered by Vision Intelligence},
author={Luoxiao Yang and Yun Wang and Xinqi Fan and Israel Cohen and jingdong chen and Zijun Zhang},
journal={Transactions on Machine Learning Research},
issn={2835-8856},
year={2025},
url={https://openreview.net/forum?id=XInsJDBIkp},
note={}
}

@inproceedings{vilt2021,
  title={Vilt: Vision-and-language transformer without convolution or region supervision},
  author={Kim, Wonjae and Son, Bokyung and Kim, Ildoo},
  booktitle={International conference on machine learning},
  pages={5583--5594},
  year={2021},
  organization={PMLR}
}

@inproceedings{zeng2024,
  title={From pixels to predictions: Spectrogram and vision transformer for better time series forecasting},
  author={Zeng, Zhen and Kaur, Rachneet and Siddagangappa, Suchetha and Balch, Tucker and Veloso, Manuela},
  booktitle={Proceedings of the Fourth ACM International Conference on AI in Finance},
  pages={82--90},
  year={2023}
}

@inproceedings{shen2023,
  title={Cross-modal fine-tuning: Align then refine},
  author={Shen, Junhong and Li, Liam and Dery, Lucio M and Staten, Corey and Khodak, Mikhail and Neubig, Graham and Talwalkar, Ameet},
  booktitle={International Conference on Machine Learning},
  pages={31030--31056},
  year={2023},
  organization={PMLR}
}

@inproceedings{
timellm2024,
title={Time-{LLM}: Time Series Forecasting by Reprogramming Large Language Models},
author={Ming Jin and Shiyu Wang and Lintao Ma and Zhixuan Chu and James Y. Zhang and Xiaoming Shi and Pin-Yu Chen and Yuxuan Liang and Yuan-Fang Li and Shirui Pan and Qingsong Wen},
booktitle={The Twelfth International Conference on Learning Representations},
year={2024},
url={https://openreview.net/forum?id=Unb5CVPtae}
}

@inproceedings{informer2021,
  title={Informer: Beyond efficient transformer for long sequence time-series forecasting},
  author={Zhou, Haoyi and Zhang, Shanghang and Peng, Jieqi and Zhang, Shuai and Li, Jianxin and Xiong, Hui and Zhang, Wancai},
  booktitle={Proceedings of the AAAI conference on artificial intelligence},
  volume={35},
  number={12},
  pages={11106--11115},
  year={2021}
}

@misc{Trindade2015,
  author       = {Trindade, Artur},
  title        = {{ElectricityLoadDiagrams20112014}},
  year         = {2015},
  howpublished = {UCI Machine Learning Repository},
}

@inproceedings{autoformer2022,
  title={Autoformer: Decomposition transformers with auto-correlation for long-term series forecasting},
  author={Wu, Haixu and Xu, Jiehui and Wang, Jianmin and Long, Mingsheng},
  booktitle={Advances in neural information processing systems},
  volume={34},
  pages={22419--22430},
  year={2021}
}

@inproceedings{fedformer2022,
  title={Fedformer: Frequency enhanced decomposed transformer for long-term series forecasting},
  author={Zhou, Tian and Ma, Ziqing and Wen, Qingsong and Wang, Xue and Sun, Liang and Jin, Rong},
  booktitle={International conference on machine learning},
  pages={27268--27286},
  year={2022},
  organization={PMLR}
}

@article{dlinear2022, title={Are Transformers Effective for Time Series Forecasting?}, volume={37}, url={https://ojs.aaai.org/index.php/AAAI/article/view/26317}, DOI={10.1609/aaai.v37i9.26317}, abstractNote={Recently, there has been a surge of Transformer-based solutions for the long-term time series forecasting (LTSF) task. Despite the growing performance over the past few years, we question the validity of this line of research in this work. Specifically, Transformers is arguably the most successful solution to extract the semantic correlations among the elements in a long sequence. However, in time series modeling, we are to extract the temporal relations in an ordered set of continuous points. While employing positional encoding and using tokens to embed sub-series in Transformers facilitate preserving some ordering information, the nature of the permutation-invariant self-attention mechanism inevitably results in temporal information loss. To validate our claim, we introduce a set of embarrassingly simple one-layer linear models named LTSF-Linear for comparison. Experimental results on nine real-life datasets show that LTSF-Linear surprisingly outperforms existing sophisticated Transformer-based LTSF models in all cases, and often by a large margin. Moreover, we conduct comprehensive empirical studies to explore the impacts of various design elements of LTSF models on their temporal relation extraction capability. We hope this surprising finding opens up new research directions for the LTSF task. We also advocate revisiting the validity of Transformer-based solutions for other time series analysis tasks (e.g., anomaly detection) in the future.}, number={9}, journal={Proceedings of the AAAI Conference on Artificial Intelligence}, author={Zeng, Ailing and Chen, Muxi and Zhang, Lei and Xu, Qiang}, year={2023}, month={Jun.}, pages={11121-11128} }

@article{Field1987,
  author    = {Field, D. J.},
  title     = {Relations between the statistics of natural images and the response 
               properties of cortical cells},
  journal   = {Journal of the Optical Society of America. A, Optics and Image Science},
  year      = {1987},
  volume    = {4},
  number    = {12},
  pages     = {2379--2394},
  doi       = {10.1364/josaa.4.002379}
}

@article{visionts++,
  title={VisionTS++: Cross-Modal Time Series Foundation Model with Continual Pre-trained Vision Backbones},
  author={Shen, Lefei and Chen, Mouxiang and Liu, Xu and Fu, Han and Ren, Xiaoxue and Sun, Jianling and Li, Zhuo and Liu, Chenghao},
  journal={arXiv preprint arXiv:2508.04379},
  year={2025}
}

@inproceedings{
text2freq2024,
title={Text2Freq: Learning Series Patterns from Text via Frequency Domain},
author={Ming-Chih Lo and Ching Chang and Wen-Chih Peng},
booktitle={NeurIPS Workshop on Time Series in the Age of Large Models},
year={2024},
url={https://openreview.net/forum?id=Pi6sA1MSSr}
}

@article{nguyen2026,
  title={Spectral Text Fusion: A Frequency-Aware Approach to Multimodal Time-Series Forecasting},
  author={Nguyen, Huu Hiep and Nguyen, Minh Hoang and Nguyen, Dung and Le, Hung},
  journal={arXiv preprint arXiv:2602.01588},
  year={2026}
}

@inproceedings{Harnessing2025,
  title={Harnessing vision models for time series analysis: a survey},
  author={Ni, Jingchao and Zhao, Ziming and Shen, ChengAo and Tong, Hanghang and Song, Dongjin and Cheng, Wei and Luo, Dongsheng and Chen, Haifeng},
  booktitle={Proceedings of the Thirty-Fourth International Joint Conference on Artificial Intelligence},
  pages={10612--10620},
  year={2025}
}

@article{timeapn2026,
  title={TimeAPN: Adaptive Amplitude-Phase Non-Stationarity Normalization for Time Series Forecasting},
  author={Hu, Yue and Tang, Jialiang and Yu, Siwei and Yu, Baosheng and Zhang, Jing and Tao, Dacheng},
  journal={arXiv preprint arXiv:2603.17436},
  year={2026}
}

@inproceedings{fredn2026,
  title={Fredn: Spectral disentanglement for time series forecasting via learnable frequency decomposition},
  author={An, Zhongde and You, Jinhong and Li, Jiyanglin and Tang, Yiming and Li, Wen and Du, Heming and Du, Shouguo},
  booktitle={Proceedings of the AAAI Conference on Artificial Intelligence},
  volume={40},
  number={24},
  pages={19623--19631},
  year={2026}
}

@article{pss2022,
  title = {Reliability and Validity of Power Spectrum Slope (PSS): A Metric for Measuring Resting-State Functional Magnetic Resonance Imaging Activity of Single Voxels},
  volume = {16},
  issn = {1662-4548, 1662-453X},
  url = {https://www.frontiersin.org/articles/10.3389/fnins.2022.871609/full},
  doi = {10.3389/fnins.2022.871609},
  language = {eng},
  urldate = {2026-04-06},
  journal = {Frontiers in Neuroscience},
  author = {Zang, Zhenxiang and Qiao, Yang and Yan, Shaozhen and Lu, Jie},
  year = {2022},
  pages = {871609},
}

@inproceedings{
patchtst2023,
title={A Time Series is Worth 64 Words:  Long-term Forecasting with Transformers},
author={Yuqi Nie and Nam H Nguyen and Phanwadee Sinthong and Jayant Kalagnanam},
booktitle={The Eleventh International Conference on Learning Representations },
pages = {1-10},
year={2023},
url={https://openreview.net/forum?id=Jbdc0vTOcol}
}

@inproceedings{
timesnet2023,
title={TimesNet: Temporal 2D-Variation Modeling for General Time Series Analysis},
author={Haixu Wu and Tengge Hu and Yong Liu and Hang Zhou and Jianmin Wang and Mingsheng Long},
booktitle={The Eleventh International Conference on Learning Representations },
year={2023},
pages = {1-10},
url={https://openreview.net/forum?id=ju_Uqw384Oq}
}

@inproceedings{Verma2024,
  title={Cross-modal projection in multimodal llms doesn’t really project visual attributes to textual space},
  author={Verma, Gaurav and Choi, Minje and Sharma, Kartik and Watson-Daniels, Jamelle and Oh, Sejoon and Kumar, Srijan},
  booktitle={Proceedings of the 62nd Annual Meeting of the Association for Computational Linguistics (Volume 2: Short Papers)},
  pages={657--664},
  year={2024}
}

@inproceedings{ghosal2023,
  title={Text-to-audio generation using instruction guided latent diffusion model},
  author={Ghosal, Deepanway and Majumder, Navonil and Mehrish, Ambuj and Poria, Soujanya},
  booktitle={Proceedings of the 31st ACM international conference on multimedia},
  pages={3590--3598},
  year={2023}
}

@inproceedings{Lu2021PretrainedTA,
  title={Frozen pretrained transformers as universal computation engines},
  author={Lu, Kevin and Grover, Aditya and Abbeel, Pieter and Mordatch, Igor},
  booktitle={Proceedings of the AAAI conference on artificial intelligence},
  volume={36},
  number={7},
  pages={7628--7636},
  year={2022}
}

@inproceedings{Hassid2023,
  title={Textually pretrained speech language models},
  author={Hassid, Michael and Remez, Tal and Nguyen, Tu Anh and Gat, Itai and Conneau, Alexis and Kreuk, Felix and Copet, Jade and Defossez, Alexandre and Synnaeve, Gabriel and Dupoux, Emmanuel and others},
  booktitle={Advances in Neural Information Processing Systems},
  volume={36},
  pages={63483--63501},
  year={2023}
}

@inproceedings{loftllm2026,
  title={LoFT-LLM: Low-Frequency Time-Series Forecasting with Large Language Models},
  author={You, Jiacheng and Yang, Jingcheng and Xie, Yuhang and Wu, Zhongxuan and Li, Xiucheng and Li, Feng and Wang, Pengjie and Xu, Jian and Zheng, Bo and Chen, Xinyang},
  booktitle={Proceedings of the 32nd ACM SIGKDD Conference on Knowledge Discovery and Data Mining V. 1},
  pages={1809--1820},
  year={2026}
}

@inproceedings{ruderman1994,
  title={Statistics of natural images: scaling in the woods},
  author={Ruderman, Daniel L and Bialek, William},
  booktitle={Proceedings of the 7th International Conference on Neural Information Processing Systems},
  pages={551--558},
  year={1993}
}

@INPROCEEDINGS{deng2009,
  author={Deng, Jia and Dong, Wei and Socher, Richard and Li, Li-Jia and Kai Li and Li Fei-Fei},
  booktitle={2009 IEEE Conference on Computer Vision and Pattern Recognition}, 
  title={ImageNet: A large-scale hierarchical image database}, 
  year={2009},
  volume={},
  number={},
  pages={248-255},
  doi={10.1109/CVPR.2009.5206848}}

@inproceedings{
test2024,
title={{TEST}: Text Prototype Aligned Embedding to Activate {LLM}'s Ability for Time Series},
author={Chenxi Sun and Hongyan Li and Yaliang Li and Shenda Hong},
booktitle={The Twelfth International Conference on Learning Representations},
year={2024},
pages={1--21}
}

@misc{adamamba2026,
      title={AdaMamba: Adaptive Frequency-Gated Mamba for Long-Term Time Series Forecasting}, 
      author={Xudong Jiang and Mingshan Loo and Hanchen Yang and Wengen Li and Mingrui Zhang and Yichao Zhang and Jihong Guan and Shuigeng Zhou},
      year={2026},
      eprint={2604.23239},
      archivePrefix={arXiv},
      primaryClass={cs.AI},
      url={https://arxiv.org/abs/2604.23239}, 
}

\clearpage
\appendix

\section{Modality Gap Analysis}
\label{apd:modality_gap_analysis}

\subsection{Experimental Principle}

\subparagraph{Frequency Distribution and Its Physical Significance}

Any signal in the spatial (or temporal) domain can be decomposed into a sum of sinusoidal components of different frequencies through Fourier transform. The frequency spectrum reveals how signal energy is distributed across different frequency components. Low frequencies correspond to slow spatial variations (e.g., overall brightness, large-scale structures), while high frequencies capture rapid changes (e.g., edges, textures, fine details).

\begin{figure}[H]
\centering

\begin{minipage}{0.24\textwidth}
    \centering
    \includegraphics[width=\linewidth]{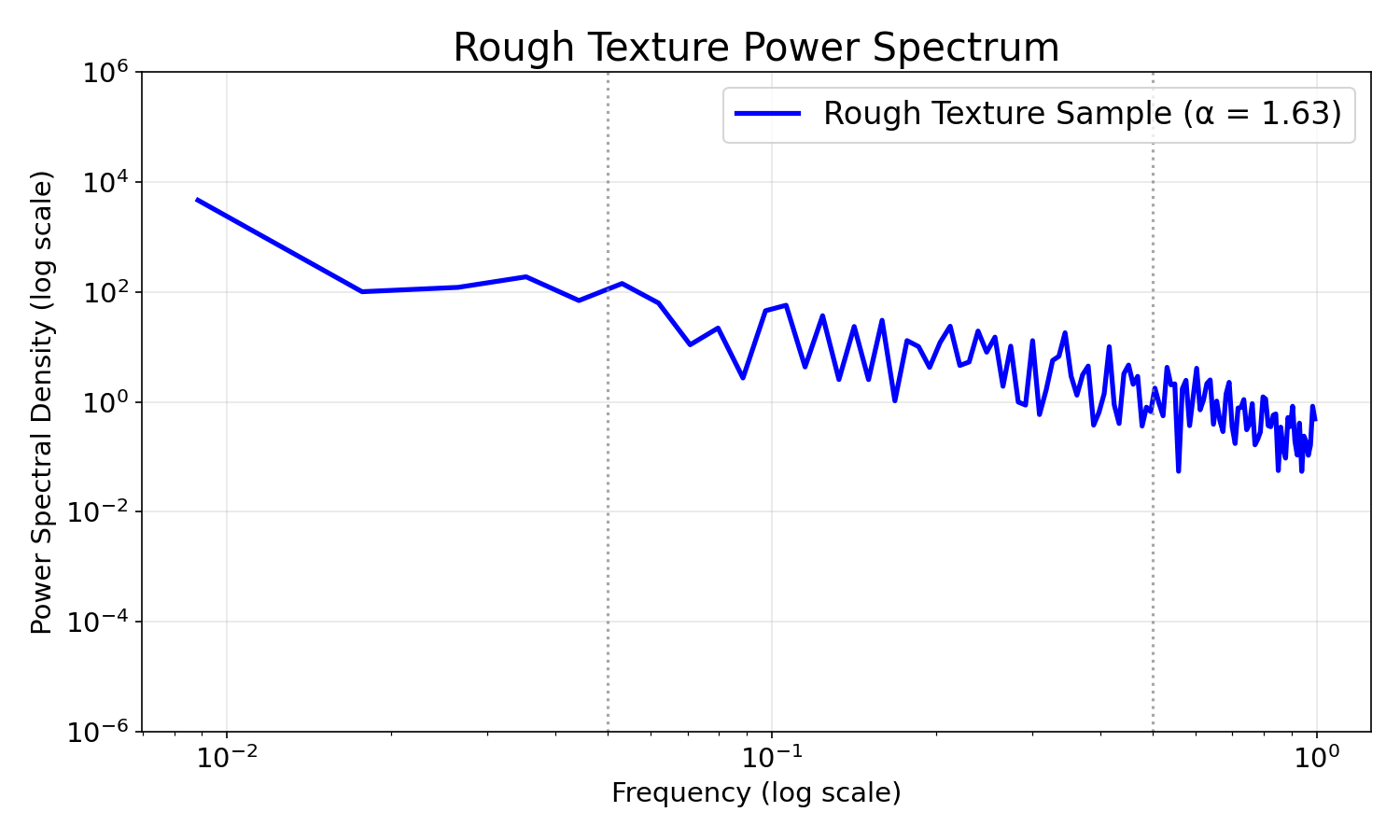}
    \vspace{2mm}
    \includegraphics[width=0.96\linewidth]{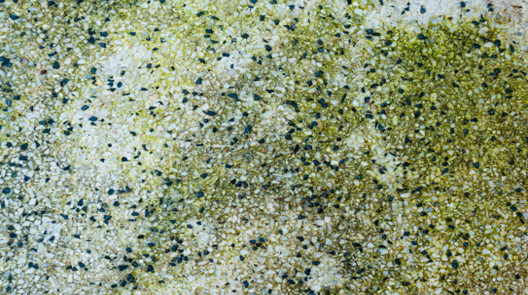}
    \small (a) Fine gravel and sand($\alpha=1.66$)
\end{minipage}
\hfill
\begin{minipage}{0.24\textwidth}
    \centering
    \includegraphics[width=\linewidth]{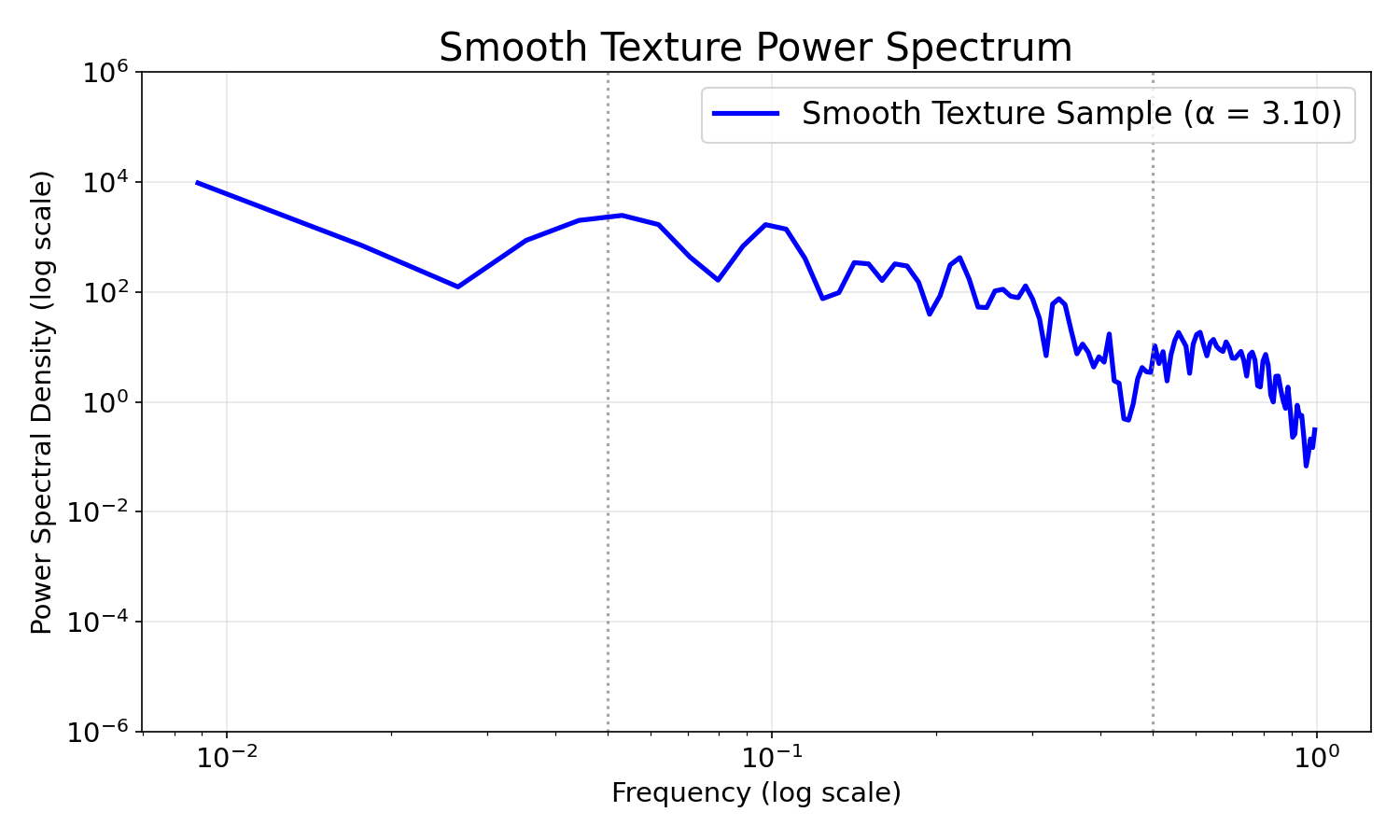}
    \vspace{2mm}
    \includegraphics[width=0.96\linewidth]{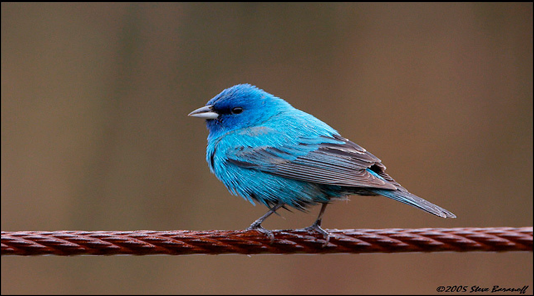}
    \small (b) Bird with blurred background($\alpha=3.10$)
\end{minipage}
\hfill
\begin{minipage}{0.24\textwidth}
    \centering
    \includegraphics[width=\linewidth]{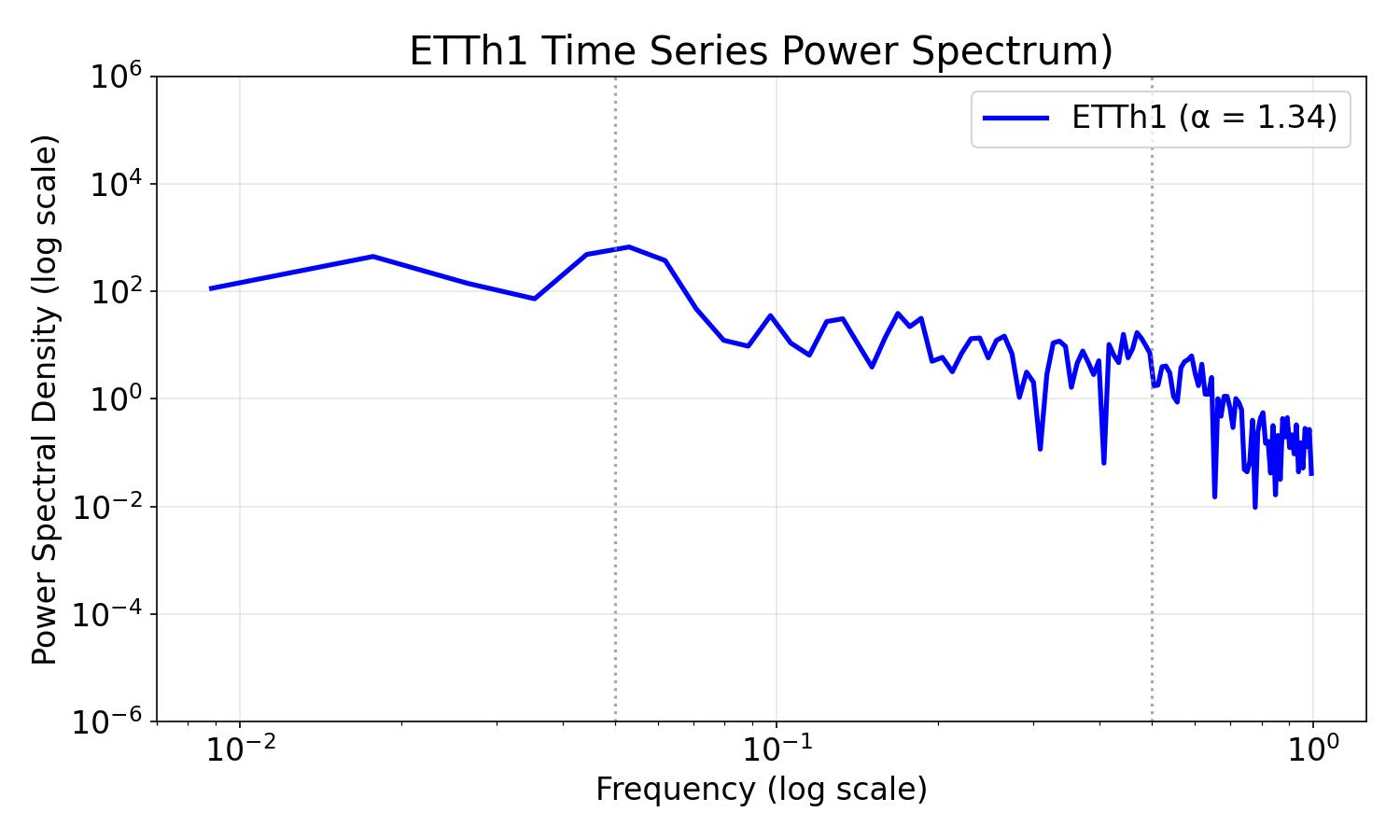}
    \vspace{2mm}
    \includegraphics[width=\linewidth]{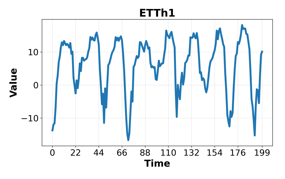}
    \small (c) Time series : ETTh1($\alpha=1.34$)
\end{minipage}
\hfill
\begin{minipage}{0.24\textwidth}
    \centering
    \includegraphics[width=\linewidth]{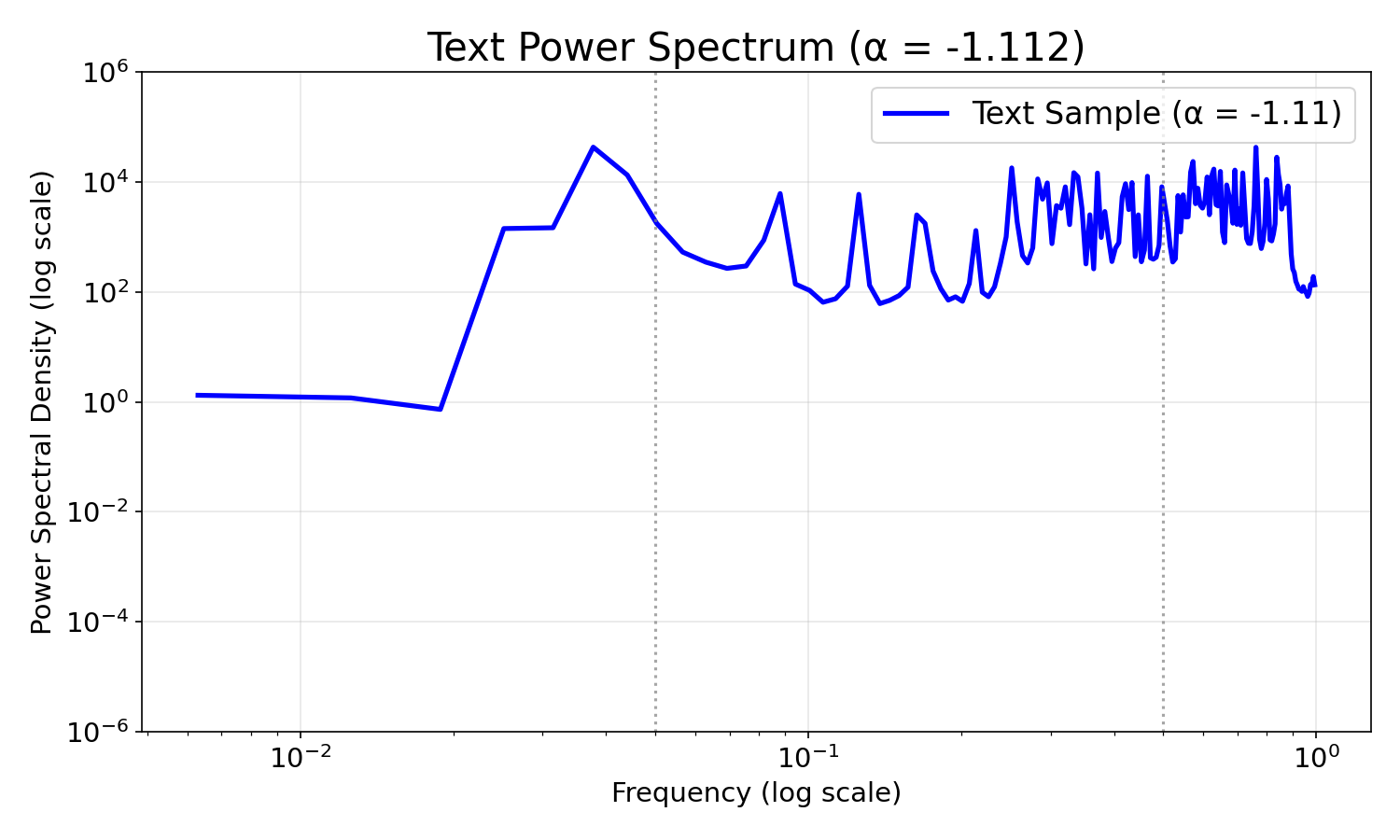}
    \vspace{2mm}
    \includegraphics[width=\linewidth]{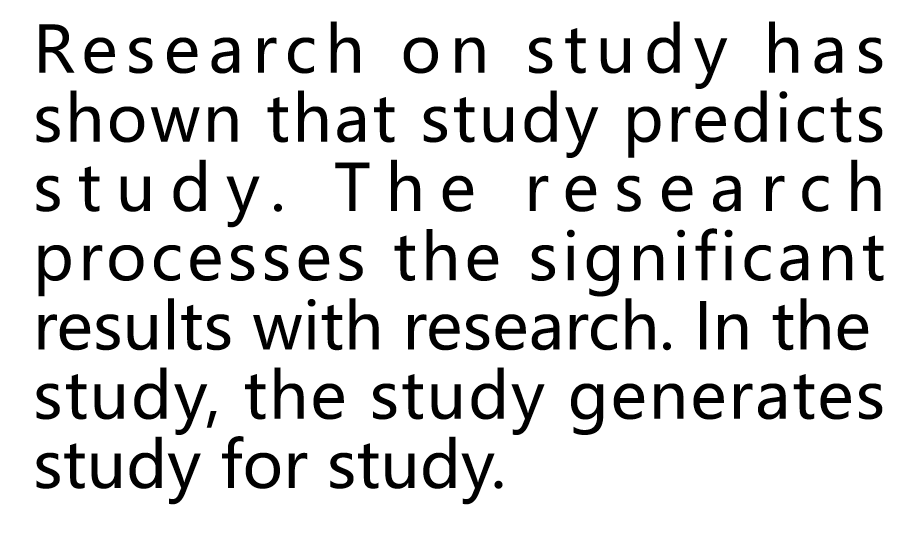}
    \small (d) Text : Wikipedia corpus($\alpha=-1.11$)
\end{minipage}

\caption{Power spectrum analysis of different modalities. Each subfigure shows the power spectrum (top) and the original modality example (bottom). (a) Rough-texture image (fine gravel and sand): $\alpha = 1.66$, indicating high-frequency content from fine textures; (b) Smooth-texture image (bird with blurred background): $\alpha = 3.10$, with gentler decay reflecting large-scale homogeneity; (c) Time series (ETTh1): $\alpha = 1.34$; (d) Text (Wikipedia corpus): $\alpha = -1.11$.}
\label{fig:modality_pss}
\end{figure}

Taking natural images as an example, as shown in Figure \ref{fig:modality_pss}(a)-(b), the texture details in images (such as the rough surface of sand or the smooth surface of leaves) manifest as different frequency characteristics. Rough textures contain more high-frequency components, while smooth regions are dominated by low frequencies. This principle extends naturally to time series data (Figure \ref{fig:modality_pss}(c)), where seasonal patterns and periodic trends correspond to low-frequency oscillations, while noise and sudden fluctuations occupy the high-frequency band. Interestingly, even text (Figure \ref{fig:modality_pss}(d)), when encoded as character sequences and transformed into spatial representations, exhibits structured frequency patterns that reflect the statistical regularities of language.

\subparagraph{Power Spectrum Slope} To quantitatively characterize these frequency-domain properties, we employ \textbf{Power Spectrum Slope (PSS)} analysis. PSS---defined as the slope of the power spectrum in log--log space---reveals distinct scaling behaviors across modalities. It quantifies the decay rate $\alpha$ in the characteristic power-law scaling $P(f) \propto f^{-\alpha}$, serving as a powerful indicator of the signal's statistical complexity and self-similarity.

In natural images, this manifests as the well-established $1/f^{\alpha}$ scaling (typically $\alpha \approx 2$) ~\cite{Field1987,ruderman1994}, where deviations encode texture properties---steeper slopes correspond to smoother textures with fewer fine details, and shallower slopes indicate rougher, more granular textures. This property arises from the statistical structure of natural scenes and has been widely used in image processing, computer vision, and neuroscience~\cite{timeapn2026,pss2022}.
As shown in Figure \ref{fig:modality_pss}, our measurements ($\alpha = 1.66$ for rough-texture, $\alpha = 3.10$ for smooth-texture, $\alpha = 1.34$ for ETTh1 time series, and $\alpha = -1.11$ for Wikipedia text) reveal substantial variations in frequency distributions across modalities. The PSS thus serves as an effective metric for quantifying these modal disparities. Consequently, by systematically comparing the power-law exponents across these modalities, we can quantitatively measure the modality gap and provide empirical evidence for addressing such distributional shifts in multi-modal time series modeling frameworks.

\subsection{Datasets}


\subparagraph{Time Series Data} We employ the Electricity Transformer Temperature (ETT) benchmark~\cite{informer2021} with all four subsets (ETTh1, ETTh2, ETTm1, ETTm2). For each subset, we randomly sample 100 sequences, each with length $L = 1440$, from the full dataset. Each sequence is rendered into a $224 \times 224$ grayscale image via SSDA's rendering pipeline.
    
\subparagraph{Natural Images} We randomly select 100 images from the ImageNet-100 dataset~\cite{deng2009}, a curated subset of ImageNet containing 100 object categories. All images are preprocessed to a uniform resolution of $224 \times 224$ pixels to ensure consistency with the ViT architecture's input requirements. Each image is normalized to zero mean and unit standard deviation.
    
\subparagraph{Text Data} We extract 100 text samples from the English Wikipedia corpus (snapshot dated 20220301.en). Each text sample is converted into a numerical representation by mapping each character to its ASCII code value (ranging from 32 to 126 for printable characters), normalized to the range $[0, 1]$. The resulting 1D sequence is then tiled or truncated to match the target size of $224 \times 224 = 50,176$ elements, and reshaped into a 2D matrix.

\subsection{Calculation Process}

The power spectrum analysis pipeline consists of four sequential stages, each detailed below:

\subparagraph{(1) Data Preprocessing}

\begin{itemize}
    \item \textbf{Time series}: Loaded from ETT datasets (ETTh1, ETTh2, ETTm1, ETTm2), standardized using training set statistics, then rendered into $224 \times 224$ images using the SSDA rendering method (periodic folding with bilinear interpolation).
    \item \textbf{Natural images}: Loaded from standard natural image datasets, resized to $224 \times 224$, normalized to zero mean and unit variance.
    \item \textbf{Text}: Converted to numerical sequence using ASCII encoding, reshaped into 2D image format, then processed identically to images.
\end{itemize}

\subparagraph{(2) 2D Discrete Fourier Transform}

For each preprocessed image $\mathbf{I}(x, y)$ of size $H \times W$, where $x \in [0, H-1]$ and $y \in [0, W-1]$ denote the spatial coordinates (row and column indices) respectively, we compute the 2D discrete Fourier transform:

\begin{equation}
    \mathcal{F}\{\mathbf{I}(x, y)\} = \hat{\mathbf{I}}(u, v) = \sum_{x=0}^{H-1}\sum_{y=0}^{W-1} \mathbf{I}(x, y) e^{-j2\pi\left(\frac{ux}{H} + \frac{vy}{W}\right)}
\end{equation}

where $u \in [0, H-1]$ and $v \in [0, W-1]$ are the frequency coordinates corresponding to the vertical and horizontal spatial frequencies respectively, $j = \sqrt{-1}$ is the imaginary unit, and the exponential term represents the complex sinusoid basis function at frequency $(u, v)$.

The power spectrum is then computed as the squared magnitude of the Fourier coefficients:

\begin{equation}
    P(u, v) = |\hat{\mathbf{I}}(u, v)|^2 = \hat{\mathbf{I}}(u, v) \cdot \hat{\mathbf{I}}^*(u, v)
\end{equation}

where $|\cdot|$ denotes the complex magnitude and $^*$ denotes complex conjugation. For efficiency, we apply the shift operation $\mathcal{F}_{shift}$ to center the zero frequency at the middle of the spectrum before computing the power.

\subparagraph{(3) Radial Averaging}

To reduce the 2D frequency plane to a 1D power spectrum, we perform radial averaging. Given the shifted power spectrum $P_{shift}(u, v)$, we compute the radial distance for each frequency component:

\begin{equation}
    r = \sqrt{(u - H/2)^2 + (v - W/2)^2}
\end{equation}

where $(H/2, W/2)$ is the center coordinates. The radial power spectrum is then computed as:

\begin{equation}
    P_{radial}(r_k) = \frac{1}{N_k} \sum_{(u,v): r \in \text{bin}(r_k)} P_{shift}(u, v)
\end{equation}

where $r_k$ denotes the $k$-th radial bin, $N_k$ is the count of frequency components falling into bin $k$, and $\text{bin}(r_k)$ assigns each $(u,v)$ pair to its corresponding radial bin based on $\lfloor r \rfloor$. The resulting 1D spectrum $P_{radial} \in \mathbb{R}^{R_+}$ represents the average power at each radial frequency $f_k = k / R_{max}$, where $R_{max} = \sqrt{(H/2)^2 + (W/2)^2}$ is the maximum radial frequency.

\subparagraph{(4) Power Law Fitting}

The radial power spectrum is fitted to the power law model in log-log space:

\begin{equation}
    P(f) \propto f^{-\alpha} \quad \Longleftrightarrow \quad \log P(f) = -\alpha \log f + C
\end{equation}

where $f$ denotes the normalized spatial frequency (in cycles per pixel, ranging from 0 to 0.5), $\alpha$ is the power law exponent (slope in log-log space) that characterizes the spectral decay rate, and $C$ is a constant offset (intercept) representing the baseline power level.

The fitting is performed using ordinary least squares linear regression on the log-transformed data within the frequency range $f \in [0.05, 0.5]$. This range is selected to exclude: (a) very low frequencies where finite image size effects dominate, and (b) very high frequencies dominated by noise and discretization artifacts. Formally, we minimize:

\begin{equation}
    \min_{\alpha, C} \sum_{i} \left[ \log P(f_i) - (-\alpha \log f_i + C) \right]^2
\end{equation}

subject to the frequency mask $\mathcal{M} = \{i : 0.05 < f_i < 0.5\}$. The fitting procedure also returns the coefficient of determination $R^2$ to assess goodness-of-fit.

For statistical robustness, we compute the mean and standard deviation of $\alpha$ across all samples within each modality:

\begin{equation}
    \bar{\alpha} = \frac{1}{N}\sum_{j=1}^{N} \alpha_j, \quad \sigma_\alpha = \sqrt{\frac{1}{N-1}\sum_{j=1}^{N}(\alpha_j - \bar{\alpha})^2}
\end{equation}

where $N$ is the number of samples per modality, and the subscript $j$ denotes the sample index.

\subsection{Experimental Results}

\begin{figure}[t]
    \centering
    \subfigure[Time Series (ETT subsets)]{\label{fig:pss_ts}\includegraphics[width=0.48\textwidth]{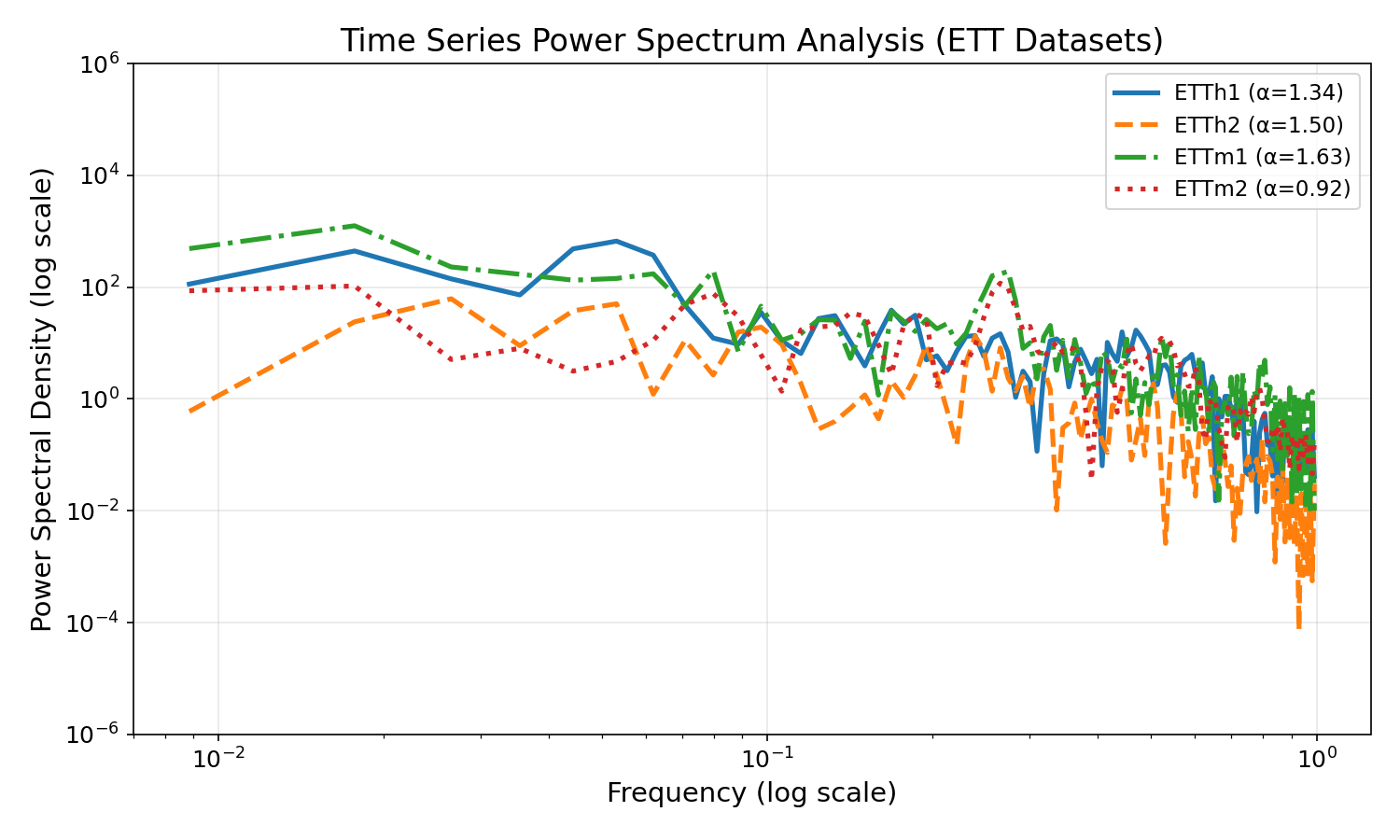}}
    \subfigure[Natural Image (ImageNet-100)]{\label{fig:pss_img}\includegraphics[width=0.48\textwidth]{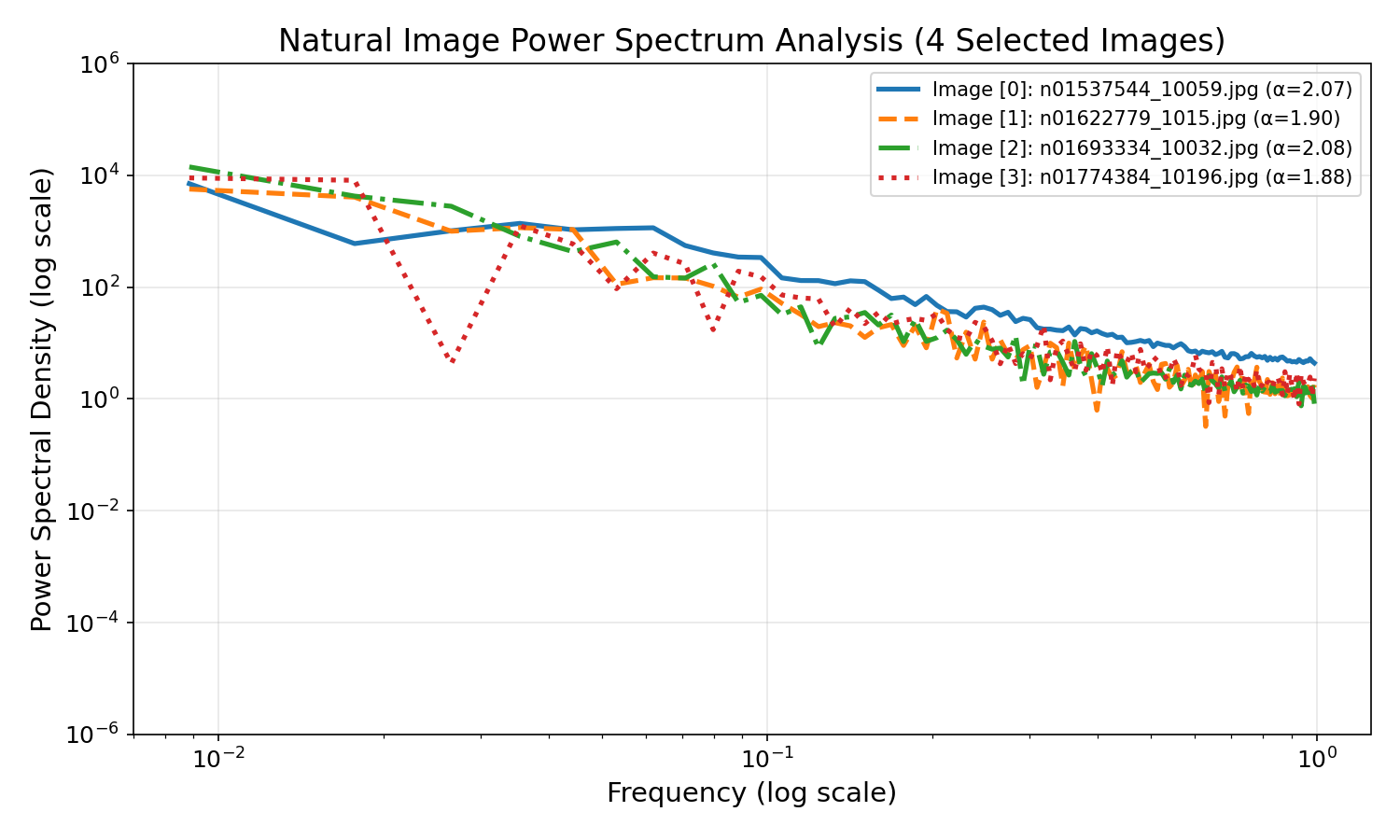}}
    \subfigure[Text (Wikipedia)]{\label{fig:pss_txt}\includegraphics[width=0.48\textwidth]{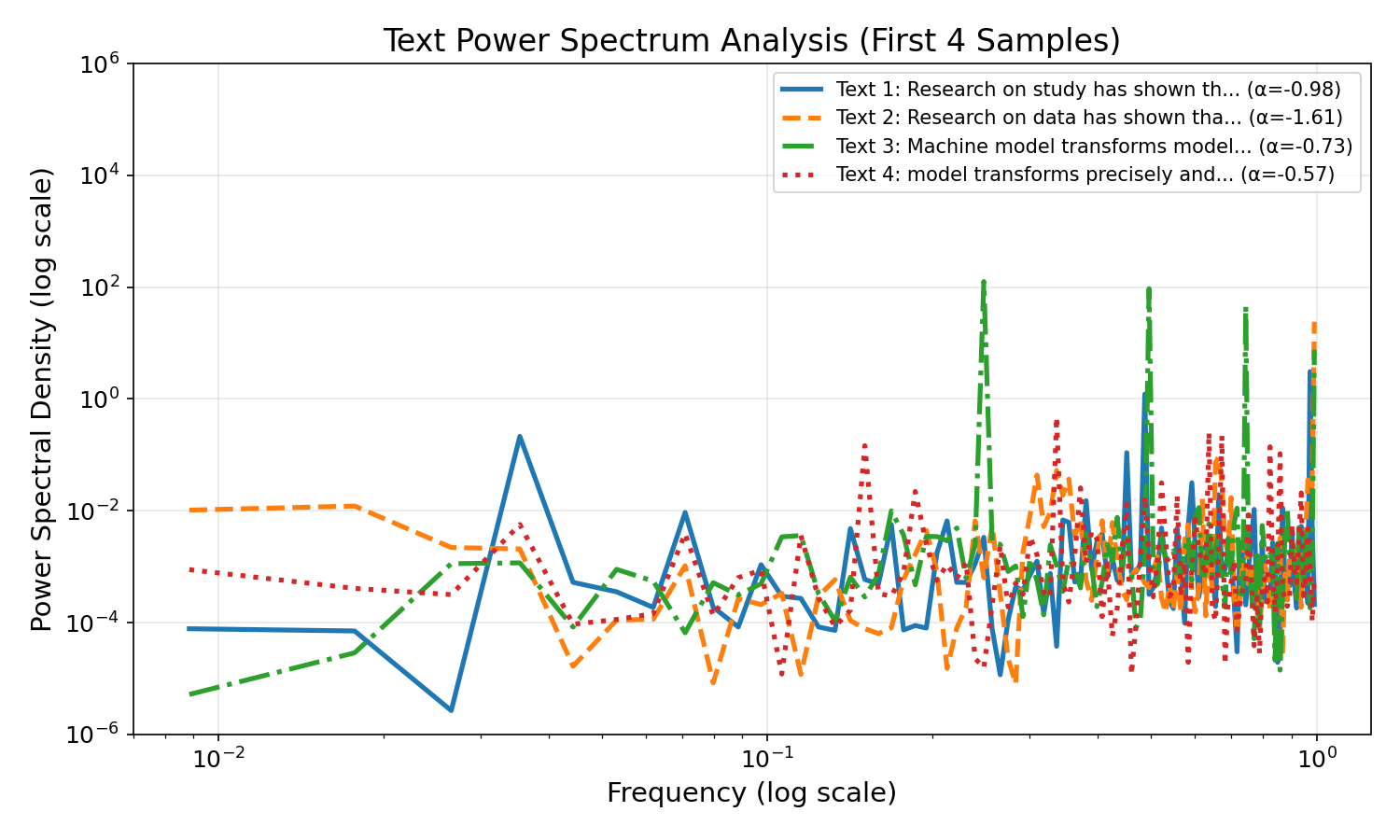}}
    \subfigure[Three Modality Comparison (Averaged)]{\label{fig:pss_avg}\includegraphics[width=0.48\textwidth]{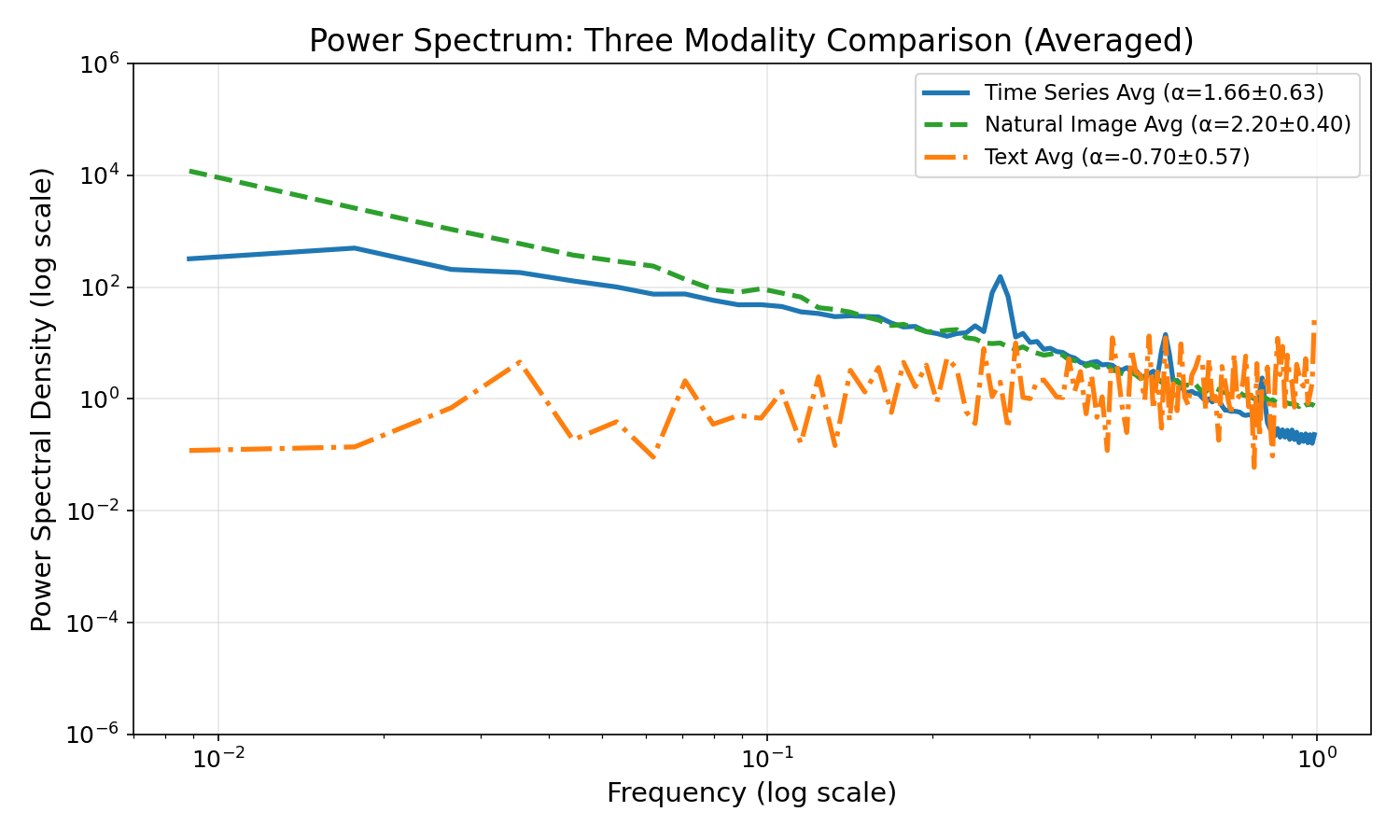}}
    \caption{Power spectral slope analysis for different modalities. Subfigures (a)-(c) show individual sample curves for time series, natural images, and text respectively. Subfigure (d) displays the averaged power spectral slope.}
    \label{fig:pss_examples}
\end{figure}

To provide intuitive understanding of the modality gap, Figure \ref{fig:pss_examples} presents representative power spectral density curves for each modality.
Table \ref{tab:power_law_results_merged} summarizes the power law fitting results for all three modalities, including data source names, sample counts, and mean exponent values with standard deviations.

\begin{table}[h]
    \centering
    \caption{Power law fitting results for different modalities. $\alpha$ denotes the fitted power law exponent, $\sigma_\alpha$ is the standard deviation.}
    \label{tab:power_law_results_merged}
    \begin{tabular}{lcccc}
        \hline
        \textbf{Modality} & \textbf{Source} & \textbf{Samples} & $\alpha \pm \sigma_\alpha$\\
        \hline
        Natural Image & ImageNet-100 & 100 & $2.19 \pm 0.40$\\
        \hline
        \multirow{4}{*}{Time Series} & ETTh1 & 100 & $1.37 \pm 0.73$\\
        & ETTh2 & 100 & $1.98 \pm 0.64$\\
        & ETTm1 & 100 & $1.57 \pm 0.53$\\
        & ETTm2 & 100 & $1.62 \pm 0.49$\\
        \cline{1-1}
        \multicolumn{1}{c}{\textbf{Average}} & \textbf{All subsets} & \textbf{400} & \textbf{$1.66 \pm 0.63$}\\
        \hline
        Text (ASCII) & Wikipedia & 100 & $-0.70 \pm 0.57$\\
        \hline
    \end{tabular}
\end{table}

The results reveal distinct spectral characteristics across modalities. Natural images exhibit the expected $1/f^2$ scaling ($\alpha \approx 2.19$), while time series rendered images show a significantly lower exponent ($\alpha = 1.66$), indicating a modality gap of $\Delta\alpha \approx 0.53$. This discrepancy arises from artificial periodicities introduced by the folding operation. Text modality shows the most distinct behavior with negative exponent ($\alpha = -0.70$), reflecting the high-frequency nature of ASCII-encoded character sequences.

\subsection{Discussion}

\begin{figure}[t]
    \centering
    \subfigure[Periodicity=24 ($\alpha=1.34$)]{\label{fig:pss_etth1_p24}\includegraphics[width=0.32\textwidth]{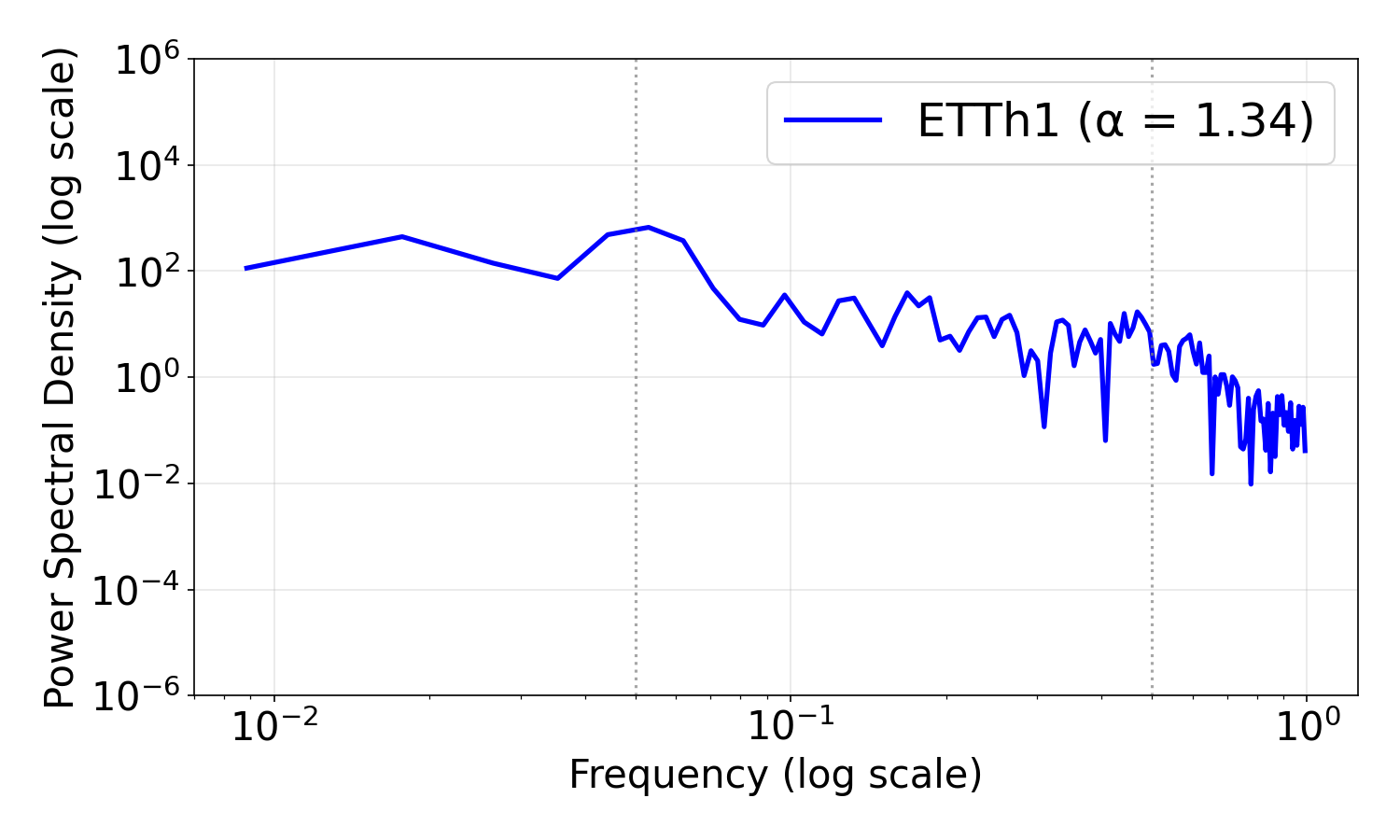}}
    \subfigure[Periodicity=28 ($\alpha=1.68$)]{\label{fig:pss_etth1_p28}\includegraphics[width=0.32\textwidth]{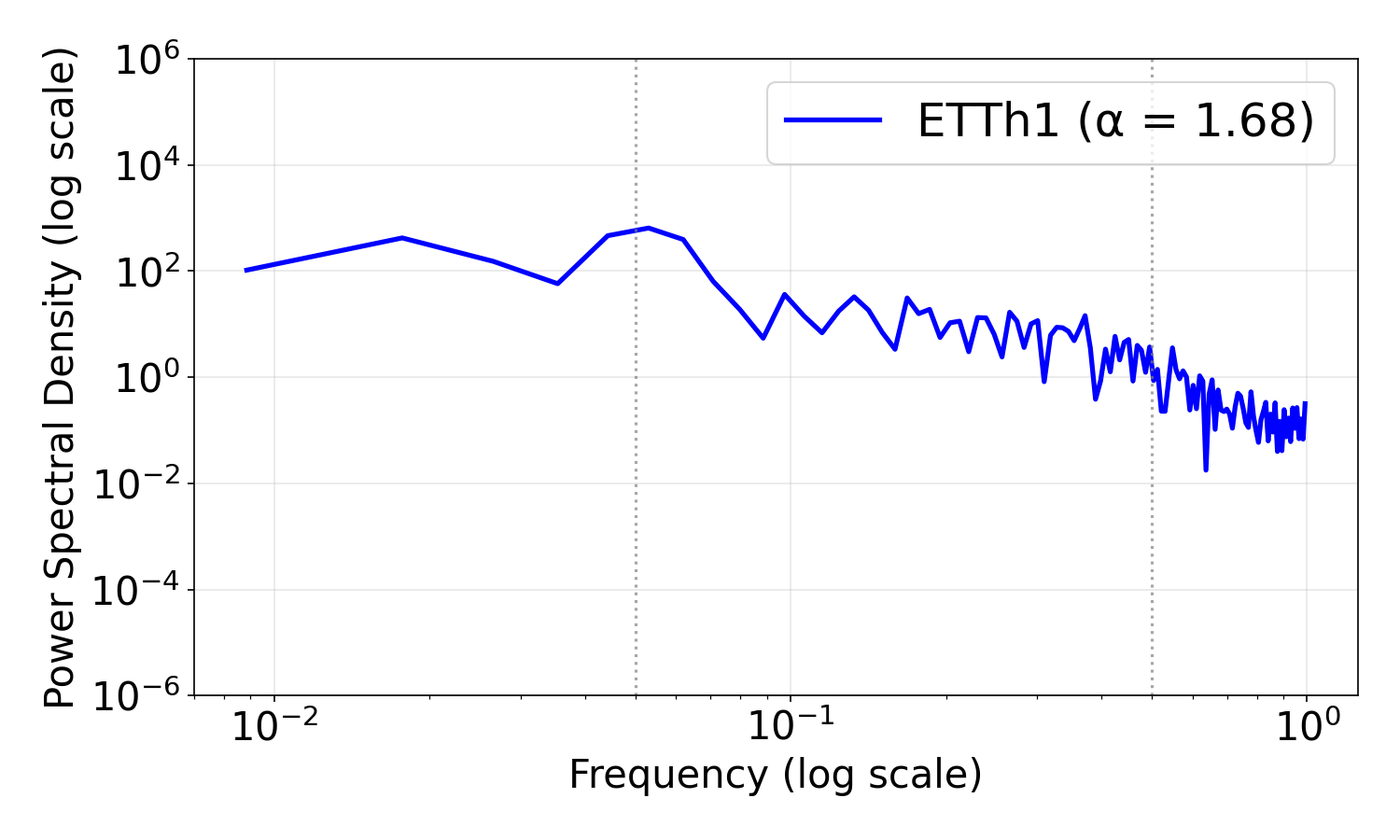}}
    \subfigure[Periodicity=30 ($\alpha=2.43$)]{\label{fig:pss_etth1_p30}\includegraphics[width=0.32\textwidth]{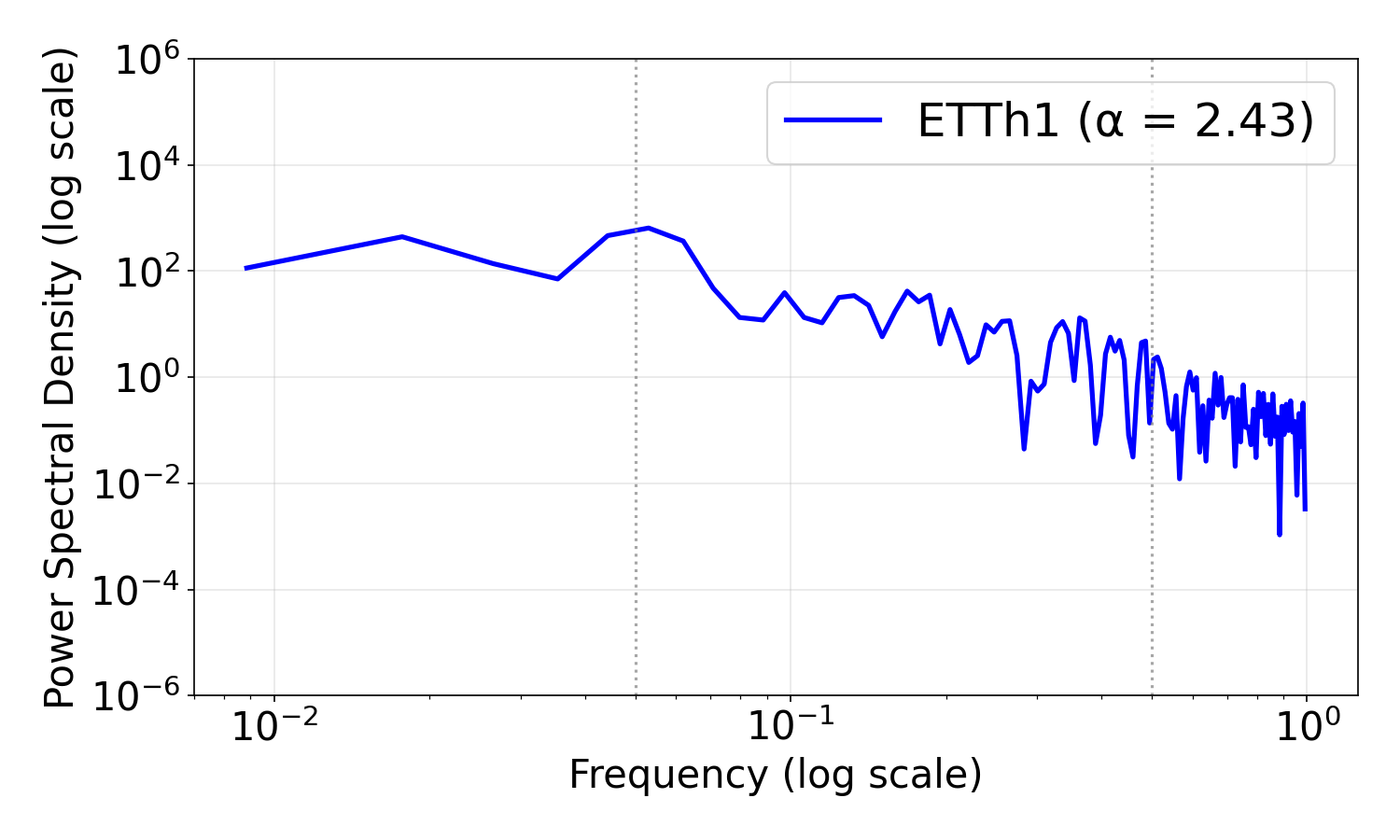}}
    \subfigure[Periodicity=32 ($\alpha=2.62$)]{\label{fig:pss_etth1_p32}\includegraphics[width=0.32\textwidth]{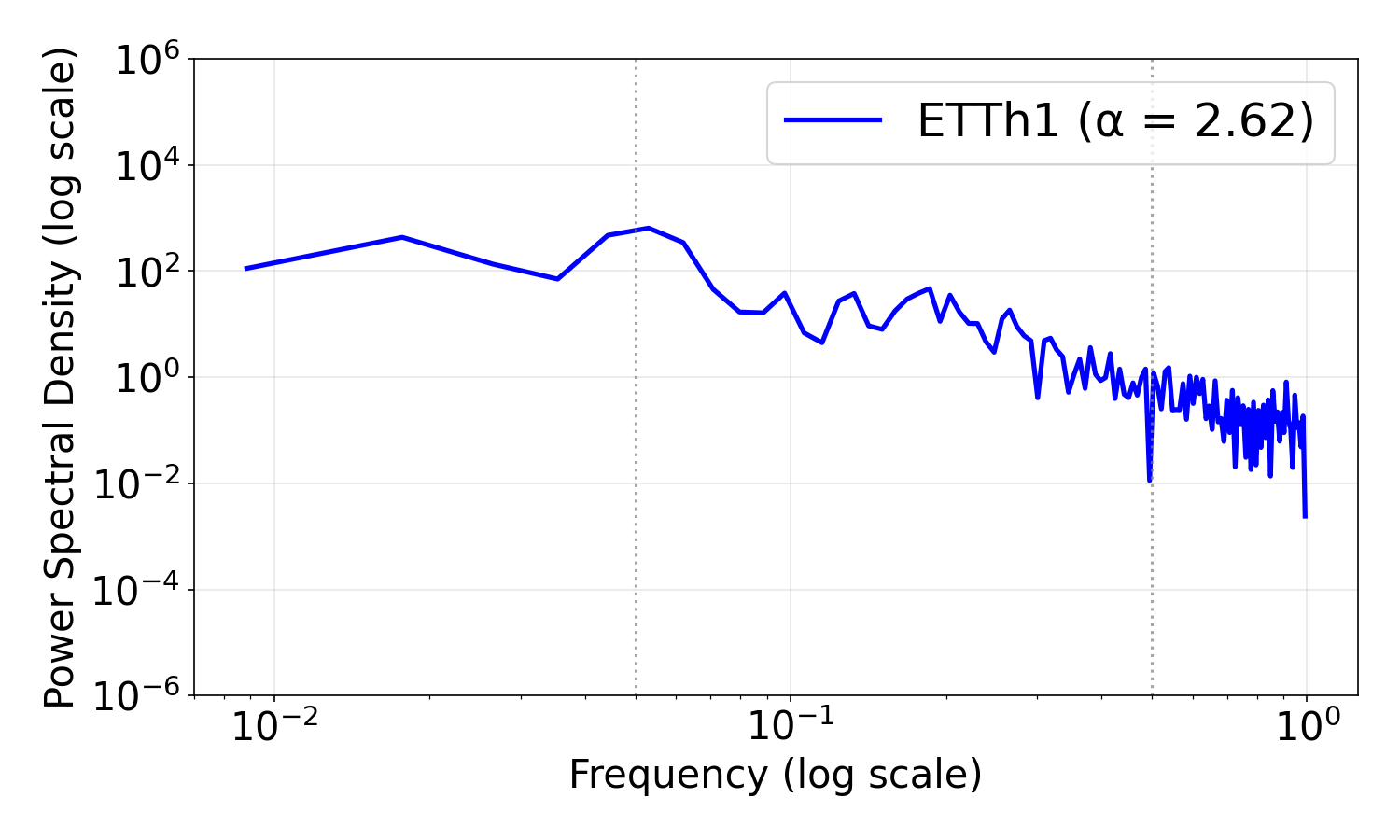}}
    \subfigure[Periodicity=48 ($\alpha=2.85$)]{\label{fig:pss_etth1_p48}\includegraphics[width=0.32\textwidth]{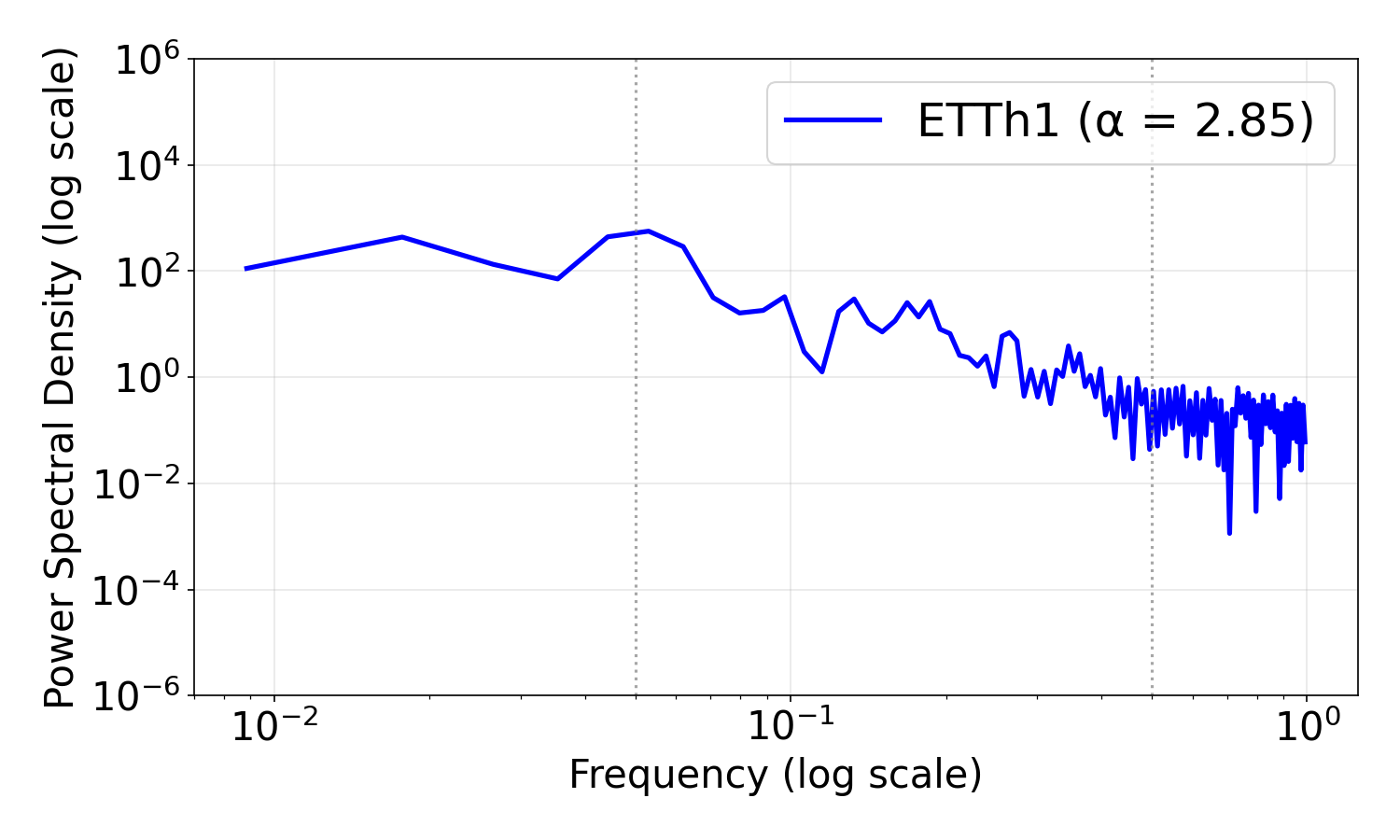}}
    \subfigure[Periodicity=64 ($\alpha=2.86$)]{\label{fig:pss_etth1_p64}\includegraphics[width=0.32\textwidth]{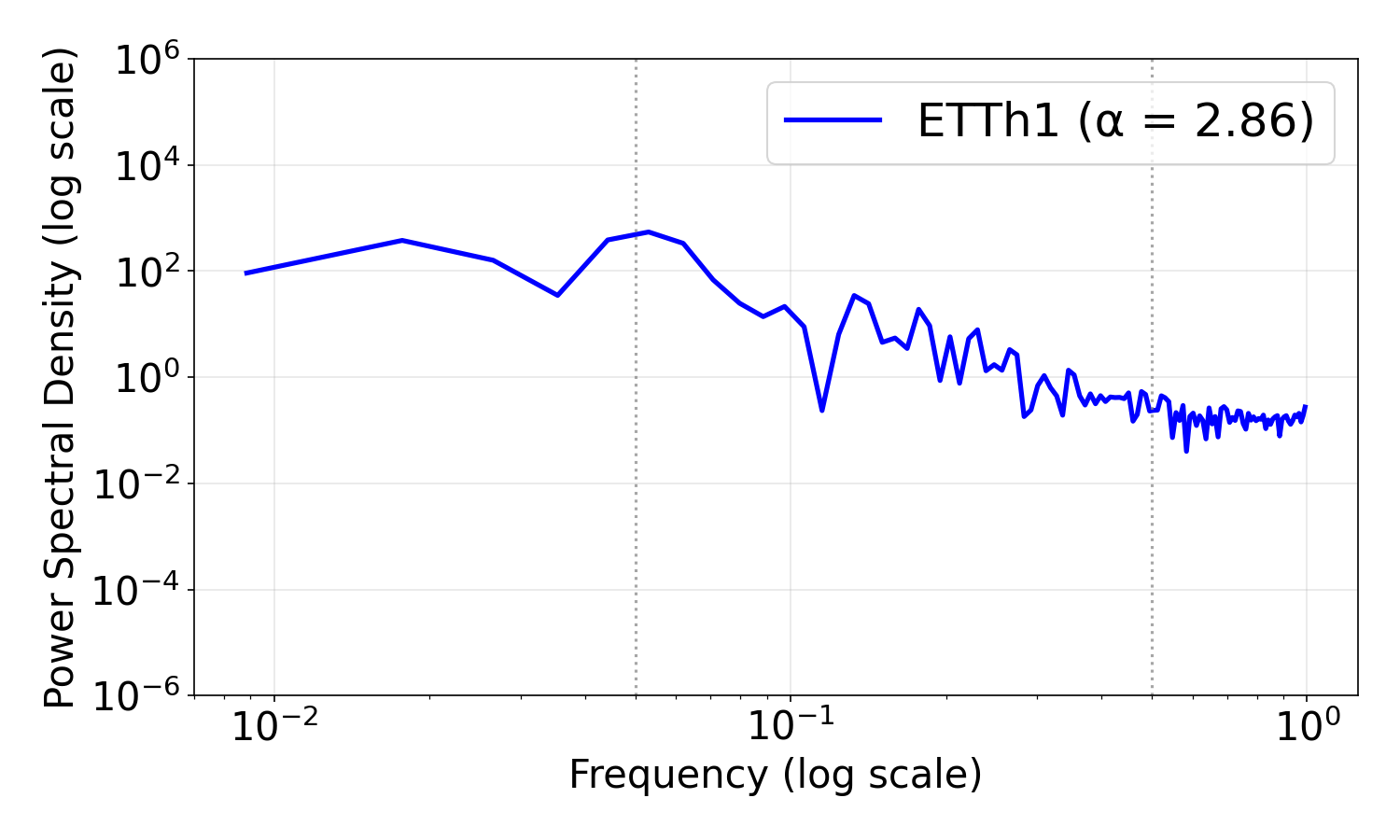}}
    \caption{Power spectral slope analysis for time series data rendered with different periodicity values. Subfigures (a)-(f) show the power spectral density (PSD) of ETTh1 data when rendered to image representation with periodicity values of 24, 28, 30, 32, 48, and 64 respectively.}
    \label{fig:pss_periodicity_analysis}
\end{figure}

As illustrated in Figure\ref{fig:pss_periodicity_analysis}, during our experiments, we observe that when converting one-dimensional time series into two-dimensional visual representations, the PSS is significantly influenced by the periodicity parameter. Specifically, the PSS $\alpha$ of time series images increases monotonically with periodicity. 
This is because periodicity determines how temporal information is spatially distributed in the image domain. Larger periodicity values compress more time steps into each row, this spatial compression amplifies high-frequency components in the visual representation, resulting in steeper spectral decay (larger $\alpha$). Conversely, smaller periodicity preserves finer temporal structures, yielding gentler spectral slopes (smaller $\alpha$).

\begin{table}[b]
\centering
\caption{Ablation study on periodicity.}
\label{tab:periodicity_ablation}
\begin{tabular}{ccccccc}
\toprule
Periodicity & 24 & 28 & 30 & 32 & 48 & 64 \\
\midrule
$\alpha$ & 1.34 & 1.68 & 2.43 & 2.62 & 2.85 & 2.86 \\
MSE       & \textcolor{red}{0.343} & 0.455 & 0.477 & 0.451 & \textcolor{blue}{0.391} & 0.444 \\
\bottomrule
\end{tabular}
\end{table}

Table\ref{tab:periodicity_ablation} demonstrates that PSS does not necessarily correlate with prediction performance, thus we do not recommend using PSS as an evaluation metric for forecasting models. In this paper, PSS serves as an analytical tool to characterize the frequency distributions of time series images and natural images, enabling frequency-domain analysis and alignment of the two modalities. Our objective is to enhance the visual representations of time series data by exploiting the underlying patterns that emerge across different modalities~\cite{dmmv2025}. Subsequently, we feed the time series images with stronger visual features into a frozen MAE pretrained on natural images, fully leveraging the knowledge learned from natural image representations, which provides more robust visual-view characteristics and patterns.

In summary, PSS provides a means to quantify the frequency-domain discrepancies between modalities, facilitating the distinction and alignment of time series and natural image distributions. However, it is crucial to avoid excessive modification of the original data during frequency redistribution, as this may introduce unnecessary noise or lead to information loss. Rather than viewing PSS as a performance metric, we advocate treating it as a novel analytical perspective for characterizing modality gaps. This approach offers a novel research direction for future work on multimodal learning and modality fusion.

\section{Implementation Details}

\subsection{Data Processing Details}

To bridge the gap between time series data and vision model inputs, SSDA implements a bidirectional transformation pipeline that converts temporal sequences into 2D image representations and recovers temporal predictions from vision model outputs. This reshape-reconstruct mechanism preserves temporal structure while enabling the use of spatial reasoning capabilities in vision models.

\subsubsection{Reshape}

The reshape process transforms time series data $\mathbf{X} \in \mathbb{R}^{B \times T \times N}$ into masked image inputs suitable for MAE processing. The transformation involves several coordinated steps that convert temporal sequences into spatial representations while maintaining temporal relationships.

\begin{algorithm}[h]
    \SetAlgoLined
    \KwData{Time series data $\mathbf{X} \in \mathbb{R}^{B \times T \times N}$, batch size $B$, sequence length $T$, number of variables $N$, periodicity $p$, padding length $p_l$, target image dimensions $H_\text{target} \times W_\text{target}$.}
    \KwResult{Masked image input $\mathbf{I}_\text{input} \in \mathbb{R}^{(B \cdot N) \times 3 \times H \times (W + W_\text{mask})}$.}
        $\mathbf{X}_\text{enc} \leftarrow \text{rearrange}(\mathbf{X}, \text{'b t n $\rightarrow$ b n t'})$\;
        $\mathbf{X}_\text{pad} \leftarrow \text{pad}(\mathbf{X}_\text{enc}, (p_l, 0), \text{mode}=\text{'replicate'})$\;
        $\mathbf{X}_{2d} \leftarrow \text{rearrange}(\mathbf{X}_\text{pad}, \text{'b n (p f) $\rightarrow$ (b n) 1 f p'})$\;
        $\mathbf{X}_\text{resize} \leftarrow \text{Resize}(\mathbf{X}_{2d}, (H_\text{target}, W_\text{target}))$\;
        \If{spectral branch}{
        $\mathbf{X}_{3c} \leftarrow \text{Enhance}(\mathbf{X}_\text{resize})$\;
        }
        \Else{
        $\mathbf{X}_{3c} \leftarrow \text{repeat}(\mathbf{X}_\text{resize}, \text{channels}=3)$\;
        }
        $\mathbf{I}_\text{input} \leftarrow \text{concat}(\mathbf{X}_{3c}, \mathbf{M}_\text{masked}, \dim=-1)$\;
    \caption{Time Series to Image Reshaping}
    \label{alg:ts_to_image}
\end{algorithm}

As illustrated in the Algorithm~\ref{alg:ts_to_image}, the process begins by rearranging the time series to group variables together, then applies left padding for periodicity alignment. The padded sequences are converted into 2D spatial representations where height corresponds to the number of periods and width corresponds to the periodicity parameter. These representations are resized to match vision model requirements, expanded to three channels, and concatenated with a masked region representing the prediction window.

\subsubsection{Reconstruct}

The reconstruction process reverses the reshaping operation to recover temporal predictions from MAE outputs $\mathbf{Z} \in \mathbb{R}^{(B \cdot N) \times L \times D}$.

As illustrated in the Algorithm~\ref{alg:image_to_ts}, starting with the latent representations from the MAE decoder, patch tokens are converted back to spatial images through unpatchify operations. The RGB reconstructions are converted to grayscale, resized to restore temporal dimensions, flattened back to temporal sequences, and finally the prediction window is extracted from the reconstructed output.

\begin{algorithm}[h]
\SetAlgoLined
\KwData{MAE decoder output $\mathbf{Z} \in \mathbb{R}^{(B \cdot N) \times L \times D}$, batch size $B$, number of variables $N$, patch sequence length $L$, embedding dimension $D$, context length $T_\text{input}$, prediction length $T_\text{pred}$.}
\KwResult{Temporal predictions $\mathbf{Y}_\text{pred} \in \mathbb{R}^{B \times T_\text{pred} \times N}$.}
$\mathbf{I}_\text{recon} \leftarrow \text{unpatchify}(\mathbf{Z})$\;
$\mathbf{I}_\text{gray} \leftarrow \frac{1}{3} \sum{c=1}^3 \mathbf{I}_\text{recon}[:,:,c,:,:]$\;
$\mathbf{I}_\text{temp} \leftarrow \text{Resize}(\mathbf{I}_\text{gray}, (p, f \cdot p))$\;
$\mathbf{Y} \leftarrow \text{rearrange}(\mathbf{I}_\text{temp}, \text{'(b n) 1 f p $\rightarrow$ b (p f) n'})$\;
$\mathbf{Y}_\text{pred} \leftarrow \mathbf{Y}[:, T_\text{input}:T_\text{input} + T_\text{pred}, :]$\;
\caption{Image to Time Series Reconstruction}
\label{alg:image_to_ts}
\end{algorithm}

\subsection{Training Configuration}

\subsubsection{Model architecture details}
\label{sec:model_arch_params}

We summarize the default model architecture parameters in Table~\ref{tab:model_arch_params}. Image representations are set to a size of $224 \times 224$, which conforms to the input requirements of MAE. The model backbone follows a standard Transformer-based design, where the hidden dimension $d\_model$ is set to $512$ and the feed-forward dimension $d\_ff$ is $2048$ to ensure sufficient model capacity. The multi-head attention module uses $8$ attention heads, while the encoder and decoder are composed of $2$ and $1$ layers, respectively, balancing performance and computational efficiency. A dropout rate of $0.1$ is applied to mitigate overfitting during training. In addition, we employ $5$ workers for data loading to accelerate the training pipeline.

\begin{table}[h]
\centering
\caption{Default Model Architecture Parameters}
\label{tab:model_arch_params}
\footnotesize
\setlength{\tabcolsep}{6pt}
\renewcommand{\arraystretch}{1.1}
\begin{tabular}{l|c|l}
\toprule
\textbf{Parameter} & \textbf{Default Value} & \textbf{Description} \\
\midrule
image\_size & 224 & Size of reshaped image representation \\
\midrule
d\_model & 512 & Dimension of hidden embeddings \\
\midrule
d\_ff & 2048 & Dimension of feed-forward network \\
\midrule
n\_heads & 8 & Number of attention heads \\
\midrule
e\_layers & 2 & Number of encoder layers \\
\midrule
d\_layers & 1 & Number of decoder layers \\
\midrule
dropout & 0.1 & Dropout rate \\
\midrule
num\_workers & 5 & Number of data loader workers \\
\bottomrule
\end{tabular}
\end{table}

\subsubsection{Training Hyperparameters}
\label{sec:training_params}
We summarize the default training parameters in Table~\ref{tab:training_params}. 
We use the Adam optimizer with a learning rate of $2\times10^{-6}$, a batch size of $32$, and train for $10$ epochs, with early stopping ($patience=3$) and MSE as the loss function. The input sequence length ($seq\_len$) is set according to dataset characteristics to capture long-range dependencies, while the prediction length ($pred\_len$) is evaluated over multiple horizons ($96, 192, 336, 720$). The output dimension ($c\_out$) is dataset-specific. For the visual representation, periodicity controls the reshaping of time series into 2D image representations (e.g., $24$ for hourly data and $96$ for $15$-minute data), and bilinear interpolation is used for resizing. The normalization and alignment constants are both set to $0.4$ to stabilize representation transformation. We adopt LoRA for parameter-efficient fine-tuning with rank $r=4$, scaling factor $16$, and dropout $0.1$, and introduce a residual weight of $0.05$ for frequency enhancement. The vision backbone is initialized from a pre-trained MAE model. Unless otherwise specified, all parameters are kept consistent across datasets for fair comparison.

\begin{table}[h]
\centering
\caption{Default Training Parameters}
\label{tab:training_params}
\footnotesize
\setlength{\tabcolsep}{6pt}
\renewcommand{\arraystretch}{1.1}
\begin{tabular}{l|c|l}
\toprule
\textbf{Parameter} & \textbf{Default Value} & \textbf{Description} \\
\midrule
batch\_size & 32 & Training batch size \\
\midrule
learning\_rate & 2e-6 & Initial learning rate \\
\midrule
training\_epochs & 10 & Number of training epochs \\
\midrule
patience & 3 & Early stopping patience \\
\midrule
loss & MSE & Mean square error loss \\
\midrule
seq\_len & 
\begin{tabular}[c]{@{}c@{}}
1440 (ETTh1/ETTh2/ETTm1/ETTm2) \\
2304 (Electricity/Traffic) \\
2360 (Weather)
\end{tabular}
& Input sequence length \\
\midrule
pred\_len & 96/192/336/720 & Prediction length \\
\midrule
c\_out &
\begin{tabular}[c]{@{}c@{}}
7 (ETTh1/ETTh2/ETTm1/ETTm2) \\
21 (Weather) \\
321 (Electricity) \\
862 (Traffic)
\end{tabular}
& Output dimension (dataset-specific) \\
\midrule
periodicity &
\begin{tabular}[c]{@{}c@{}}
24 (ETTh1/ETTh2/Electricity/Traffic) \\
96 (ETTm1/ETTm2) \\
144 (Weather)
\end{tabular}
& Periodicity for time-series to image conversion \\
\midrule
interpolation & bilinear & Image resizing method \\
\midrule
norm\_const & 0.4 & Normalization coefficient \\
\midrule
align\_const & 0.4 & Alignment scaling factor \\
\midrule
LoRA\_rank (r) & 4 & Rank of low-rank adaptation \\
\midrule
LoRA\_alpha & 16 & Scaling factor of LoRA \\
\midrule
LoRA\_dropout & 0.1 & Dropout in LoRA layers \\
\midrule
residual\_weight & 0.05 & Weight of frequency enhancement residual \\
\midrule
vision\_backbone & mae\_base & Vision model architecture \\

\bottomrule
\end{tabular}
\end{table}

\subsection{Running Environment}
\label{sec:implementation}

All experiments are conducted on a workstation equipped with 8 NVIDIA GeForce RTX 4090 GPUs (24GB memory each), running CUDA 12.1 and cuDNN 8.9. The model is implemented in Python 3.10 using PyTorch 2.2.2, with key dependencies including NumPy 1.23.5, pandas 1.5.3, and torchvision 0.17.2. More training details can be found in Sections~\ref{sec:model_arch_params} and~\ref{sec:training_params}.

\section{Benchmark and Baseline}
\subsection{Benchmark}
Following prior work~\cite{visionts2025,timevlm2025,dmmv2025}, we evaluate our method on seven widely used long-term time series forecasting (LTSF) benchmarks, covering diverse sampling frequencies, variate dimensions, periodic patterns, and real-world scenarios.

The benchmarks include four Electricity Transformer Temperature (ETT) datasets (ETTh1, ETTh2, ETTm1, ETTm2), which record oil temperature from electric transformers with hourly and 15-minute sampling frequencies~\cite{informer2021}. In addition, the Weather dataset contains 10-minute resolution meteorological measurements~\cite{autoformer2022}. The Traffic dataset collects hourly road occupancy rates from freeway sensors~\cite{autoformer2022}, and the Electricity dataset provides hourly electricity consumption data~\cite{Trindade2015}.



\subsection{Baselines}

\subparagraph{Baselines}
We compare our method with a diverse set of representative baselines spanning vision-based, large-model-based, and classical time series forecasting approaches, ensuring a comprehensive evaluation.

1. \textbf{DMMV}~\cite{dmmv2025} introduces a dual-modality modeling framework that jointly leverages visual representations and temporal dynamics for time series forecasting, demonstrating the effectiveness of multimodal fusion.

2. \textbf{Time-VLM}~\cite{timevlm2025} adapts vision-language models to time series tasks by encoding temporal data into visual formats and leveraging pre-trained multimodal representations for forecasting.

3. \textbf{VisionTS}~\cite{visionts2025} converts time series into images and directly applies large vision models (LVMs), showcasing the potential of transferring visual knowledge to time series forecasting.

4. \textbf{Time-LLM}~\cite{timellm2024} reformulates time series forecasting as a language modeling problem, enabling large language models (LLMs) to process temporal data through prompt-based learning.

5. \textbf{GPT4TS}~\cite{zhou2023} extends the idea of LLM-based forecasting by further adapting GPT-style architectures to capture temporal dependencies via sequence modeling and instruction tuning.

6. \textbf{TimeCMA}~\cite{timecma2025} proposes a cross-modality alignment mechanism to retrieve disentangled time series embeddings from LLM-based prompts via channel-wise similarity, enhancing robustness in multivariate forecasting.

7. \textbf{PatchTST}~\cite{patchtst2023} adopts a patch-based Transformer architecture, where time series are segmented into patches to improve long-range dependency modeling and computational efficiency.

8. \textbf{TimesNet}~\cite{timesnet2023} proposes a temporal 2D-variation modeling paradigm by transforming time series into multiple periodic representations, effectively capturing complex temporal patterns.

9. \textbf{FEDformer}~\cite{fedformer2022} introduces frequency-enhanced decomposition with attention mechanisms, combining Fourier analysis with Transformer architectures to better model long-term dependencies.

10. \textbf{Informer}~\cite{informer2021} is a Transformer-based model designed for long sequence forecasting, utilizing ProbSparse self-attention to reduce computational complexity while preserving performance.

11. \textbf{DLinear}~\cite{dlinear2022} is a lightweight linear model that decomposes time series into trend and seasonal components, serving as a strong and efficient baseline despite its simplicity.

\section{Full Experimental Results}
\subsection{Few-Shot Forecasting}
\label{apd:few_shot}

\begin{table}[H]
\centering
\caption{\small Full few-shot learning results on 10\% training data with forecasting horizons $H \in \{96, 192, 336, 720\}$. Lower values indicate better performance. \textcolor{red}{Red} indicates the best performance, and \textcolor{blue}{Blue} indicates the second best.}
\label{tab:full10_results}
\renewcommand{\arraystretch}{0.95}
\resizebox{\textwidth}{!}{
\begin{tabular}{ll|cc|cc|cc|cc|cc|cc|cc|cc|cc|cc}
\toprule

\multicolumn{2}{c}{Models} &
\multicolumn{2}{c}{SSDA} &
\multicolumn{2}{c}{Time-VLM} &
\multicolumn{2}{c}{VisionTS} &
\multicolumn{2}{c}{Time-LLM} &
\multicolumn{2}{c}{GPT4TS} &
\multicolumn{2}{c}{PatchTST} &
\multicolumn{2}{c}{TimesNet} &
\multicolumn{2}{c}{FEDformer} &
\multicolumn{2}{c}{Informer} &
\multicolumn{2}{c}{DLinear} \\

\cmidrule(lr){3-4}\cmidrule(lr){5-6}\cmidrule(lr){7-8}\cmidrule(lr){9-10}
\cmidrule(lr){11-12}\cmidrule(lr){13-14}\cmidrule(lr){15-16}
\cmidrule(lr){17-18}\cmidrule(lr){19-20}\cmidrule(lr){21-22}

\multicolumn{2}{c}{Metrics} &
MSE & MAE & MSE & MAE & MSE & MAE & MSE & MAE &
MSE & MAE & MSE & MAE & MSE & MAE & MSE & MAE &
MSE & MAE & MSE & MAE \\
\midrule

\multirow{5}{*}{\rotatebox{90}{ETTh1}}
& 96  & \textcolor{red}{0.359} & \textcolor{red}{0.376} & 0.391 & 0.404 & \textcolor{blue}{0.365} & \textcolor{blue}{0.376} & 0.448 & 0.460 & 0.458 & 0.456 & 0.516 & 0.485 & 0.861 & 0.628 & 0.512 & 0.499 & 1.179 & 0.792 & 0.492 & 0.495 \\
& 192 & \textcolor{red}{0.389} & \textcolor{red}{0.403} & 0.420 & 0.431 & \textcolor{blue}{0.407} & \textcolor{blue}{0.404} & 0.484 & 0.483 & 0.570 & 0.516 & 0.598 & 0.524 & 0.797 & 0.593 & 0.624 & 0.555 & 1.199 & 0.806 & 0.565 & 0.538 \\
& 336 & \textcolor{red}{0.417} & \textcolor{red}{0.420} & 0.439 & 0.448 & \textcolor{blue}{0.438} & \textcolor{blue}{0.426} & 0.589 & 0.540 & 0.608 & 0.535 & 0.657 & 0.550 & 0.941 & 0.648 & 0.691 & 0.574 & 1.202 & 0.811 & 0.721 & 0.622 \\
& 720 & \textcolor{red}{0.418} & \textcolor{blue}{0.442} & 0.476 & 0.484 & \textcolor{blue}{0.426} & \textcolor{red}{0.440} & 0.700 & 0.604 & 0.725 & 0.591 & 0.762 & 0.610 & 0.877 & 0.641 & 0.728 & 0.614 & 1.217 & 0.825 & 0.986 & 0.743 \\
& Avg & \textcolor{red}{0.396} & \textcolor{red}{0.410} & 0.431 & 0.442 & \textcolor{blue}{0.409} & \textcolor{blue}{0.412} & 0.556 & 0.522 & 0.590 & 0.525 & 0.633 & 0.542 & 0.869 & 0.628 & 0.639 & 0.561 & 1.199 & 0.809 & 0.691 & 0.600 \\
\midrule

\multirow{5}{*}{\rotatebox{90}{ETTh2}}
& 96  & \textcolor{red}{0.263} & \textcolor{red}{0.325} & 0.284 & 0.347 & 0.294 & 0.340 & \textcolor{blue}{0.275} & \textcolor{blue}{0.326} & 0.331 & 0.374 & 0.353 & 0.389 & 0.378 & 0.409 & 0.382 & 0.416 & 3.837 & 1.508 & 0.357 & 0.411 \\
& 192 & \textcolor{red}{0.319} & \textcolor{red}{0.367} & \textcolor{blue}{0.349} & 0.398 & 0.355 & 0.383 & 0.374 & \textcolor{blue}{0.373} & 0.402 & 0.411 & 0.403 & 0.414 & 0.490 & 0.467 & 0.478 & 0.474 & 3.856 & 1.513 & 0.569 & 0.519 \\
& 336 & \textcolor{red}{0.346} & \textcolor{red}{0.386} & \textcolor{blue}{0.370} & 0.412 & 0.374 & \textcolor{blue}{0.405} & 0.406 & 0.429 & 0.406 & 0.433 & 0.426 & 0.441 & 0.537 & 0.494 & 0.504 & 0.501 & 3.952 & 1.526 & 0.671 & 0.572 \\
& 720 & \textcolor{red}{0.368} & \textcolor{red}{0.412} & 0.441 & 0.466 & \textcolor{blue}{0.390} & \textcolor{blue}{0.427} & 0.427 & 0.449 & 0.449 & 0.464 & 0.477 & 0.480 & 0.510 & 0.491 & 0.499 & 0.509 & 3.842 & 1.503 & 0.824 & 0.648 \\
& Avg & \textcolor{red}{0.324} & \textcolor{red}{0.373} & 0.361 & 0.405 & \textcolor{blue}{0.353} & \textcolor{blue}{0.389} & 0.370 & 0.394 & 0.397 & 0.421 & 0.415 & 0.431 & 0.479 & 0.465 & 0.466 & 0.475 & 3.872 & 1.513 & 0.605 & 0.538 \\
\midrule

\multirow{5}{*}{\rotatebox{90}{ETTm1}}
& 96  & \textcolor{blue}{0.323} & \textcolor{red}{0.341} & \textcolor{red}{0.310} & \textcolor{blue}{0.354} & 0.444 & 0.428 & 0.346 & 0.388 & 0.390 & 0.404 & 0.410 & 0.419 & 0.583 & 0.501 & 0.578 & 0.518 & 1.162 & 0.785 & 0.352 & 0.392 \\
& 192 & \textcolor{red}{0.340} & \textcolor{red}{0.355} & \textcolor{blue}{0.340} & \textcolor{blue}{0.370} & 0.502 & 0.457 & 0.373 & 0.416 & 0.429 & 0.423 & 0.437 & 0.434 & 0.630 & 0.528 & 0.617 & 0.546 & 1.172 & 0.793 & 0.382 & 0.412 \\
& 336 & \textcolor{red}{0.364} & \textcolor{red}{0.371} & \textcolor{blue}{0.369} & \textcolor{blue}{0.387} & 0.665 & 0.547 & 0.413 & 0.426 & 0.469 & 0.439 & 0.476 & 0.454 & 0.725 & 0.568 & 0.998 & 0.775 & 1.227 & 0.908 & 0.419 & 0.434 \\
& 720 & \textcolor{red}{0.414} & \textcolor{red}{0.409} & \textcolor{blue}{0.423} & \textcolor{blue}{0.417} & 0.767 & 0.601 & 0.485 & 0.476 & 0.569 & 0.498 & 0.681 & 0.556 & 0.769 & 0.549 & 0.693 & 0.579 & 1.207 & 0.797 & 0.490 & 0.477 \\
& Avg & \textcolor{red}{0.360} & \textcolor{red}{0.369} & \textcolor{blue}{0.360} & \textcolor{blue}{0.382} & 0.595 & 0.508 & 0.404 & 0.427 & 0.464 & 0.441 & 0.501 & 0.466 & 0.677 & 0.537 & 0.722 & 0.605 & 1.192 & 0.821 & 0.411 & 0.429 \\
\midrule

\multirow{5}{*}{\rotatebox{90}{ETTm2}}
& 96  & 0.209 & 0.276 & \textcolor{red}{0.169} & \textcolor{red}{0.260} & 0.192 & 0.274 & \textcolor{blue}{0.177} & \textcolor{blue}{0.261} & 0.188 & 0.269 & 0.191 & 0.274 & 0.212 & 0.285 & 0.291 & 0.399 & 3.203 & 1.407 & 0.213 & 0.303 \\
& 192 & 0.242 & \textcolor{blue}{0.300} & \textcolor{red}{0.222} & \textcolor{red}{0.296} & 0.243 & 0.308 & \textcolor{blue}{0.241} & 0.314 & 0.251 & 0.309 & 0.252 & 0.317 & 0.270 & 0.323 & 0.307 & 0.379 & 3.112 & 1.387 & 0.278 & 0.345 \\
& 336 & 0.283 & \textcolor{blue}{0.328} & \textcolor{blue}{0.278} & 0.335 & 0.299 & 0.347 & \textcolor{red}{0.274} & \textcolor{red}{0.327} & 0.307 & 0.346 & 0.306 & 0.353 & 0.323 & 0.353 & 0.543 & 0.559 & 3.255 & 1.421 & 0.338 & 0.385 \\
& 720 & \textcolor{red}{0.346} & \textcolor{red}{0.375} & \textcolor{blue}{0.381} & \textcolor{blue}{0.401} & 0.433 & 0.439 & 0.417 & 0.390 & 0.426 & 0.417 & 0.433 & 0.427 & 0.474 & 0.449 & 0.712 & 0.614 & 3.909 & 1.543 & 0.436 & 0.440 \\
& Avg & \textcolor{blue}{0.270} & \textcolor{red}{0.320} & \textcolor{red}{0.263} & \textcolor{blue}{0.323} & 0.292 & 0.342 & 0.277 & 0.323 & 0.293 & 0.335 & 0.296 & 0.343 & 0.320 & 0.353 & 0.463 & 0.488 & 3.370 & 1.440 & 0.316 & 0.368 \\
\midrule

\multirow{5}{*}{\rotatebox{90}{Weather}}
& 96  & \textcolor{red}{0.153} & \textcolor{red}{0.208} & 0.174 & 0.228 & 0.167 & 0.215 & \textcolor{blue}{0.161} & \textcolor{blue}{0.210} & 0.163 & 0.215 & 0.165 & 0.215 & 0.184 & 0.230 & 0.188 & 0.253 & 0.374 & 0.401 & 0.171 & 0.224 \\
& 192 & \textcolor{red}{0.199} & \textcolor{blue}{0.251} & 0.217 & 0.262 & 0.210 & 0.262 & \textcolor{blue}{0.204} & \textcolor{red}{0.248} & 0.210 & \textcolor{blue}{0.254} & 0.210 & 0.257 & 0.245 & 0.283 & 0.250 & 0.304 & 0.552 & 0.478 & 0.215 & 0.263 \\
& 336 & 0.262 & 0.299 & 0.263 & \textcolor{blue}{0.296} & 0.267 & 0.306 & 0.261 & 0.302 & \textcolor{red}{0.256} & \textcolor{red}{0.292} & 0.259 & 0.297 & 0.305 & 0.321 & 0.312 & 0.346 & 0.724 & 0.541 & \textcolor{blue}{0.258} & 0.299 \\
& 720 & \textcolor{red}{0.306} & \textcolor{red}{0.327} & 0.326 & 0.340 & 0.344 & 0.356 & \textcolor{blue}{0.309} & \textcolor{blue}{0.332} & 0.321 & 0.339 & 0.332 & 0.346 & 0.381 & 0.371 & 0.387 & 0.393 & 0.739 & 0.558 & 0.320 & 0.346 \\
& Avg & \textcolor{red}{0.230} & \textcolor{red}{0.271} & 0.245 & 0.282 & 0.247 & 0.285 & \textcolor{blue}{0.234} & \textcolor{blue}{0.273} & 0.238 & 0.275 & 0.242 & 0.279 & 0.279 & 0.301 & 0.284 & 0.324 & 0.597 & 0.495 & 0.241 & 0.283 \\
\midrule

\multirow{5}{*}{\rotatebox{90}{Electricity}}
& 96  & \textcolor{blue}{0.137} & \textcolor{blue}{0.228} & 0.160 & 0.269 & \textcolor{red}{0.135} & \textcolor{red}{0.224} & 0.139 & 0.241 & 0.139 & 0.237 & 0.140 & 0.238 & 0.299 & 0.373 & 0.231 & 0.323 & 1.259 & 0.919 & 0.150 & 0.253 \\
& 192 & 0.163 & 0.252 & 0.174 & 0.279 & \textcolor{blue}{0.154} & \textcolor{red}{0.243} & \textcolor{red}{0.151} & \textcolor{blue}{0.248} & 0.156 & 0.252 & 0.160 & 0.255 & 0.305 & 0.379 & 0.261 & 0.356 & 1.160 & 0.873 & 0.164 & 0.264 \\
& 336 & 0.185 & 0.273 & 0.190 & 0.294 & 0.178 & \textcolor{red}{0.267} & \textcolor{red}{0.169} & \textcolor{blue}{0.270} & \textcolor{blue}{0.175} & \textcolor{blue}{0.270} & 0.180 & 0.276 & 0.319 & 0.391 & 0.360 & 0.445 & 1.157 & 0.872 & 0.181 & 0.282 \\
& 720 & 0.241 & \textcolor{blue}{0.319} & \textcolor{blue}{0.229} & 0.323 & 0.253 & 0.326 & 0.240 & 0.322 & 0.233 & \textcolor{red}{0.317} & 0.241 & 0.323 & 0.369 & 0.426 & 0.530 & 0.585 & 1.203 & 0.898 & \textcolor{red}{0.223} & 0.321 \\
& Avg & 0.199 & \textcolor{blue}{0.268} & 0.198 & 0.291 & 0.180 & \textcolor{red}{0.265} & \textcolor{red}{0.175} & 0.270 & \textcolor{blue}{0.176} & 0.269 & 0.180 & 0.273 & 0.323 & 0.392 & 0.346 & 0.427 & 1.195 & 0.891 & 0.180 & 0.280 \\
\midrule

\multirow{5}{*}{\rotatebox{90}{Traffic}}
& 96  & \textcolor{blue}{0.404} & \textcolor{red}{0.277} & 0.465 & 0.349 & 0.515 & 0.367 & 0.418 & 0.291 & 0.414 & 0.297 & \textcolor{red}{0.403} & \textcolor{blue}{0.289} & 0.719 & 0.416 & 0.639 & 0.400 & 1.557 & 0.821 & 0.419 & 0.298 \\
& 192 & 0.469 & 0.315 & 0.468 & 0.350 & 0.464 & 0.343 & \textcolor{red}{0.414} & \textcolor{red}{0.296} & 0.426 & \textcolor{blue}{0.301} & \textcolor{blue}{0.415} & \textcolor{red}{0.296} & 0.748 & 0.428 & 0.637 & 0.416 & 1.454 & 0.765 & 0.434 & 0.305 \\
& 336 & 0.453 & \textcolor{red}{0.301} & 0.483 & 0.356 & 0.451 & 0.318 & \textcolor{red}{0.421} & 0.311 & 0.434 & \textcolor{blue}{0.303} & \textcolor{blue}{0.426} & 0.304 & 0.853 & 0.471 & 0.655 & 0.427 & 1.521 & 0.812 & 0.449 & 0.313 \\
& 720 & 0.544 & 0.346 & 0.520 & 0.373 & 0.502 & 0.347 & \textcolor{red}{0.462} & \textcolor{red}{0.327} & 0.487 & 0.337 & \textcolor{blue}{0.474} & \textcolor{blue}{0.331} & 1.485 & 0.825 & 0.722 & 0.456 & 1.605 & 0.846 & 0.484 & 0.336 \\
& Avg & 0.468 & 0.310 & 0.484 & 0.357 & 0.483 & 0.344 & \textcolor{red}{0.429} & \textcolor{red}{0.306} & 0.440 & \textcolor{blue}{0.310} & \textcolor{blue}{0.430} & 0.305 & 0.951 & 0.535 & 0.663 & 0.425 & 1.534 & 0.811 & 0.447 & 0.313 \\

\midrule

\multicolumn{2}{c|}{\textbf{1st count}} 
& \multicolumn{2}{c|}{\textbf{35}} 
& \multicolumn{2}{c|}{6} 
& \multicolumn{2}{c|}{6} 
& \multicolumn{2}{c|}{18} 
& \multicolumn{2}{c|}{3} 
& \multicolumn{2}{c|}{2} 
& \multicolumn{2}{c|}{0} 
& \multicolumn{2}{c|}{0} 
& \multicolumn{2}{c|}{0} 
& \multicolumn{2}{c}{1} \\

\bottomrule
\end{tabular}}
\end{table}

\begin{table}[H]
\centering
\caption{\small Full few-shot learning results on 5\% training data. ''-'' denotes that 5\% of the time series is insufficient to form a valid training set. }
\label{tab:full5_results}
\renewcommand{\arraystretch}{0.95}
\resizebox{\textwidth}{!}{
\begin{tabular}{ll|cc|cc|cc|cc|cc|cc|cc|cc|cc|cc}
\toprule

\multicolumn{2}{c}{Models} &
\multicolumn{2}{c}{SSDA} &
\multicolumn{2}{c}{Time-VLM} &
\multicolumn{2}{c}{VisionTS} &
\multicolumn{2}{c}{Time-LLM} &
\multicolumn{2}{c}{GPT4TS} &
\multicolumn{2}{c}{PatchTST} &
\multicolumn{2}{c}{TimesNet} &
\multicolumn{2}{c}{FEDformer} &
\multicolumn{2}{c}{Informer} &
\multicolumn{2}{c}{DLinear} \\

\cmidrule(lr){3-4}\cmidrule(lr){5-6}\cmidrule(lr){7-8}\cmidrule(lr){9-10}
\cmidrule(lr){11-12}\cmidrule(lr){13-14}\cmidrule(lr){15-16}
\cmidrule(lr){17-18}\cmidrule(lr){19-20}\cmidrule(lr){21-22}

\multicolumn{2}{c}{Metrics} &
MSE & MAE & MSE & MAE & MSE & MAE & MSE & MAE &
MSE & MAE & MSE & MAE & MSE & MAE & MSE & MAE &
MSE & MAE & MSE & MAE \\
\midrule

\multirow{5}{*}{\rotatebox{90}{ETTh1}}
& 96  & \textcolor{red}{0.359} & \textcolor{red}{0.376} & 0.417 & 0.435 & \textcolor{blue}{0.369} & \textcolor{blue}{0.377} & 0.483 & 0.464 & 0.543 & 0.506 & 0.557 & 0.519 & 0.892 & 0.625 & 0.593 & 0.529 & 1.225 & 0.812 & 0.547 & 0.503 \\
& 192 & \textcolor{red}{0.396} & \textcolor{red}{0.406} & 0.450 & 0.458 & \textcolor{blue}{0.416} & \textcolor{blue}{0.408} & 0.629 & 0.540 & 0.748 & 0.580 & 0.711 & 0.570 & 0.940 & 0.665 & 0.652 & 0.563 & 1.249 & 0.828 & 0.720 & 0.604 \\
& 336 & \textcolor{red}{0.415} & \textcolor{red}{0.421} & 0.460 & 0.465 & \textcolor{blue}{0.442} & \textcolor{blue}{0.424} & 0.768 & 0.626 & 0.754 & 0.595 & 0.816 & 0.619 & 0.945 & 0.653 & 0.731 & 0.594 & 1.202 & 0.811 & 0.984 & 0.727 \\
& 720 & -- & -- & -- & -- & -- & -- & -- & -- & -- & -- & -- & -- & -- & -- & -- & -- & -- & -- & -- & -- \\
& Avg & \textcolor{red}{0.390} & \textcolor{red}{0.401} & 0.442 & 0.453 & \textcolor{blue}{0.409} & \textcolor{blue}{0.403} & 0.627 & 0.543 & 0.681 & 0.560 & 0.694 & 0.569 & 0.925 & 0.647 & 0.658 & 0.562 & 1.225 & 0.817 & 0.750 & 0.611 \\
\midrule

\multirow{5}{*}{\rotatebox{90}{ETTh2}}
& 96  & \textcolor{red}{0.265} & \textcolor{red}{0.325} & 0.302 & 0.365 & \textcolor{blue}{0.300} & \textcolor{blue}{0.337} & 0.336 & 0.397 & 0.376 & 0.421 & 0.401 & 0.421 & 0.409 & 0.420 & 0.390 & 0.424 & 3.837 & 1.508 & 0.442 & 0.456 \\
& 192 & \textcolor{red}{0.319} & \textcolor{red}{0.363} & \textcolor{blue}{0.361} & 0.406 & 0.365 & \textcolor{blue}{0.388} & 0.406 & 0.425 & 0.418 & 0.441 & 0.452 & 0.455 & 0.483 & 0.464 & 0.457 & 0.465 & 3.975 & 1.933 & 0.617 & 0.542 \\
& 336 & \textcolor{red}{0.349} & \textcolor{red}{0.385} & 0.398 & 0.434 & \textcolor{blue}{0.379} & \textcolor{blue}{0.402} & 0.405 & 0.432 & 0.408 & 0.439 & 0.464 & 0.469 & 0.499 & 0.479 & 0.477 & 0.483 & 3.956 & 1.520 & 1.424 & 0.849 \\
& 720 & -- & -- & -- & -- & -- & -- & -- & -- & -- & -- & -- & -- & -- & -- & -- & -- & -- & -- & -- & -- \\
& Avg & \textcolor{red}{0.311} & \textcolor{red}{0.358} & 0.354 & 0.402 & \textcolor{blue}{0.348} & \textcolor{blue}{0.376} & 0.382 & 0.418 & 0.401 & 0.433 & 0.439 & 0.448 & 0.463 & 0.454 & 0.441 & 0.457 & 3.923 & 1.654 & 0.694 & 0.577 \\
\midrule

\multirow{5}{*}{\rotatebox{90}{ETTm1}}
& 96  & 0.330 & \textcolor{red}{0.342} & \textcolor{red}{0.314} & \textcolor{blue}{0.357} & 0.444 & 0.433 & \textcolor{blue}{0.316} & 0.377 & 0.386 & 0.405 & 0.399 & 0.414 & 0.606 & 0.518 & 0.628 & 0.544 & 1.130 & 0.775 & 0.332 & 0.374 \\
& 192 & \textcolor{blue}{0.347} & \textcolor{red}{0.358} & \textcolor{red}{0.343} & \textcolor{blue}{0.373} & 0.497 & 0.458 & 0.450 & 0.464 & 0.440 & 0.438 & 0.441 & 0.436 & 0.681 & 0.539 & 0.666 & 0.566 & 1.150 & 0.788 & 0.358 & 0.390 \\
& 336 & \textcolor{red}{0.369} & \textcolor{red}{0.373} & \textcolor{blue}{0.373} & \textcolor{blue}{0.391} & 0.653 & 0.530 & 0.450 & 0.424 & 0.485 & 0.459 & 0.499 & 0.467 & 0.786 & 0.597 & 0.807 & 0.628 & 1.198 & 0.809 & 0.402 & 0.416 \\
& 720 & \textcolor{red}{0.404} & \textcolor{red}{0.403} & \textcolor{blue}{0.425} & \textcolor{blue}{0.420} & 0.702 & 0.572 & 0.483 & 0.471 & 0.577 & 0.499 & 0.767 & 0.587 & 0.796 & 0.593 & 0.822 & 0.633 & 1.175 & 0.794 & 0.511 & 0.489 \\
& Avg & \textcolor{red}{0.363} & \textcolor{red}{0.369} & \textcolor{blue}{0.364} & \textcolor{blue}{0.385} & 0.574 & 0.498 & 0.425 & 0.434 & 0.472 & 0.450 & 0.526 & 0.476 & 0.717 & 0.561 & 0.730 & 0.592 & 1.163 & 0.791 & 0.400 & 0.417 \\
\midrule

\multirow{5}{*}{\rotatebox{90}{ETTm2}}
& 96  & 0.207 & 0.272 & \textcolor{red}{0.169} & \textcolor{red}{0.260} & 0.196 & 0.277 & \textcolor{blue}{0.174} & \textcolor{blue}{0.261} & 0.199 & 0.280 & 0.206 & 0.288 & 0.220 & 0.299 & 0.229 & 0.320 & 3.599 & 1.478 & 0.236 & 0.326 \\
& 192 & 0.282 & 0.327 & \textcolor{blue}{0.224} & \textcolor{blue}{0.298} & 0.255 & 0.316 & \textcolor{red}{0.215} & \textcolor{red}{0.287} & 0.256 & 0.316 & 0.264 & 0.324 & 0.311 & 0.361 & 0.394 & 0.361 & 3.578 & 1.475 & 0.306 & 0.373 \\
& 336 & 0.285 & \textcolor{red}{0.329} & \textcolor{blue}{0.282} & 0.338 & 0.321 & 0.364 & \textcolor{red}{0.273} & \textcolor{blue}{0.330} & 0.318 & 0.353 & 0.334 & 0.367 & 0.338 & 0.366 & 0.378 & 0.427 & 3.561 & 1.473 & 0.380 & 0.423 \\
& 720 & \textcolor{red}{0.350} & \textcolor{red}{0.377} & \textcolor{blue}{0.375} & \textcolor{blue}{0.397} & 0.459 & 0.444 & 0.433 & 0.412 & 0.460 & 0.436 & 0.454 & 0.432 & 0.509 & 0.465 & 0.523 & 0.510 & 3.896 & 1.533 & 0.674 & 0.583 \\
& Avg & 0.281 & \textcolor{blue}{0.326} & \textcolor{red}{0.262} & \textcolor{red}{0.323} & 0.308 & 0.350 & \textcolor{blue}{0.274} & \textcolor{red}{0.323} & 0.308 & 0.346 & 0.314 & 0.352 & 0.344 & 0.372 & 0.381 & 0.404 & 3.658 & 1.489 & 0.399 & 0.426 \\
\midrule

\multirow{5}{*}{\rotatebox{90}{Weather}}
& 96  & 0.175 & \textcolor{red}{0.224} & 0.176 & 0.231 & 0.174 & \textcolor{blue}{0.228} & \textcolor{blue}{0.172} & 0.263 & 0.175 & 0.230 & \textcolor{red}{0.171} & \textcolor{red}{0.224} & 0.207 & 0.253 & 0.229 & 0.309 & 0.497 & 0.497 & 0.184 & 0.242 \\
& 192 & \textcolor{red}{0.211} & \textcolor{red}{0.254} & \textcolor{blue}{0.216} & \textcolor{blue}{0.263} & 0.225 & 0.275 & 0.224 & 0.271 & 0.227 & 0.276 & 0.230 & 0.277 & 0.272 & 0.307 & 0.265 & 0.317 & 0.620 & 0.545 & 0.228 & 0.283 \\
& 336 & \textcolor{red}{0.262} & \textcolor{red}{0.289} & \textcolor{blue}{0.264} & \textcolor{blue}{0.298} & 0.284 & 0.323 & 0.282 & 0.321 & 0.286 & 0.322 & 0.294 & 0.326 & 0.313 & 0.328 & 0.353 & 0.392 & 0.649 & 0.547 & 0.279 & 0.322 \\
& 720 & \textcolor{red}{0.326} & \textcolor{red}{0.332} & \textcolor{blue}{0.327} & \textcolor{blue}{0.342} & 0.338 & 0.357 & 0.366 & 0.381 & 0.366 & 0.379 & 0.384 & 0.387 & 0.400 & 0.385 & 0.391 & 0.394 & 0.570 & 0.522 & 0.364 & 0.388 \\
& Avg & \textcolor{red}{0.244} & \textcolor{red}{0.275} & \textcolor{blue}{0.246} & \textcolor{blue}{0.284} & 0.255 & 0.296 & 0.260 & 0.309 & 0.263 & 0.301 & 0.269 & 0.303 & 0.298 & 0.318 & 0.309 & 0.353 & 0.584 & 0.527 & 0.263 & 0.308 \\
\midrule

\multirow{5}{*}{\rotatebox{90}{Electricity}}
& 96  & 0.149 & \textcolor{blue}{0.240} & 0.185 & 0.296 & \textcolor{red}{0.141} & \textcolor{red}{0.230} & 0.147 & 0.242 & \textcolor{blue}{0.143} & 0.241 & 0.145 & 0.244 & 0.315 & 0.389 & 0.235 & 0.322 & 1.265 & 0.919 & 0.150 & 0.251 \\
& 192 & 0.166 & 0.256 & 0.194 & 0.302 & 0.160 & \textcolor{blue}{0.249} & \textcolor{red}{0.158} & \textcolor{red}{0.241} & \textcolor{blue}{0.159} & 0.255 & 0.163 & 0.260 & 0.318 & 0.396 & 0.247 & 0.341 & 1.298 & 0.939 & 0.163 & 0.263 \\
& 336 & 0.199 & 0.283 & 0.210 & 0.315 & 0.189 & \textcolor{blue}{0.277} & \textcolor{blue}{0.178} & \textcolor{blue}{0.277} & 0.179 & \textcolor{red}{0.274} & 0.183 & 0.281 & 0.340 & 0.415 & 0.267 & 0.356 & 1.302 & 0.942 & \textcolor{red}{0.175} & 0.278 \\
& 720 & 0.248 & 0.324 & 0.251 & 0.346 & 0.243 & 0.320 & \textcolor{blue}{0.224} & \textcolor{blue}{0.312} & 0.233 & 0.323 & 0.233 & 0.323 & 0.635 & 0.613 & 0.318 & 0.394 & 1.259 & 0.919 & \textcolor{red}{0.219} & \textcolor{red}{0.311} \\
& Avg & 0.191 & 0.276 & 0.218 & 0.315 & 0.183 & \textcolor{blue}{0.269} & 0.179 & \textcolor{red}{0.268} & \textcolor{blue}{0.178} & 0.273 & 0.181 & 0.277 & 0.402 & 0.453 & 0.266 & 0.353 & 1.281 & 0.929 & \textcolor{red}{0.176} & 0.275 \\
\midrule

\multirow{5}{*}{\rotatebox{90}{Traffic}}
& 96  & 0.469 & 0.315 & 0.550 & 0.408 & 1.278 & 0.796 & \textcolor{blue}{0.414} & \textcolor{blue}{0.291} & 0.419 & 0.298 & \textcolor{red}{0.404} & \textcolor{red}{0.286} & 0.854 & 0.492 & 0.670 & 0.421 & 1.557 & 0.821 & 0.427 & 0.304 \\
& 192 & 0.482 & 0.320 & 0.552 & 0.408 & 0.423 & 0.298 & \textcolor{blue}{0.419} & \textcolor{red}{0.291} & 0.434 & 0.305 & \textcolor{red}{0.412} & \textcolor{blue}{0.294} & 0.894 & 0.517 & 0.653 & 0.405 & 1.596 & 0.834 & 0.447 & 0.315 \\
& 336 & 0.528 & 0.354 & 0.572 & 0.414 & 0.444 & \textcolor{blue}{0.312} & \textcolor{red}{0.437} & 0.314 & 0.449 & 0.313 & \textcolor{blue}{0.439} & \textcolor{red}{0.310} & 0.853 & 0.471 & 0.707 & 0.445 & 1.621 & 0.841 & 0.478 & 0.333 \\
& 720 & -- & -- & -- & -- & -- & -- & -- & -- & -- & -- & -- & -- & -- & -- & -- & -- & -- & -- & -- & -- \\
& Avg & 0.493 & 0.330 & 0.558 & 0.410 & 0.715 & 0.469 & \textcolor{blue}{0.423} & \textcolor{blue}{0.298} & 0.434 & 0.305 & \textcolor{red}{0.418} & \textcolor{red}{0.296} & 0.867 & 0.493 & 0.676 & 0.423 & 1.591 & 0.832 & 0.450 & 0.317 \\

\midrule

\multicolumn{2}{c|}{\textbf{1st count}} 
& \multicolumn{2}{c|}{\textbf{29}} 
& \multicolumn{2}{c|}{12} 
& \multicolumn{2}{c|}{2} 
& \multicolumn{2}{c|}{9} 
& \multicolumn{2}{c|}{1} 
& \multicolumn{2}{c|}{8} 
& \multicolumn{2}{c|}{0} 
& \multicolumn{2}{c|}{0} 
& \multicolumn{2}{c|}{0} 
& \multicolumn{2}{c}{4} \\

\bottomrule
\end{tabular}}
\end{table}

\subsection{Ablation Study}
\label{apd:ablation}

\begin{table}[H]
\centering
\caption{Full resutls of the ablation study.}
\label{tab:full_ablation}

\resizebox{0.92\textwidth}{!}{
\begin{tabular}{cc|cc|cc|cc|cc|cc}
\toprule
\multicolumn{2}{c}{Models} 
& \multicolumn{2}{c}{SSDA} 
& \multicolumn{2}{c}{w/o Structural Branch} 
& \multicolumn{2}{c}{w/o Spectral Branch} 
& \multicolumn{2}{c}{VisionTS+LoRA} 
& \multicolumn{2}{c}{VisionTS} \\
\cmidrule(lr){3-4} \cmidrule(lr){5-6} \cmidrule(lr){7-8} \cmidrule(lr){9-10} \cmidrule(lr){11-12}
\multicolumn{2}{c}{Metric} 
& MSE & MAE & MSE & MAE & MSE & MAE & MSE & MAE & MSE & MAE \\
\midrule

\multirow{5}{*}{ETTh1}
& 96  & \textcolor{red}{0.343} & \textcolor{red}{0.372} & 0.373 & \textcolor{blue}{0.386} & \textcolor{blue}{0.344} & \textcolor{red}{0.372} & 0.473 & 0.428 & 0.355 & \textcolor{blue}{0.386} \\
& 192 & \textcolor{red}{0.387} & \textcolor{red}{0.402} & 0.398 & 0.408 & \textcolor{blue}{0.395} & \textcolor{blue}{0.406} & 0.473 & 0.436 & \textcolor{blue}{0.395} & 0.407 \\
& 336 & \textcolor{blue}{0.409} & \textcolor{blue}{0.420} & 0.418 & 0.424 & \textcolor{red}{0.405} & \textcolor{red}{0.416} & 0.464 & 0.439 & 0.419 & 0.421 \\
& 720 & \textcolor{red}{0.408} & \textcolor{red}{0.430} & 0.423 & 0.448 & \textcolor{blue}{0.420} & \textcolor{blue}{0.439} & 0.436 & 0.448 & 0.458 & 0.460 \\
& Avg & \textcolor{red}{0.387} & \textcolor{red}{0.406} & 0.403 & 0.417 & \textcolor{blue}{0.391} & \textcolor{blue}{0.408} & 0.462 & 0.438 & 0.407 & 0.419 \\

\midrule
\multirow{5}{*}{ETTh2}
& 96  & \textcolor{red}{0.267} & \textcolor{red}{0.328} & 0.279 & 0.336 & \textcolor{blue}{0.274} & \textcolor{blue}{0.329} & 0.402 & 0.399 & 0.288 & 0.334 \\
& 192 & \textcolor{red}{0.324} & \textcolor{red}{0.368} & \textcolor{blue}{0.326} & \textcolor{blue}{0.369} & 0.331 & 0.373 & 0.396 & 0.406 & 0.349 & 0.380 \\
& 336 & \textcolor{red}{0.347} & \textcolor{red}{0.392} & 0.351 & \textcolor{blue}{0.393} & \textcolor{blue}{0.350} & \textcolor{blue}{0.393} & 0.397 & 0.415 & 0.364 & 0.398 \\
& 720 & \textcolor{red}{0.370} & \textcolor{blue}{0.415} & \textcolor{blue}{0.372} & \textcolor{red}{0.414} & \textcolor{blue}{0.372} & 0.418 & 0.390 & 0.421 & 0.403 & 0.431 \\
& Avg & \textcolor{red}{0.327} & \textcolor{red}{0.376} & \textcolor{blue}{0.332} & \textcolor{blue}{0.378} & \textcolor{blue}{0.332} & \textcolor{blue}{0.378} & 0.396 & 0.410 & 0.351 & 0.386 \\

\midrule
\multirow{5}{*}{ETTm1}
& 96  & \textcolor{red}{0.283} & \textcolor{red}{0.328} & 0.327 & 0.347 & 0.302 & 0.338 & 0.319 & 0.342 & \textcolor{blue}{0.284} & \textcolor{blue}{0.332} \\
& 192 & \textcolor{red}{0.316} & \textcolor{red}{0.353} & 0.343 & 0.358 & \textcolor{blue}{0.325} & \textcolor{blue}{0.354} & 0.337 & 0.357 & 0.327 & 0.362 \\
& 336 & \textcolor{red}{0.340} & \textcolor{red}{0.370} & 0.368 & 0.382 & \textcolor{blue}{0.349} & \textcolor{blue}{0.373} & 0.355 & \textcolor{blue}{0.373} & 0.354 & 0.382 \\
& 720 & \textcolor{red}{0.388} & \textcolor{red}{0.400} & 0.411 & 0.409 & \textcolor{blue}{0.390} & \textcolor{blue}{0.401} & 0.401 & 0.403 & 0.411 & 0.415 \\
& Avg & \textcolor{red}{0.332} & \textcolor{red}{0.363} & 0.362 & 0.374 & \textcolor{blue}{0.342} & \textcolor{blue}{0.367} & 0.353 & 0.369 & 0.344 & 0.373 \\

\midrule
\multirow{5}{*}{ETTm2}
& 96  & \textcolor{red}{0.168} & \textcolor{red}{0.256} & 0.218 & 0.281 & 0.179 & 0.264 & 0.206 & 0.276 & \textcolor{blue}{0.174} & \textcolor{blue}{0.262} \\
& 192 & \textcolor{red}{0.217} & \textcolor{red}{0.290} & 0.253 & 0.307 & 0.229 & \textcolor{blue}{0.296} & 0.244 & 0.305 & \textcolor{blue}{0.228} & 0.297 \\
& 336 & \textcolor{red}{0.268} & \textcolor{red}{0.323} & 0.292 & 0.334 & \textcolor{blue}{0.277} & \textcolor{blue}{0.330} & 0.284 & 0.331 & 0.281 & 0.337 \\
& 720 & \textcolor{red}{0.341} & \textcolor{red}{0.375} & 0.390 & 0.400 & 0.355 & 0.380 & \textcolor{blue}{0.354} & \textcolor{blue}{0.378} & 0.384 & 0.410 \\
& Avg & \textcolor{red}{0.249} & \textcolor{red}{0.311} & 0.288 & 0.331 & \textcolor{blue}{0.260} & \textcolor{blue}{0.318} & 0.272 & 0.323 & 0.267 & 0.327 \\

\bottomrule
\end{tabular}}
\end{table}

\subsection{Efficiency Analysis}
\label{apd:efficiency}

\begin{figure}[b]
    \centering
    \includegraphics[width=0.7\linewidth]{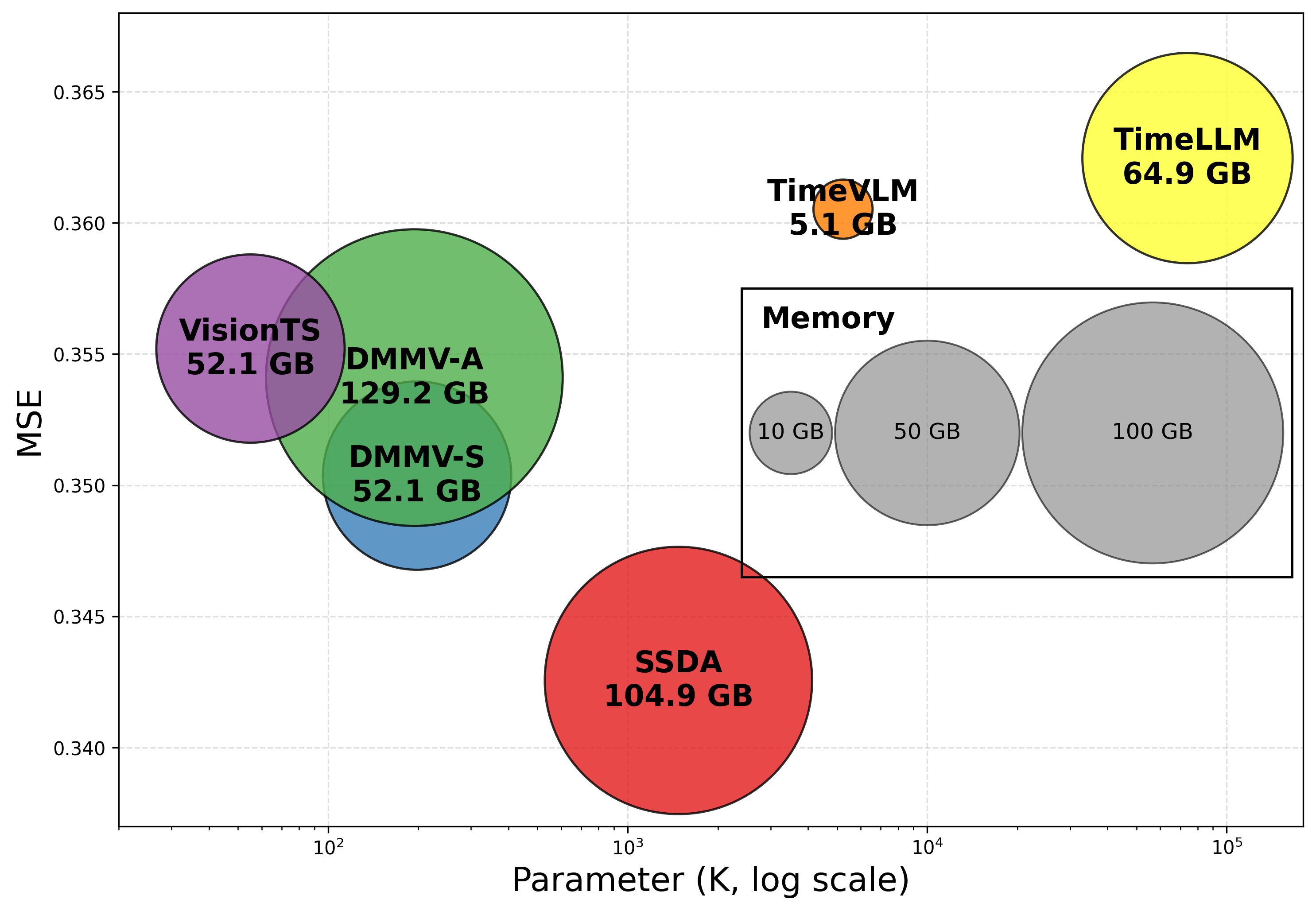}
    \caption{The bubble chart of performance vs.\ efficiency.}
    \label{fig:bubble}
\end{figure}

We conduct an efficiency analysis to evaluate the trade-off among forecasting performance, model complexity, and memory consumption. Specifically, we compare different models from three perspectives: (1) prediction accuracy measured by MSE, (2) model size quantified by the number of parameters, and (3) runtime memory usage. The results are visualized in a bubble chart, where the x-axis denotes the parameter scale (in log scale), the y-axis represents MSE, and the bubble size encodes memory consumption.

From the results, several important observations can be drawn. First, \textbf{SSDA achieves the best predictive performance} with the lowest MSE (0.343), while maintaining a moderate parameter scale (1.5M). Compared to lightweight models such as VisionTS and DMMV-S, SSDA significantly improves accuracy with only a moderate increase in memory usage, demonstrating a favorable efficiency-performance balance.

Second, although models such as TimeLLM and Time-VLM employ substantially larger parameter sizes (up to 76M), they do not yield corresponding performance gains. In fact, their MSE values are consistently higher than those of smaller or medium-scale models. This suggests that \textbf{simply scaling up model parameters does not necessarily translate into better forecasting performance} in the time series domain.

Overall, SSDA lies on a more favorable region of the performance-efficiency trade-off curve, achieving a better balance among accuracy, parameter efficiency, and memory usage. These results demonstrate that our design not only improves forecasting performance but also maintains strong efficiency characteristics, making it more suitable for practical deployment scenarios.


\subsection{Visualization}
\label{apd:visualizaiton}

In this section, we present a comprehensive visualization analysis of SSDA from both branch-level and model-level perspectives. Specifically, we visualize the input and reconstructed images of both the spectral branch and the structural branch, enabling a direct comparison of their respective reconstruction behaviors and prediction characteristics.
Furthermore, we compare the forecasting results of SSDA with those of DMMV by visualizing their predicted sequences alongside the ground truth. To provide a quantitative assessment, we also report the corresponding MSE and MAE values on selected samples.

The results consistently demonstrate that SSDA achieves superior prediction accuracy compared to DMMV. From the visualizations, SSDA produces predictions that are more closely aligned with the ground truth, exhibiting better trend fidelity and reduced noise. This improvement can be attributed to the complementary design of the two branches: the spectral branch enhances frequency-domain consistency, while the structural branch strengthens temporal dependency modeling. Together, these components effectively mitigate modality discrepancies, leading to more accurate and robust forecasting performance.

Beyond this overall improvement, an important and consistent phenomenon can be observed in the reconstructed images of the structural branch: clear vertical patterns persist across different samples.
These vertical structures indicate that the model effectively captures and reinforces temporal consistency along the time axis, where each column corresponds to a temporal slice of the multivariate sequence, while horizontal patterns indicate underlying periodicity in the data. The emergence and stability of such patterns indicate that the structural branch suppresses spurious spatial correlations introduced during the image transformation process, while preserving meaningful temporal dependencies.

\newpage

\begin{figure}[H]
    \centering

    \begin{minipage}[t]{0.20\textwidth}
        \centering
        \includegraphics[width=1\textwidth]{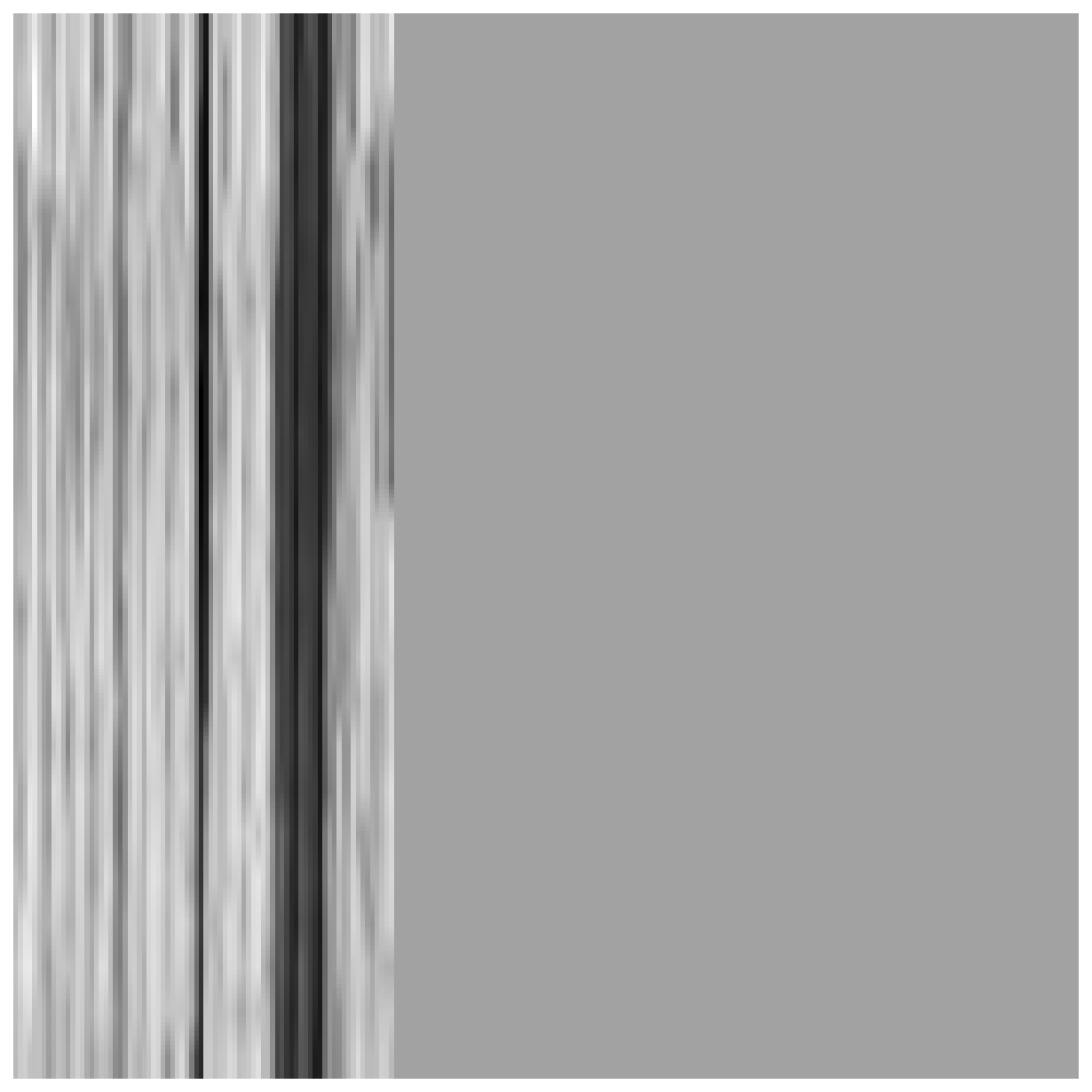}
        \small (a) Structural branch input image
        \label{fig:ettm2_var2_ti}
    \end{minipage}
    \hspace{2em}
    \begin{minipage}[t]{0.20\textwidth}
        \centering
        \includegraphics[width=1\textwidth]{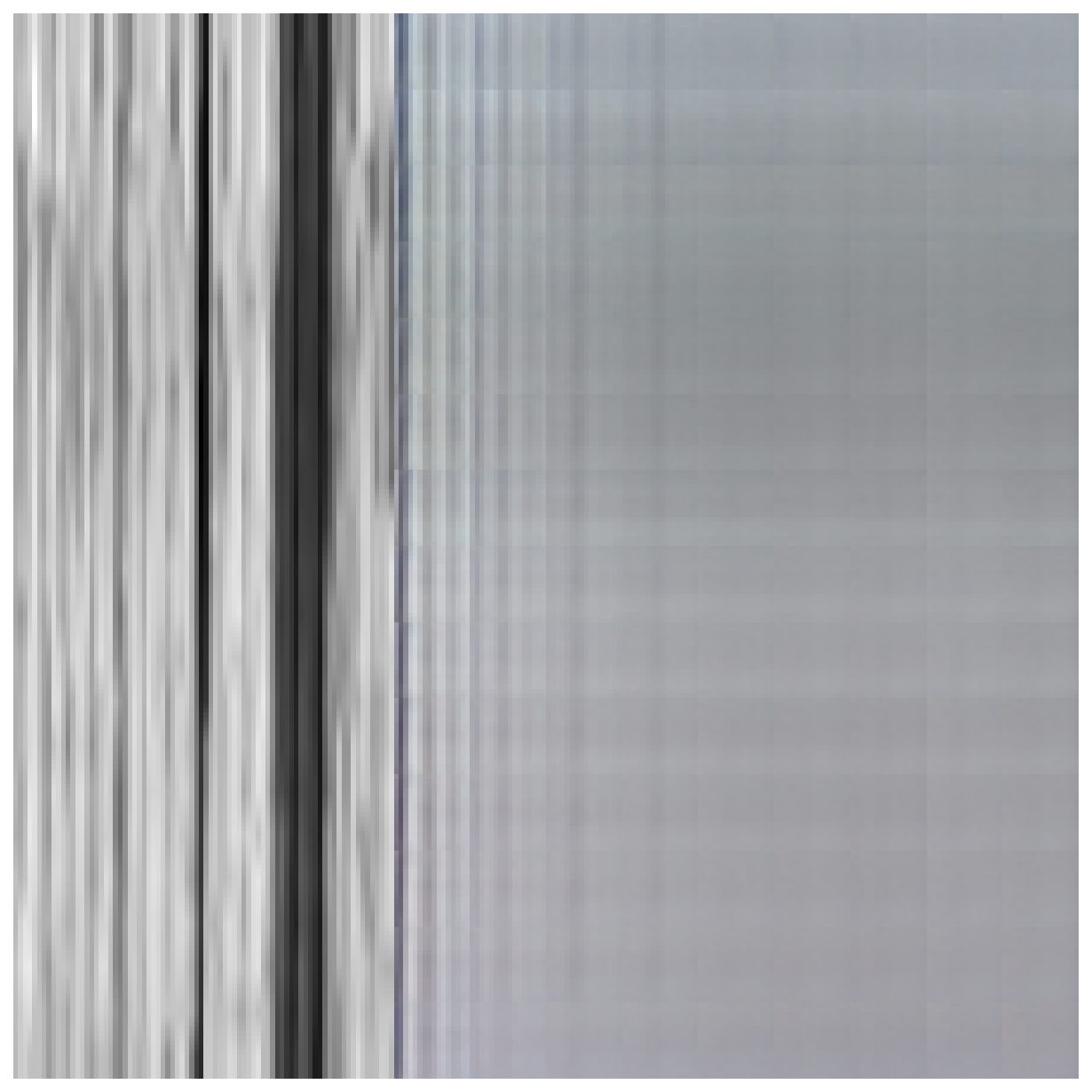}
        \small (b) Structural branch reconstructed image
        \label{fig:ettm2_var2_tr}
    \end{minipage}
    \hspace{2em}
    \begin{minipage}[t]{0.20\textwidth}
        \centering
        \includegraphics[width=1\textwidth]{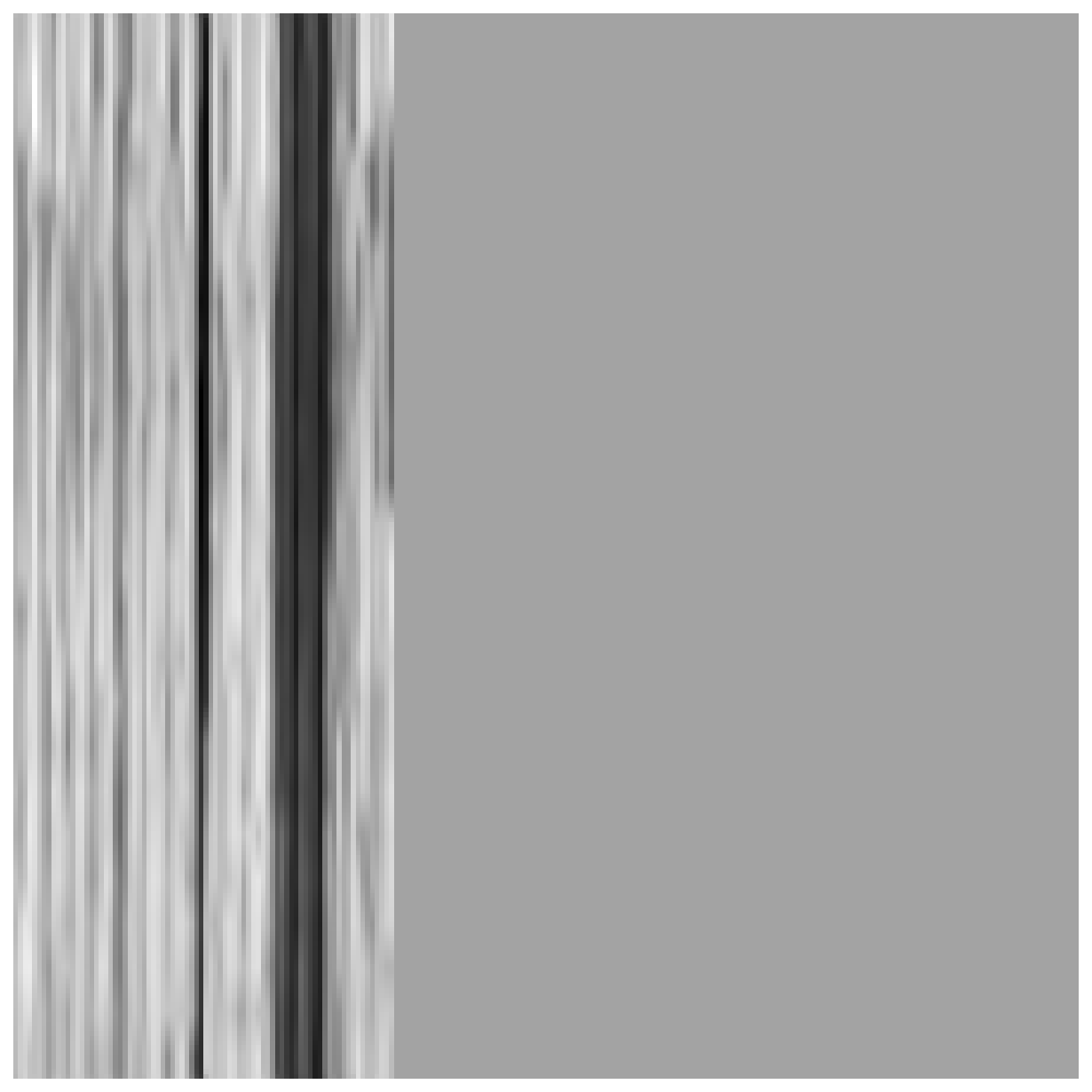}
        \small (c) Spectral branch input image
        \label{fig:ettm2_var2_vi}
    \end{minipage}
    \hspace{2em}
    \begin{minipage}[t]{0.20\textwidth}
        \centering
        \includegraphics[width=1\textwidth]{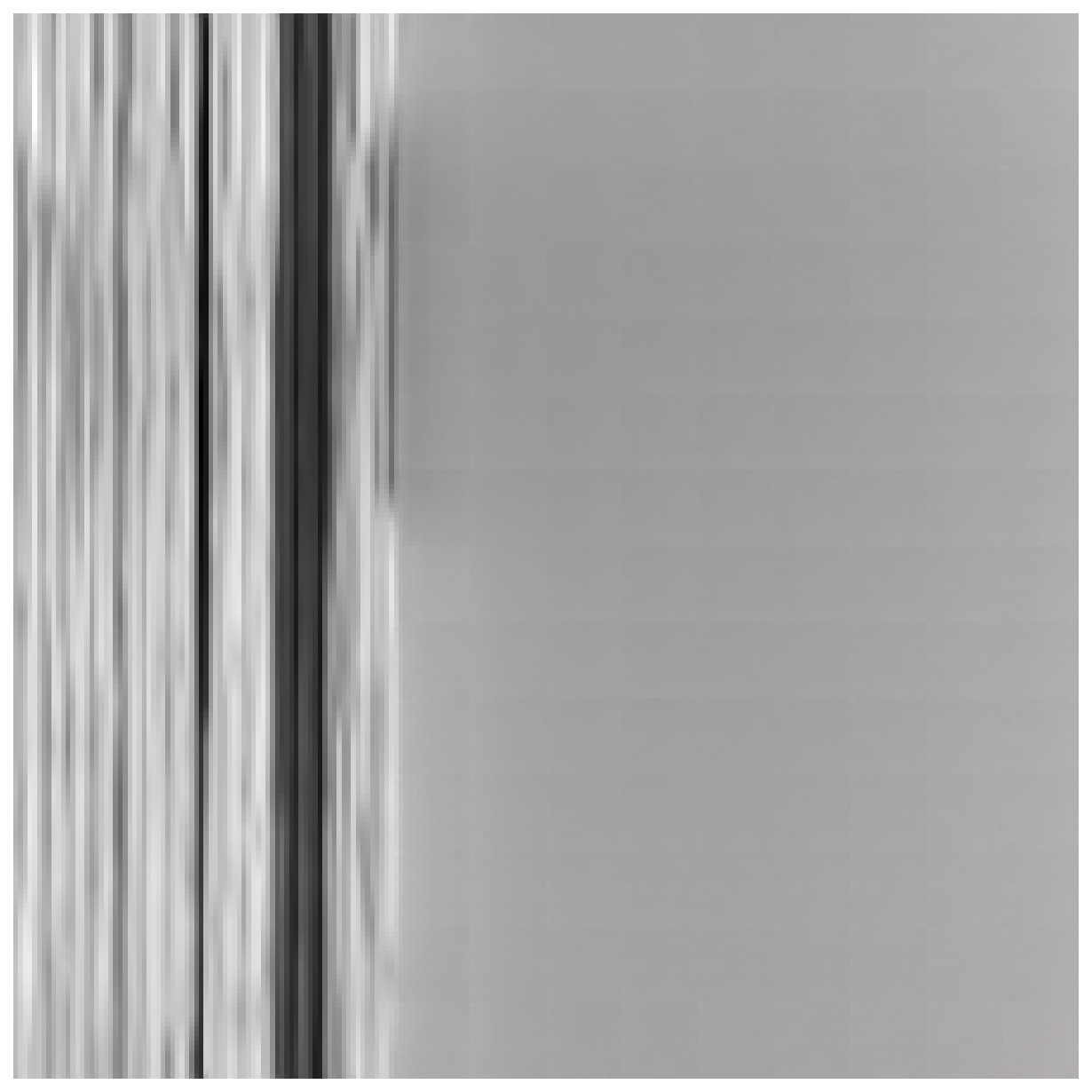}
        \small (d) Spectral branch reconstructed image
        \label{fig:ettm2_var2_vr}
    \end{minipage}

    \vspace{1em}

    \begin{minipage}[t]{1\textwidth}
        \centering
        \includegraphics[width=1\textwidth]{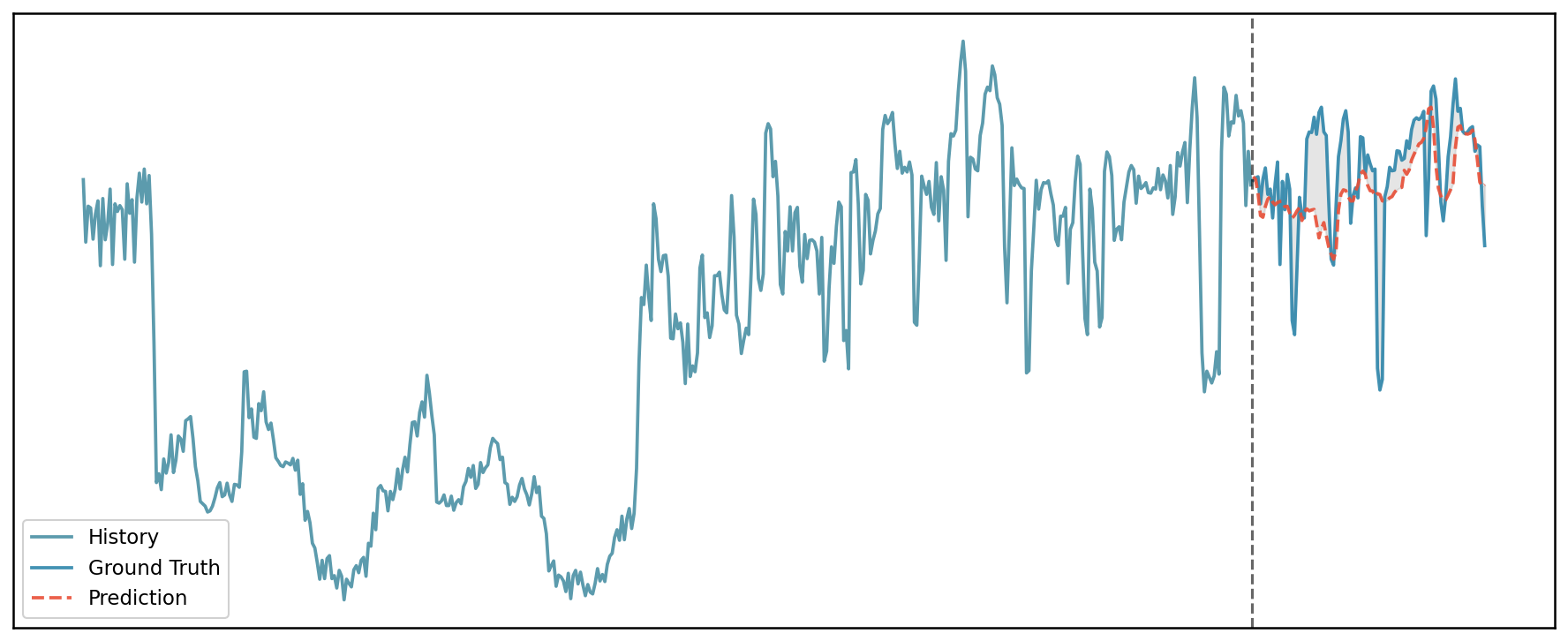}
        \small (e) SSDA (MSE=0.155,MAE=0.308).
        \label{fig:ettm2_var2_ssda}
    \end{minipage}

    \vspace{1em}

    \begin{minipage}[t]{1\textwidth}
        \centering
        \includegraphics[width=1\textwidth]{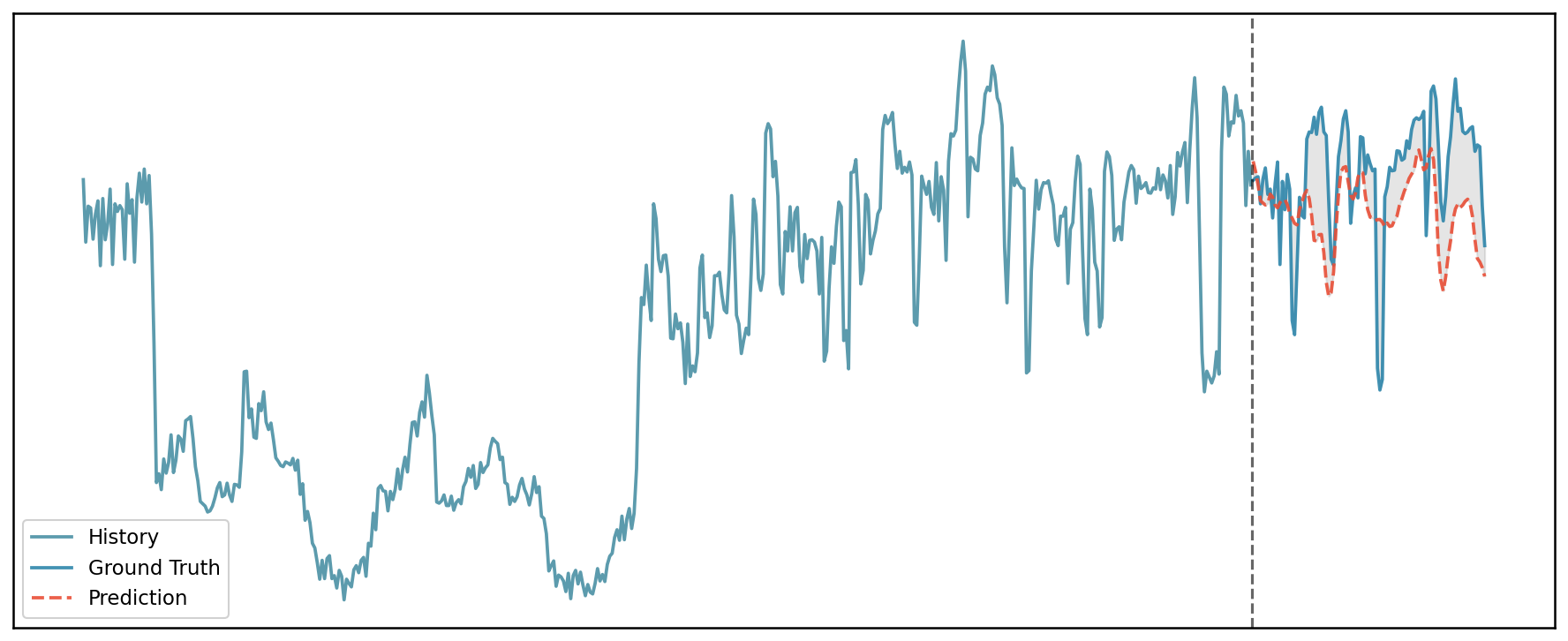}
        \small (f) DMMV (MSE=0.201,MAE=0.350).
        \label{fig:ettm2_var2_dmmv}
    \end{minipage}

    \caption{Forecasting visualization on a sample from ETTm2. (a) Input image of structural branch. (b) Output image of structural branch. (c) Input image of spectral branch(enhanced by Spectral Magnitude Aligner). (d) Output image of spectral branch. (e-f) Comparison of predictions and ground truths by SSDA and DMMV.}
    \label{fig:ettm2_var2_combined}
\end{figure}

\begin{figure}[H]
    \centering

    \begin{minipage}[t]{0.20\textwidth}
        \centering
        \includegraphics[width=1\textwidth]{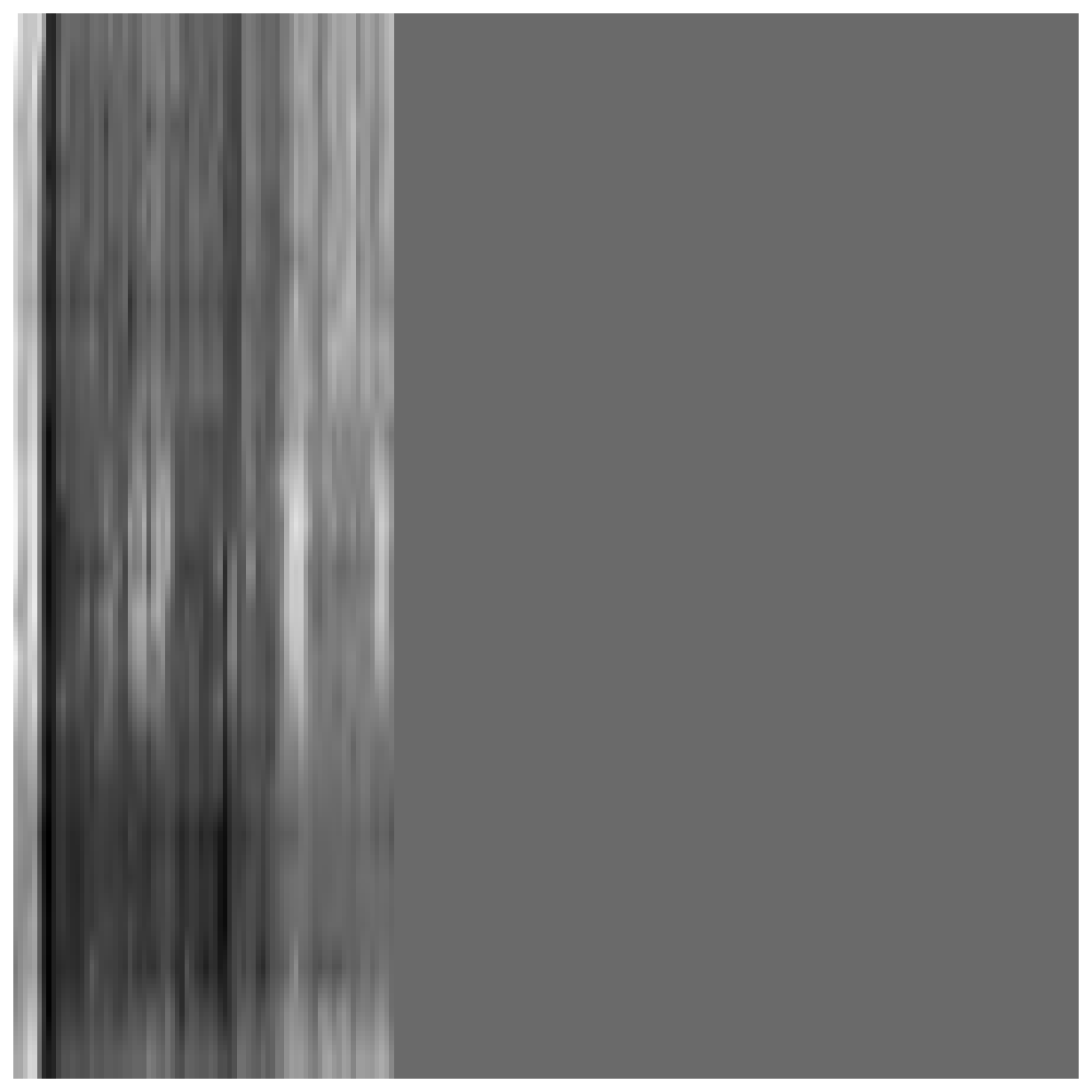}
        \small (a) Structural branch input image
        \label{fig:etth1_var1_ti}
    \end{minipage}
    \hspace{2em}
    \begin{minipage}[t]{0.20\textwidth}
        \centering
        \includegraphics[width=1\textwidth]{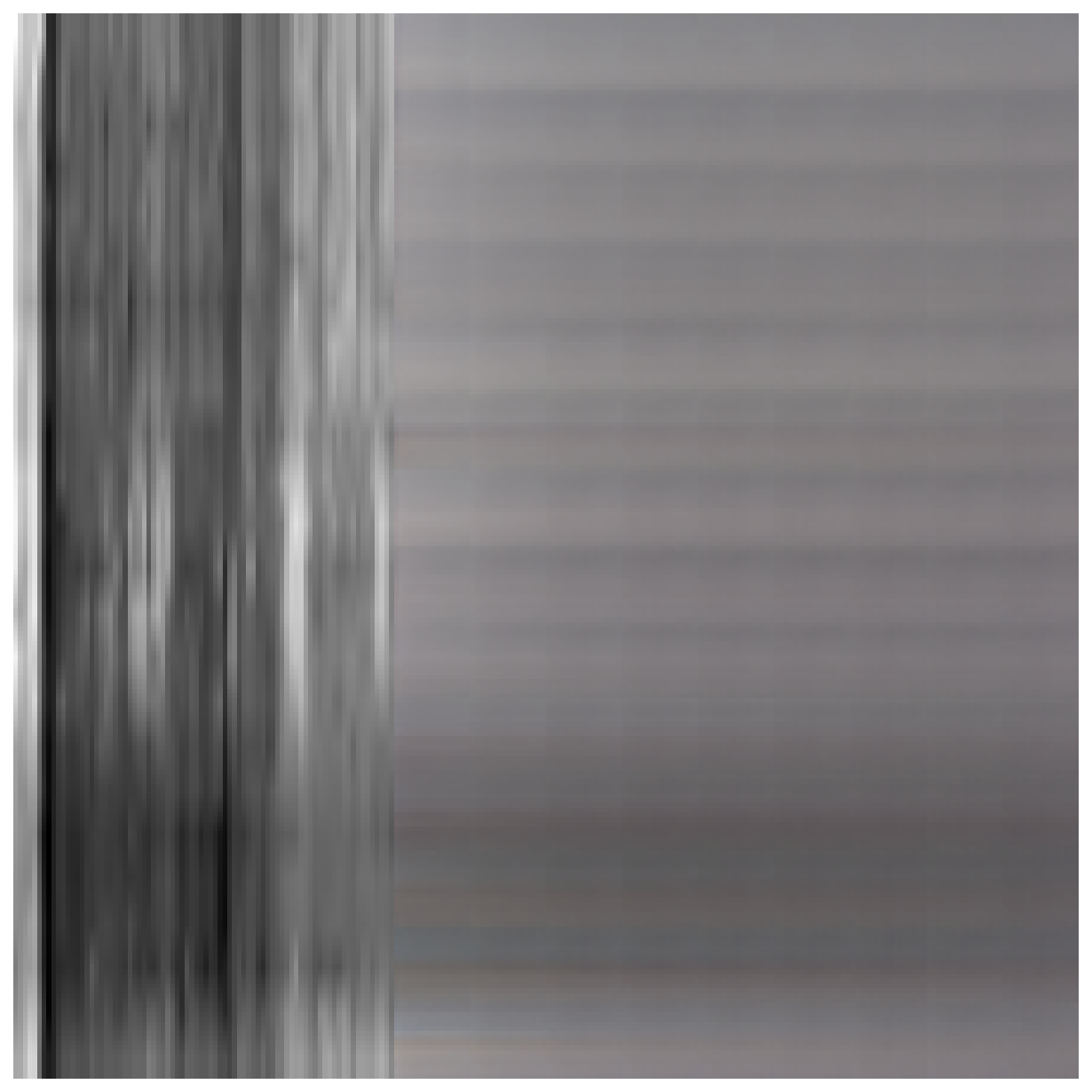}
        \small (b) Structural branch reconstructed image
        \label{fig:etth1_var1_tr}
    \end{minipage}
    \hspace{2em}
    \begin{minipage}[t]{0.20\textwidth}
        \centering
        \includegraphics[width=1\textwidth]{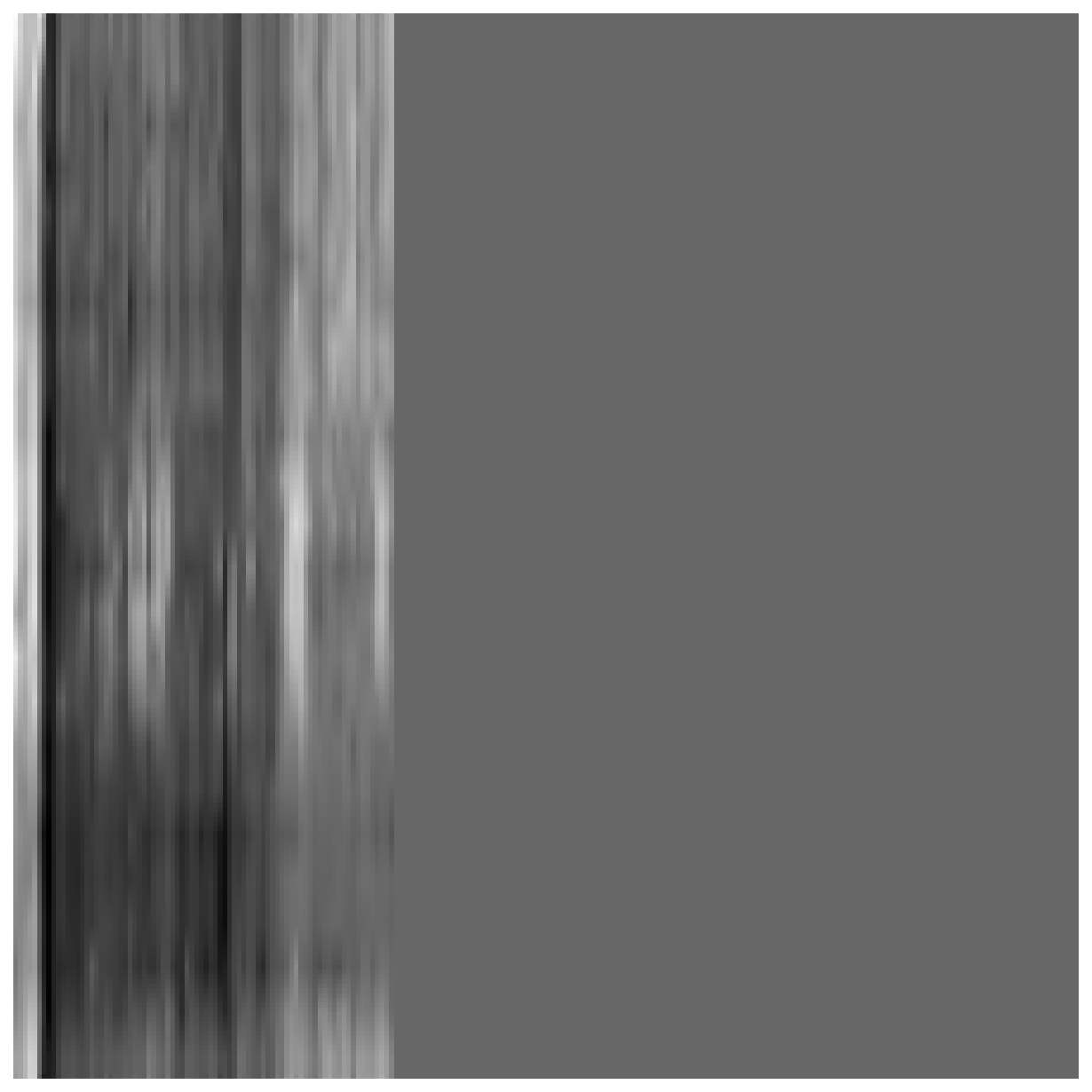}
        \small (c) Spectral branch input image
        \label{fig:etth1_var1_vi}
    \end{minipage}
    \hspace{2em}
    \begin{minipage}[t]{0.20\textwidth}
        \centering
        \includegraphics[width=1\textwidth]{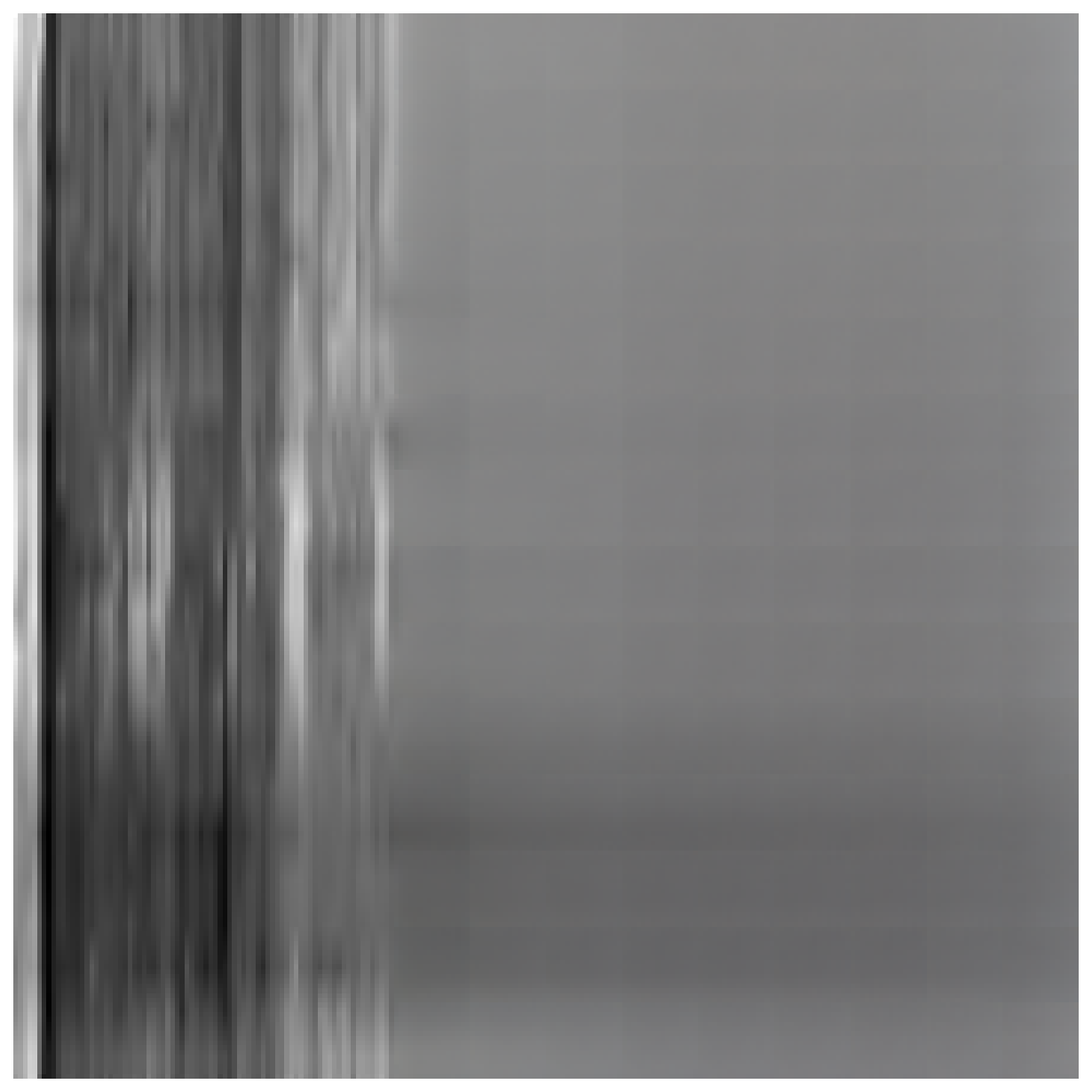}
        \small (d) Spectral branch reconstructed image
        \label{fig:etth1_var1_vr}
    \end{minipage}

    \vspace{1em}

    \begin{minipage}[t]{1\textwidth}
        \centering
        \includegraphics[width=1\textwidth]{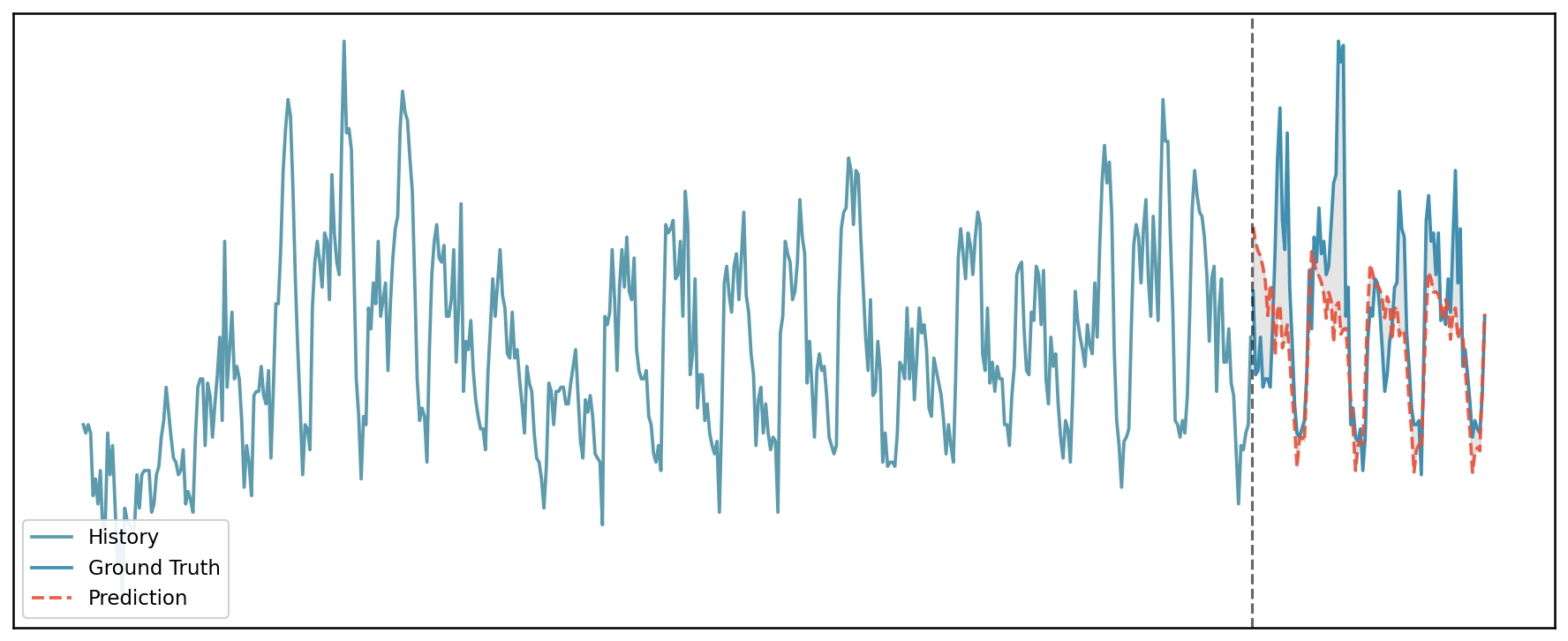}
        \small (e) SSDA (MSE=0.718,MAE=0.678).
        \label{fig:etth1_ssda}
    \end{minipage}

    \vspace{1em}

    \begin{minipage}[t]{1\textwidth}
        \centering
        \includegraphics[width=1\textwidth]{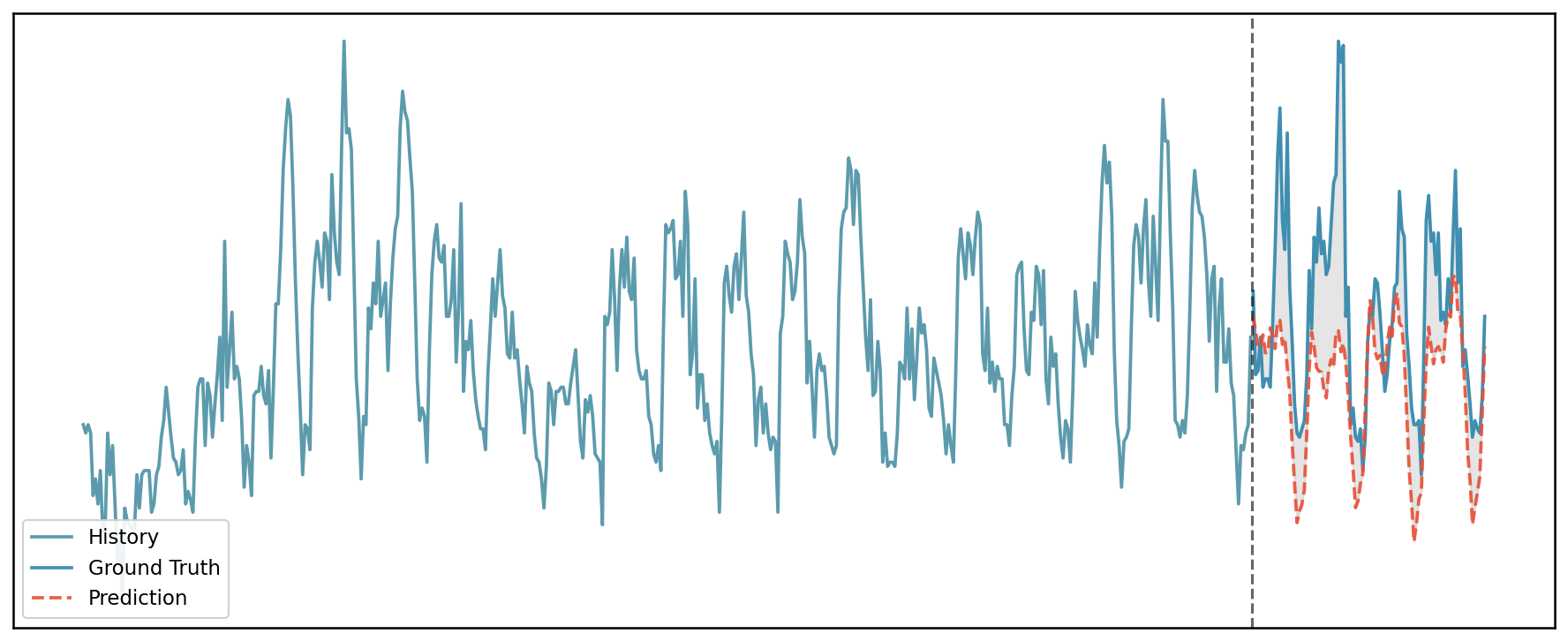}
        \small (f) DMMV (MSE=0.765,MAE=0.694).
        \label{fig:etth1_dmmv}
    \end{minipage}

    \caption{Forecasting visualization on a sample from ETTh1. (a) Input image of structural branch. (b) Output image of structural branch. (c) Input image of spectral branch(enhanced by Spectral Magnitude Aligner). (d) Output image of spectral branch. (e-f) Comparison of predictions and ground truths by SSDA and DMMV.}
    \label{fig:ettm2_combined_visualizations}
\end{figure}

\begin{figure}[H]
    \centering

    \begin{minipage}[t]{0.20\textwidth}
        \centering
        \includegraphics[width=1\textwidth]{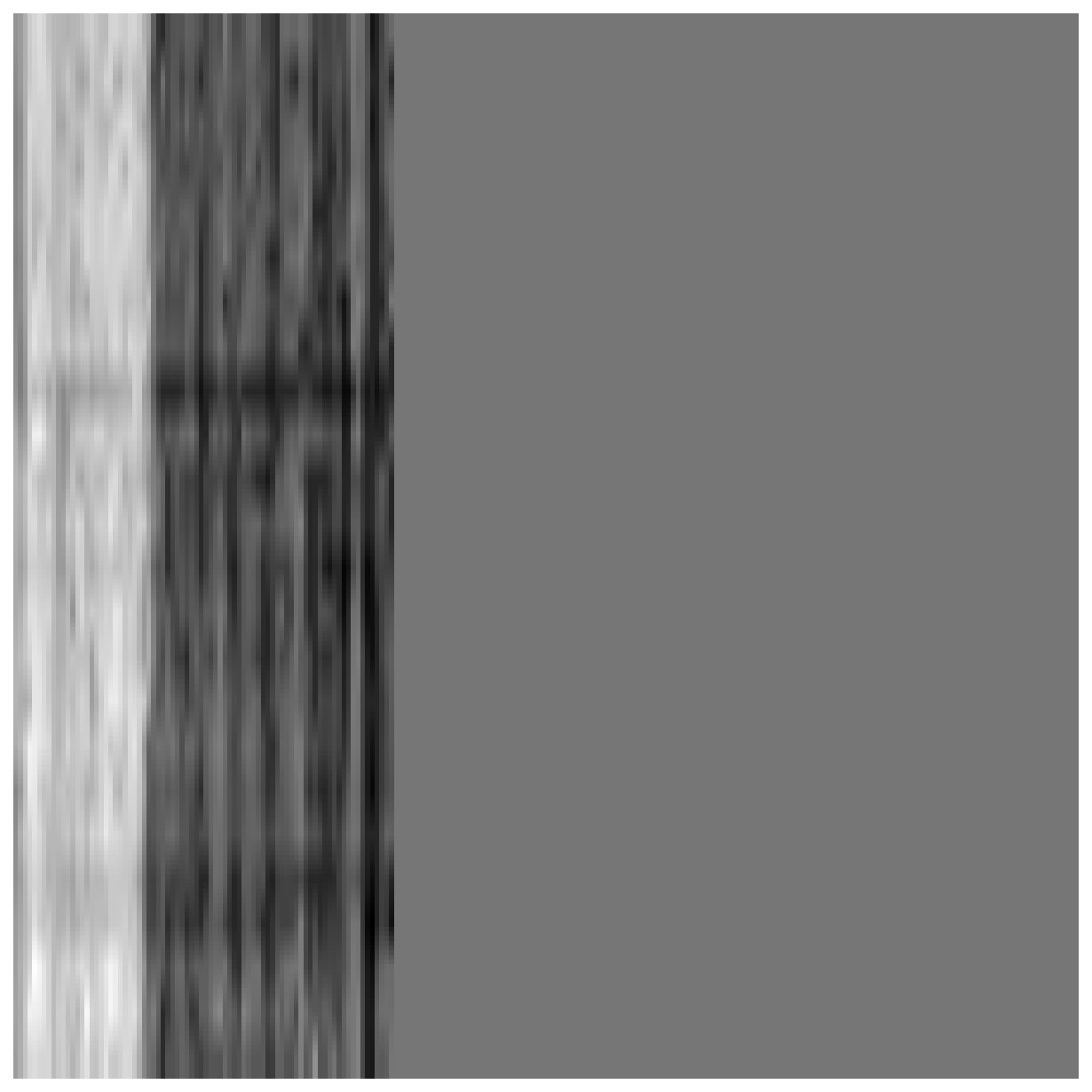}
        \small (a) Structural branch input image
        \label{fig:etth2_var3_ti}
    \end{minipage}
    \hspace{2em}
    \begin{minipage}[t]{0.20\textwidth}
        \centering
        \includegraphics[width=1\textwidth]{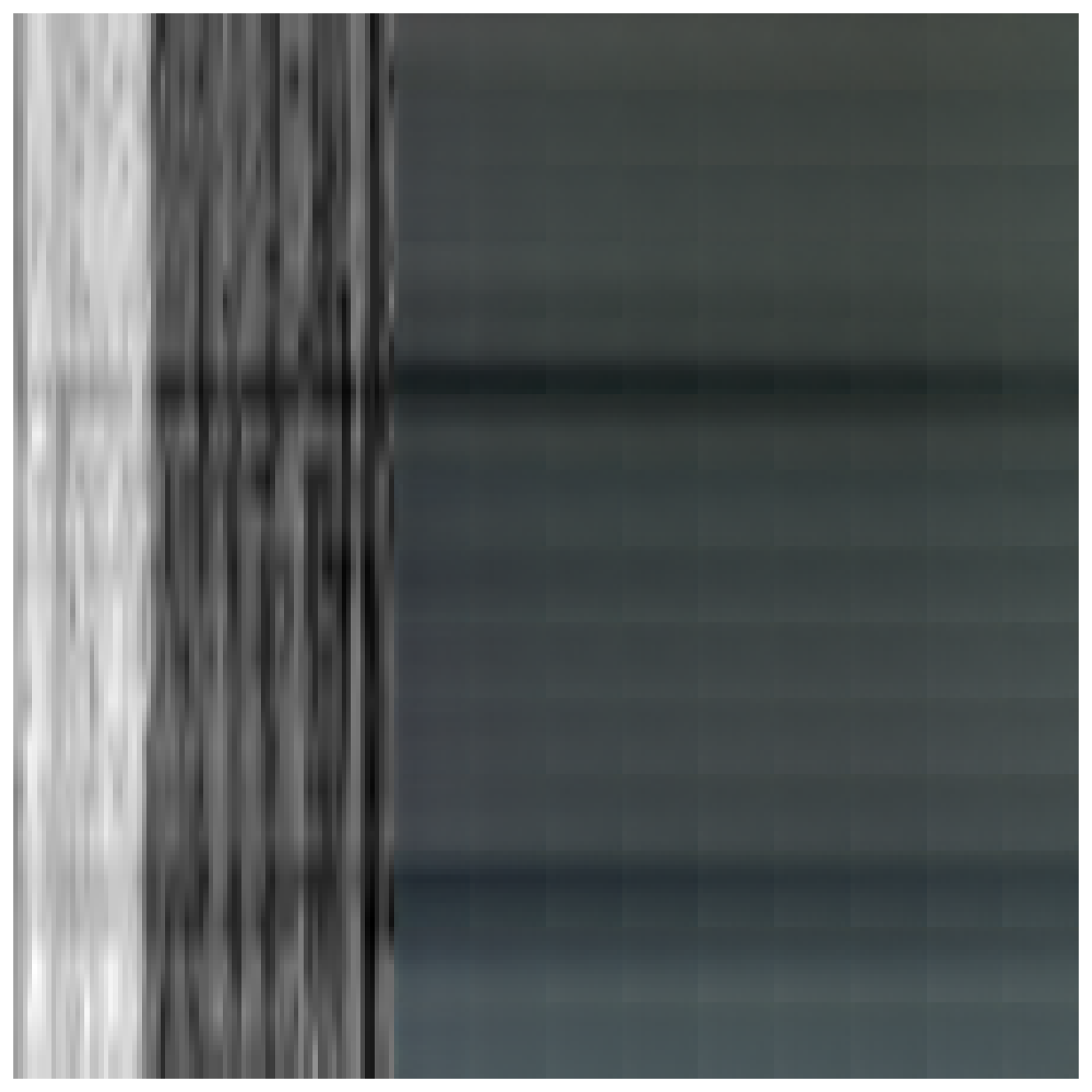}
        \small (b) Structural branch reconstructed image
        \label{fig:etth2_var3_tr}
    \end{minipage}
    \hspace{2em}
    \begin{minipage}[t]{0.20\textwidth}
        \centering
        \includegraphics[width=1\textwidth]{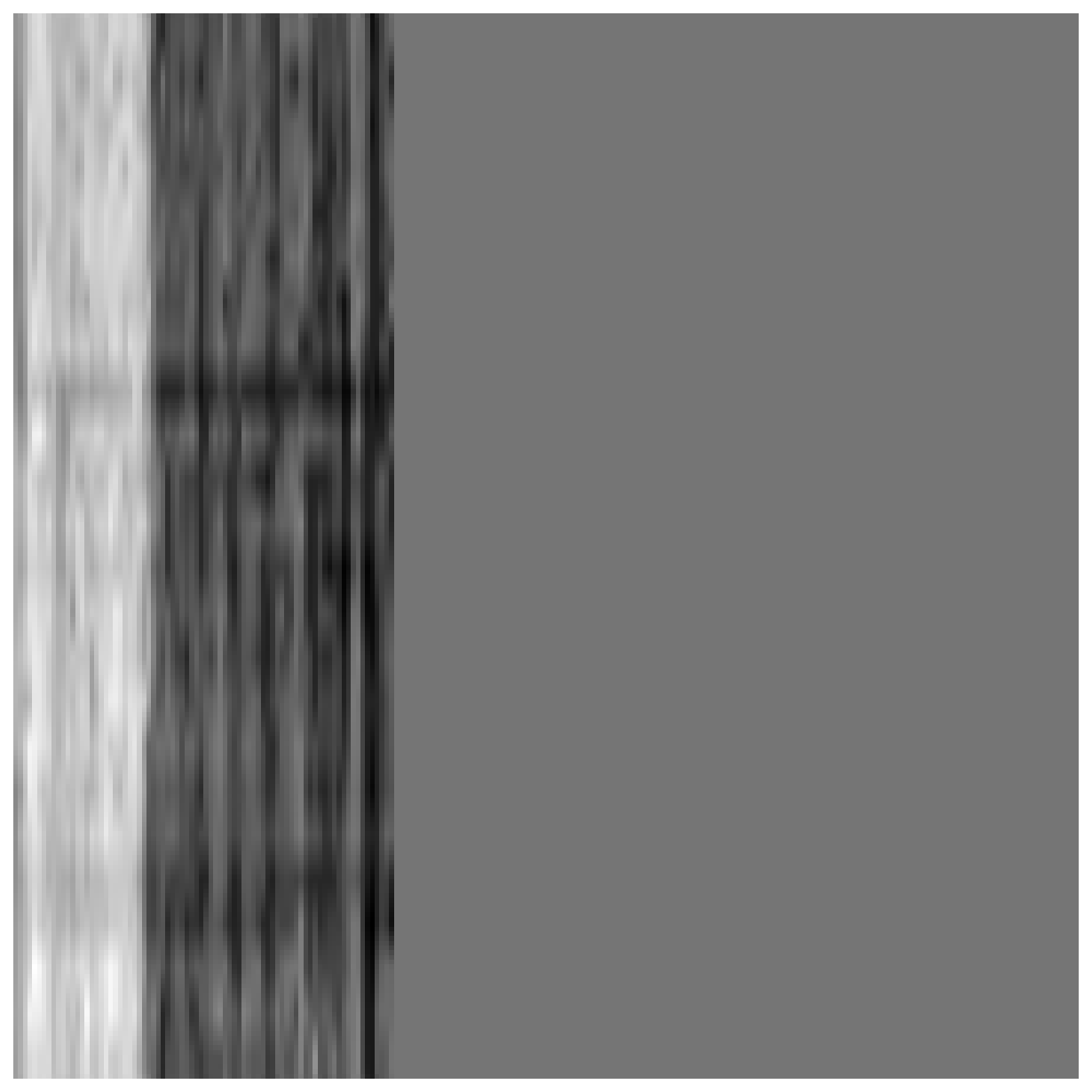}
        \small (c) Spectral branch input image
        \label{fig:etth2_var3_vi}
    \end{minipage}
    \hspace{2em}
    \begin{minipage}[t]{0.20\textwidth}
        \centering
        \includegraphics[width=1\textwidth]{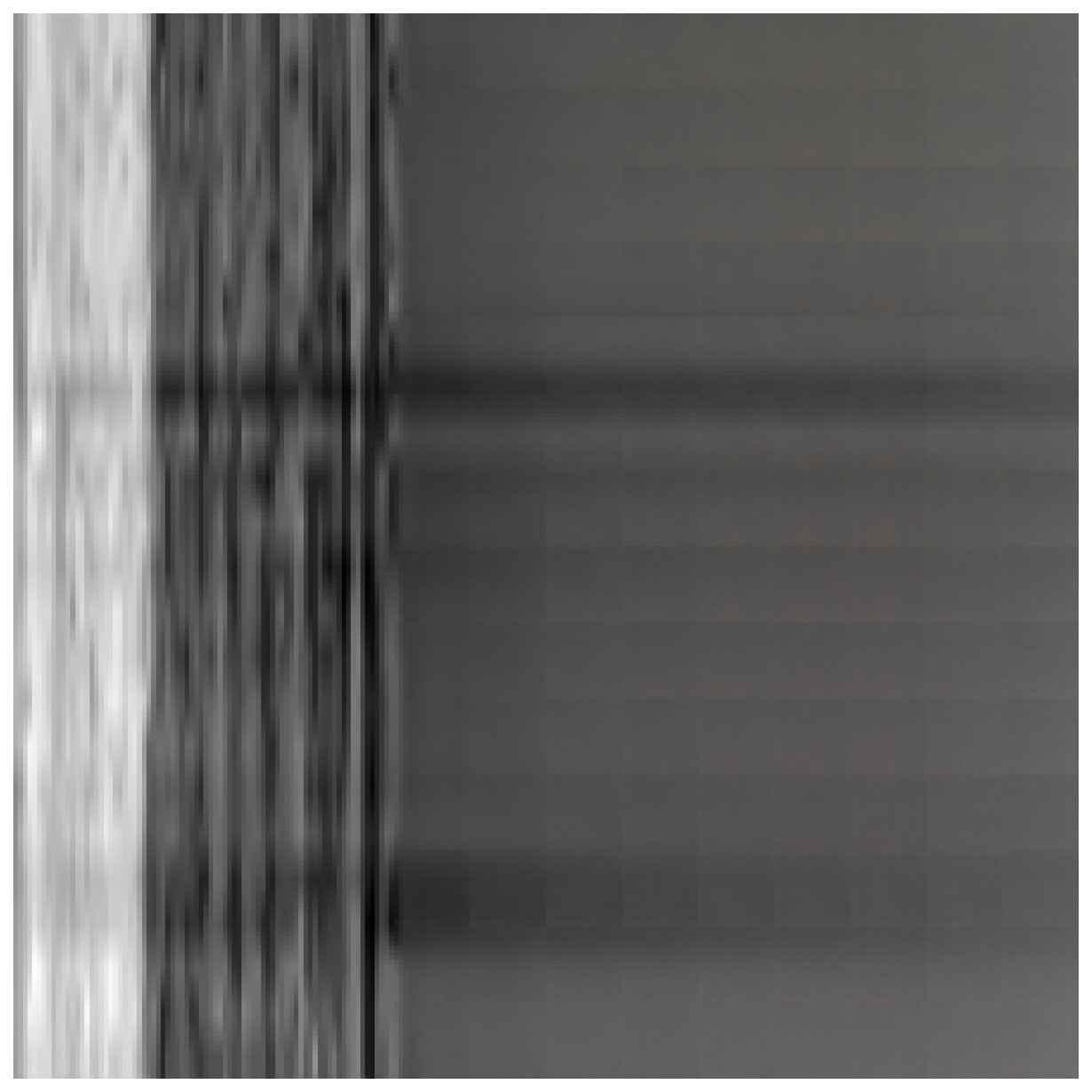}
        \small (d) Spectral branch reconstructed image
        \label{fig:etth2_var3_vr}
    \end{minipage}

    \vspace{1em}

    \begin{minipage}[t]{1\textwidth}
        \centering
        \includegraphics[width=1\textwidth]{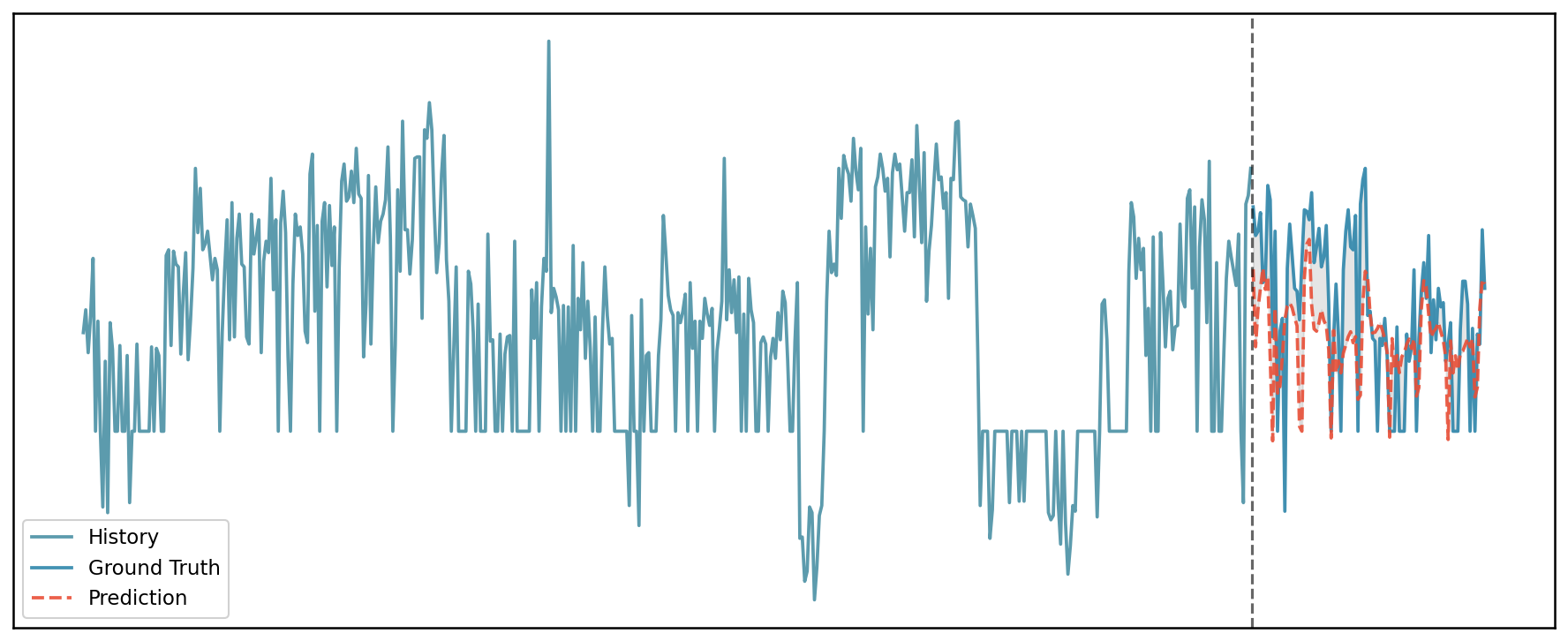}
        \small (e) SSDA (MSE=0.439,MAE=0.522).
        \label{fig:etth2_ssda}
    \end{minipage}

    \vspace{1em}

    \begin{minipage}[t]{1\textwidth}
        \centering
        \includegraphics[width=1\textwidth]{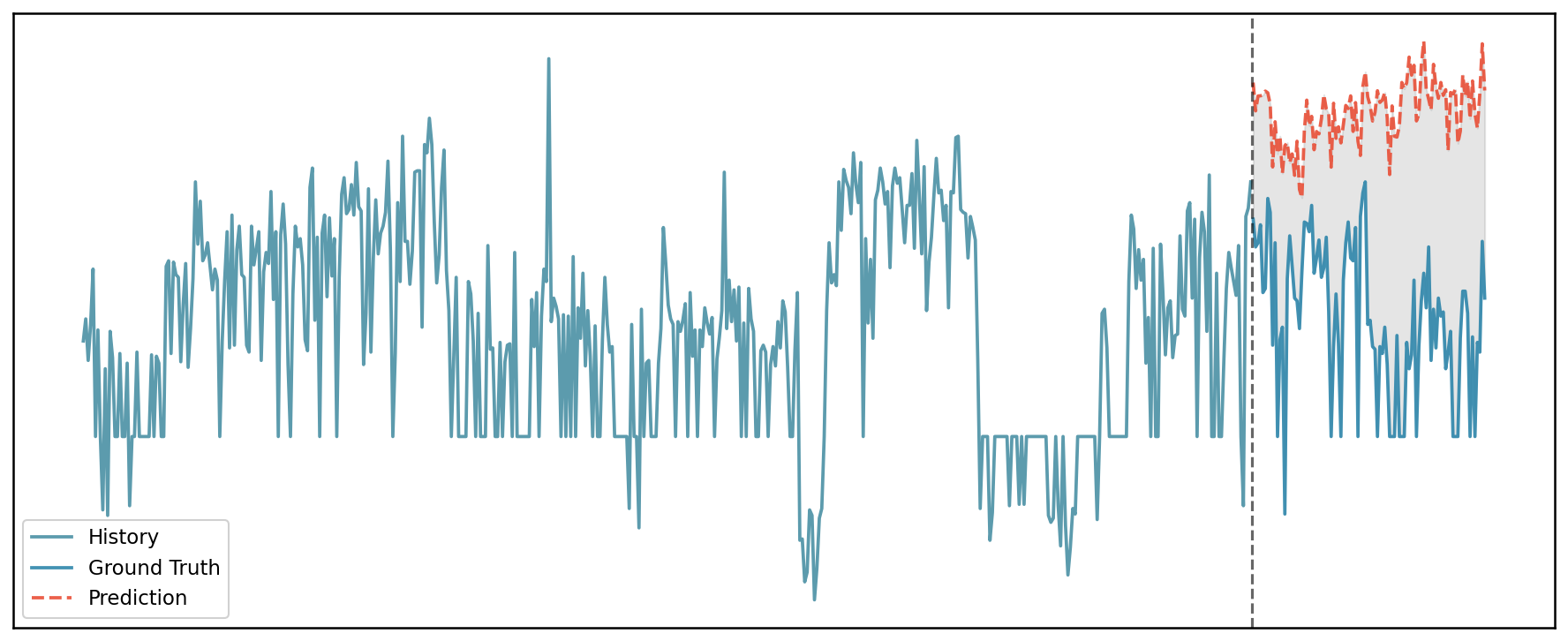}
        \small (f) DMMV (MSE=2.242,MAE=1.370).
        \label{fig:etth2_dmmv}
    \end{minipage}

    \caption{Forecasting visualization on a sample from ETTh2. (a) Input image of structural branch. (b) Output image of structural branch. (c) Input image of spectral branch(enhanced by Spectral Magnitude Aligner). (d) Output image of spectral branch. (e-f) Comparison of predictions and ground truths by SSDA and DMMV.}
    \label{fig:etth2_combined_visualizations}
\end{figure}

\newpage
\section*{NeurIPS Paper Checklist}

\begin{enumerate}

\item {\bf Claims}
    \item[] Question: Do the main claims made in the abstract and introduction accurately reflect the paper's contributions and scope?
    \item[] Answer: \answerYes{}
    \item[] Justification: Yes, the abstract and introduction sections accurately reflect the paper's contributions and scope.
    \item[] Guidelines:
    \begin{itemize}
        \item The answer \answerNA{} means that the abstract and introduction do not include the claims made in the paper.
        \item The abstract and/or introduction should clearly state the claims made, including the contributions made in the paper and important assumptions and limitations. A \answerNo{} or \answerNA{} answer to this question will not be perceived well by the reviewers. 
        \item The claims made should match theoretical and experimental results, and reflect how much the results can be expected to generalize to other settings. 
        \item It is fine to include aspirational goals as motivation as long as it is clear that these goals are not attained by the paper. 
    \end{itemize}

\item {\bf Limitations}
    \item[] Question: Does the paper discuss the limitations of the work performed by the authors?
    \item[] Answer: \answerYes{} 
    \item[] Justification: Yes, for instance, in Appendix \ref{apd:modality_gap_analysis}, we discuss the limitations of PSS, with the hope that other researchers will be able to develop improved approaches for analyzing the modality gap.
    \item[] Guidelines:
    \begin{itemize}
        \item The answer \answerNA{} means that the paper has no limitation while the answer \answerNo{} means that the paper has limitations, but those are not discussed in the paper. 
        \item The authors are encouraged to create a separate ``Limitations'' section in their paper.
        \item The paper should point out any strong assumptions and how robust the results are to violations of these assumptions (e.g., independence assumptions, noiseless settings, model well-specification, asymptotic approximations only holding locally). The authors should reflect on how these assumptions might be violated in practice and what the implications would be.
        \item The authors should reflect on the scope of the claims made, e.g., if the approach was only tested on a few datasets or with a few runs. In general, empirical results often depend on implicit assumptions, which should be articulated.
        \item The authors should reflect on the factors that influence the performance of the approach. For example, a facial recognition algorithm may perform poorly when image resolution is low or images are taken in low lighting. Or a speech-to-text system might not be used reliably to provide closed captions for online lectures because it fails to handle technical jargon.
        \item The authors should discuss the computational efficiency of the proposed algorithms and how they scale with dataset size.
        \item If applicable, the authors should discuss possible limitations of their approach to address problems of privacy and fairness.
        \item While the authors might fear that complete honesty about limitations might be used by reviewers as grounds for rejection, a worse outcome might be that reviewers discover limitations that aren't acknowledged in the paper. The authors should use their best judgment and recognize that individual actions in favor of transparency play an important role in developing norms that preserve the integrity of the community. Reviewers will be specifically instructed to not penalize honesty concerning limitations.
    \end{itemize}

\item {\bf Theory assumptions and proofs}
    \item[] Question: For each theoretical result, does the paper provide the full set of assumptions and a complete (and correct) proof?
    \item[] Answer: \answerYes{} 
    \item[] Justification: In Section~\ref{sec:effectiveness_analysis}, we conduct a comprehensive analysis of the proposed SMA and SG-LoRA modules, verifying their effectiveness in practice. Specifically, the results demonstrate that SMA effectively reduces the modality gap between time series and natural images, while SG-LoRA facilitates large vision models (LVMs) in capturing richer temporal order and structural dependencies.
    \item[] Guidelines:
    \begin{itemize}
        \item The answer \answerNA{} means that the paper does not include theoretical results. 
        \item All the theorems, formulas, and proofs in the paper should be numbered and cross-referenced.
        \item All assumptions should be clearly stated or referenced in the statement of any theorems.
        \item The proofs can either appear in the main paper or the supplemental material, but if they appear in the supplemental material, the authors are encouraged to provide a short proof sketch to provide intuition. 
        \item Inversely, any informal proof provided in the core of the paper should be complemented by formal proofs provided in appendix or supplemental material.
        \item Theorems and Lemmas that the proof relies upon should be properly referenced. 
    \end{itemize}

    \item {\bf Experimental result reproducibility}
    \item[] Question: Does the paper fully disclose all the information needed to reproduce the main experimental results of the paper to the extent that it affects the main claims and/or conclusions of the paper (regardless of whether the code and data are provided or not)?
    \item[] Answer: \answerYes{} 
    \item[] Justification: The paper provides sufficient details to reproduce the main experimental results. Specifically, we clearly describe the datasets, preprocessing procedures, and data splits used in all experiments. The model architecture, including the design of SMA and SG-LoRA, is presented with adequate technical details. Training configurations such as optimizer, learning rate, batch size, and training epochs are explicitly specified. In addition, we report evaluation metrics and experimental settings consistently across all baselines, and ensure fair comparisons by following widely adopted protocols in prior work. These details collectively enable reproducibility of the reported results.
    \item[] Guidelines:
    \begin{itemize}
        \item The answer \answerNA{} means that the paper does not include experiments.
        \item If the paper includes experiments, a \answerNo{} answer to this question will not be perceived well by the reviewers: Making the paper reproducible is important, regardless of whether the code and data are provided or not.
        \item If the contribution is a dataset and\slash or model, the authors should describe the steps taken to make their results reproducible or verifiable. 
        \item Depending on the contribution, reproducibility can be accomplished in various ways. For example, if the contribution is a novel architecture, describing the architecture fully might suffice, or if the contribution is a specific model and empirical evaluation, it may be necessary to either make it possible for others to replicate the model with the same dataset, or provide access to the model. In general. releasing code and data is often one good way to accomplish this, but reproducibility can also be provided via detailed instructions for how to replicate the results, access to a hosted model (e.g., in the case of a large language model), releasing of a model checkpoint, or other means that are appropriate to the research performed.
        \item While NeurIPS does not require releasing code, the conference does require all submissions to provide some reasonable avenue for reproducibility, which may depend on the nature of the contribution. For example
        \begin{enumerate}
            \item If the contribution is primarily a new algorithm, the paper should make it clear how to reproduce that algorithm.
            \item If the contribution is primarily a new model architecture, the paper should describe the architecture clearly and fully.
            \item If the contribution is a new model (e.g., a large language model), then there should either be a way to access this model for reproducing the results or a way to reproduce the model (e.g., with an open-source dataset or instructions for how to construct the dataset).
            \item We recognize that reproducibility may be tricky in some cases, in which case authors are welcome to describe the particular way they provide for reproducibility. In the case of closed-source models, it may be that access to the model is limited in some way (e.g., to registered users), but it should be possible for other researchers to have some path to reproducing or verifying the results.
        \end{enumerate}
    \end{itemize}

\item {\bf Open access to data and code}
    \item[] Question: Does the paper provide open access to the data and code, with sufficient instructions to faithfully reproduce the main experimental results, as described in supplemental material?
    \item[] Answer: \answerYes{} 
    \item[] Justification: We provide open access to both the code and data through an anonymized repository included in the supplementary material. The repository contains complete training and evaluation pipelines, along with scripts to reproduce all main experimental results reported in the paper. Detailed instructions are provided, including environment setup (e.g., Python version and required dependencies), dataset acquisition and preprocessing steps, and exact commands to run each experiment. 
    \item[] Guidelines:
    \begin{itemize}
        \item The answer \answerNA{} means that paper does not include experiments requiring code.
        \item Please see the NeurIPS code and data submission guidelines (\url{https://neurips.cc/public/guides/CodeSubmissionPolicy}) for more details.
        \item While we encourage the release of code and data, we understand that this might not be possible, so \answerNo{} is an acceptable answer. Papers cannot be rejected simply for not including code, unless this is central to the contribution (e.g., for a new open-source benchmark).
        \item The instructions should contain the exact command and environment needed to run to reproduce the results. See the NeurIPS code and data submission guidelines (\url{https://neurips.cc/public/guides/CodeSubmissionPolicy}) for more details.
        \item The authors should provide instructions on data access and preparation, including how to access the raw data, preprocessed data, intermediate data, and generated data, etc.
        \item The authors should provide scripts to reproduce all experimental results for the new proposed method and baselines. If only a subset of experiments are reproducible, they should state which ones are omitted from the script and why.
        \item At submission time, to preserve anonymity, the authors should release anonymized versions (if applicable).
        \item Providing as much information as possible in supplemental material (appended to the paper) is recommended, but including URLs to data and code is permitted.
    \end{itemize}

\item {\bf Experimental setting/details}
    \item[] Question: Does the paper specify all the training and test details (e.g., data splits, hyperparameters, how they were chosen, type of optimizer) necessary to understand the results?
    \item[] Answer: \answerYes{} 
    \item[] Justification: The paper specifies all essential experimental settings required to understand and interpret the results. We clearly describe the data splits for training, validation, and testing across all datasets, along with preprocessing procedures. Key training details, including model architecture, optimizer type, learning rate, batch size, and number of training epochs, are explicitly provided. We also explain the selection of hyperparameters, following standard practices or prior work where applicable, and ensure consistency across all compared methods for fair evaluation. Additional implementation details and configuration settings are included in the appendix and supplementary material, providing a complete reference for reproducing and understanding the experimental outcomes.
    \item[] Guidelines:
    \begin{itemize}
        \item The answer \answerNA{} means that the paper does not include experiments.
        \item The experimental setting should be presented in the core of the paper to a level of detail that is necessary to appreciate the results and make sense of them.
        \item The full details can be provided either with the code, in appendix, or as supplemental material.
    \end{itemize}

\item {\bf Experiment statistical significance}
    \item[] Question: Does the paper report error bars suitably and correctly defined or other appropriate information about the statistical significance of the experiments?
    \item[] Answer: \answerYes{} 
    \item[] Justification: The reported results are obtained by selecting the best-performing configuration after multiple experimental runs and hyperparameter tuning. 
    \item[] Guidelines:
    \begin{itemize}
        \item The answer \answerNA{} means that the paper does not include experiments.
        \item The authors should answer \answerYes{} if the results are accompanied by error bars, confidence intervals, or statistical significance tests, at least for the experiments that support the main claims of the paper.
        \item The factors of variability that the error bars are capturing should be clearly stated (for example, train/test split, initialization, random drawing of some parameter, or overall run with given experimental conditions).
        \item The method for calculating the error bars should be explained (closed form formula, call to a library function, bootstrap, etc.)
        \item The assumptions made should be given (e.g., Normally distributed errors).
        \item It should be clear whether the error bar is the standard deviation or the standard error of the mean.
        \item It is OK to report 1-sigma error bars, but one should state it. The authors should preferably report a 2-sigma error bar than state that they have a 96\% CI, if the hypothesis of Normality of errors is not verified.
        \item For asymmetric distributions, the authors should be careful not to show in tables or figures symmetric error bars that would yield results that are out of range (e.g., negative error rates).
        \item If error bars are reported in tables or plots, the authors should explain in the text how they were calculated and reference the corresponding figures or tables in the text.
    \end{itemize}

\item {\bf Experiments compute resources}
    \item[] Question: For each experiment, does the paper provide sufficient information on the computer resources (type of compute workers, memory, time of execution) needed to reproduce the experiments?
    \item[] Answer: \answerYes{} 
    \item[] Justification: All experiments were conducted on  NVIDIA RTX 4090 24GB GPUs. We report the GPU type, memory consumption, training time, and number of parameters for each main experiment.
    \item[] Guidelines:
    \begin{itemize}
        \item The answer \answerNA{} means that the paper does not include experiments.
        \item The paper should indicate the type of compute workers CPU or GPU, internal cluster, or cloud provider, including relevant memory and storage.
        \item The paper should provide the amount of compute required for each of the individual experimental runs as well as estimate the total compute. 
        \item The paper should disclose whether the full research project required more compute than the experiments reported in the paper (e.g., preliminary or failed experiments that didn't make it into the paper). 
    \end{itemize}
    
\item {\bf Code of ethics}
    \item[] Question: Does the research conducted in the paper conform, in every respect, with the NeurIPS Code of Ethics \url{https://neurips.cc/public/EthicsGuidelines}?
\item[] Answer: \answerYes{}
\item[] Justification: We have reviewed the NeurIPS Code of Ethics and confirm that our research conforms with it in every respect. The work involves only publicly available time series benchmarks and does not involve human subjects, sensitive personal data, or applications with foreseeable harm.
    \item[] Guidelines:
    \begin{itemize}
        \item The answer \answerNA{} means that the authors have not reviewed the NeurIPS Code of Ethics.
        \item If the authors answer \answerNo, they should explain the special circumstances that require a deviation from the Code of Ethics.
        \item The authors should make sure to preserve anonymity (e.g., if there is a special consideration due to laws or regulations in their jurisdiction).
    \end{itemize}

\item {\bf Broader impacts}
    \item[] Question: Does the paper discuss both potential positive societal impacts and negative societal impacts of the work performed?
    \item[] Answer: \answerYes{}
    \item[] Justification: Our method improves time series forecasting by leveraging large vision models, which can benefit applications such as energy consumption prediction, weather forecasting, traffic management, and healthcare monitoring. 
    \item[] Guidelines:
    \begin{itemize}
        \item The answer \answerNA{} means that there is no societal impact of the work performed.
        \item If the authors answer \answerNA{} or \answerNo, they should explain why their work has no societal impact or why the paper does not address societal impact.
        \item Examples of negative societal impacts include potential malicious or unintended uses (e.g., disinformation, generating fake profiles, surveillance), fairness considerations (e.g., deployment of technologies that could make decisions that unfairly impact specific groups), privacy considerations, and security considerations.
        \item The conference expects that many papers will be foundational research and not tied to particular applications, let alone deployments. However, if there is a direct path to any negative applications, the authors should point it out. For example, it is legitimate to point out that an improvement in the quality of generative models could be used to generate Deepfakes for disinformation. On the other hand, it is not needed to point out that a generic algorithm for optimizing neural networks could enable people to train models that generate Deepfakes faster.
        \item The authors should consider possible harms that could arise when the technology is being used as intended and functioning correctly, harms that could arise when the technology is being used as intended but gives incorrect results, and harms following from (intentional or unintentional) misuse of the technology.
        \item If there are negative societal impacts, the authors could also discuss possible mitigation strategies (e.g., gated release of models, providing defenses in addition to attacks, mechanisms for monitoring misuse, mechanisms to monitor how a system learns from feedback over time, improving the efficiency and accessibility of ML).
    \end{itemize}
    
\item {\bf Safeguards}
    \item[] Question: Does the paper describe safeguards that have been put in place for responsible release of data or models that have a high risk for misuse (e.g., pre-trained language models, image generators, or scraped datasets)?
    \item[] Answer: \answerNA{}
    \item[] Justification: The paper does not release pre-trained generative models, scraped datasets, or any other assets with high risk for misuse. Our released code and trained checkpoints are for time series analysis on standard public benchmarks and pose no such risks.
    \item[] Guidelines:
    \begin{itemize}
        \item The answer \answerNA{} means that the paper poses no such risks.
        \item Released models that have a high risk for misuse or dual-use should be released with necessary safeguards to allow for controlled use of the model, for example by requiring that users adhere to usage guidelines or restrictions to access the model or implementing safety filters. 
        \item Datasets that have been scraped from the Internet could pose safety risks. The authors should describe how they avoided releasing unsafe images.
        \item We recognize that providing effective safeguards is challenging, and many papers do not require this, but we encourage authors to take this into account and make a best faith effort.
    \end{itemize}

\item {\bf Licenses for existing assets}
    \item[] Question: Are the creators or original owners of assets (e.g., code, data, models), used in the paper, properly credited and are the license and terms of use explicitly mentioned and properly respected?
    \item[] Answer: \answerYes{} 
    \item[] Justification: All datasets used (e.g., ETT, Weather, Traffic, Electricity) are publicly available time series benchmarks that have been widely used in prior work, and we cite their original sources. The pre-trained vision models (e.g., CLIP and MAE) are used under their respective licenses. All third-party code is properly credited in the paper and repository.
    \item[] Guidelines:
    \begin{itemize}
        \item The answer \answerNA{} means that the paper does not use existing assets.
        \item The authors should cite the original paper that produced the code package or dataset.
        \item The authors should state which version of the asset is used and, if possible, include a URL.
        \item The name of the license (e.g., CC-BY 4.0) should be included for each asset.
        \item For scraped data from a particular source (e.g., website), the copyright and terms of service of that source should be provided.
        \item If assets are released, the license, copyright information, and terms of use in the package should be provided. For popular datasets, \url{paperswithcode.com/datasets} has curated licenses for some datasets. Their licensing guide can help determine the license of a dataset.
        \item For existing datasets that are re-packaged, both the original license and the license of the derived asset (if it has changed) should be provided.
        \item If this information is not available online, the authors are encouraged to reach out to the asset's creators.
    \end{itemize}

\item {\bf New assets}
    \item[] Question: Are new assets introduced in the paper well documented and is the documentation provided alongside the assets?
    \item[] Answer: \answerYes{}
    \item[] Justification: We release our code as a new asset, accompanied by a README with setup instructions, training/evaluation scripts, configuration files, and documentation of the SMA and SG-LoRA modules. The anonymized repository is provided in the supplementary material, and a non-anonymized version with a license (e.g., MIT) will be released upon acceptance.
    \item[] Guidelines:
    \begin{itemize}
        \item The answer \answerNA{} means that the paper does not release new assets.
        \item Researchers should communicate the details of the dataset\slash code\slash model as part of their submissions via structured templates. This includes details about training, license, limitations, etc. 
        \item The paper should discuss whether and how consent was obtained from people whose asset is used.
        \item At submission time, remember to anonymize your assets (if applicable). You can either create an anonymized URL or include an anonymized zip file.
    \end{itemize}

\item {\bf Crowdsourcing and research with human subjects}
    \item[] Question: For crowdsourcing experiments and research with human subjects, does the paper include the full text of instructions given to participants and screenshots, if applicable, as well as details about compensation (if any)? 
    \item[] Answer: \answerNA{}
    \item[] Justification: The paper does not involve crowdsourcing or research with human subjects. All experiments are conducted on publicly available time series benchmarks.
    \item[] Guidelines:
    \begin{itemize}
        \item The answer \answerNA{} means that the paper does not involve crowdsourcing nor research with human subjects.
        \item Including this information in the supplemental material is fine, but if the main contribution of the paper involves human subjects, then as much detail as possible should be included in the main paper. 
        \item According to the NeurIPS Code of Ethics, workers involved in data collection, curation, or other labor should be paid at least the minimum wage in the country of the data collector. 
    \end{itemize}

\item {\bf Institutional review board (IRB) approvals or equivalent for research with human subjects}
    \item[] Question: Does the paper describe potential risks incurred by study participants, whether such risks were disclosed to the subjects, and whether Institutional Review Board (IRB) approvals (or an equivalent approval/review based on the requirements of your country or institution) were obtained?
    \item[] Answer: \answerNA{}
    \item[] Justification: The paper does not involve research with human subjects, so IRB approval is not applicable.
    \item[] Guidelines:
    \begin{itemize}
        \item The answer \answerNA{} means that the paper does not involve crowdsourcing nor research with human subjects.
        \item Depending on the country in which research is conducted, IRB approval (or equivalent) may be required for any human subjects research. If you obtained IRB approval, you should clearly state this in the paper. 
        \item We recognize that the procedures for this may vary significantly between institutions and locations, and we expect authors to adhere to the NeurIPS Code of Ethics and the guidelines for their institution. 
        \item For initial submissions, do not include any information that would break anonymity (if applicable), such as the institution conducting the review.
    \end{itemize}

\item {\bf Declaration of LLM usage}
    \item[] Question: Does the paper describe the usage of LLMs if it is an important, original, or non-standard component of the core methods in this research? Note that if the LLM is used only for writing, editing, or formatting purposes and does \emph{not} impact the core methodology, scientific rigor, or originality of the research, declaration is not required.
\item[] Answer: \answerNA{}
\item[] Justification: The use of LLMs was limited to writing assistance, editing, and formatting, which does not require declaration per the NeurIPS LLM policy.
    \item[] Guidelines:
    \begin{itemize}
        \item The answer \answerNA{} means that the core method development in this research does not involve LLMs as any important, original, or non-standard components.
        \item Please refer to our LLM policy in the NeurIPS handbook for what should or should not be described.
    \end{itemize}

\end{enumerate}

\end{document}